\documentclass{article}

 \usepackage[eandd, final]{neurips_2026}

\usepackage[utf8]{inputenc} 
\usepackage[T1]{fontenc}    
\usepackage{hyperref}       
\usepackage{url}            
\usepackage{booktabs}       
\usepackage{amsfonts}       
\usepackage{nicefrac}       
\usepackage{microtype}      
\usepackage{xcolor}         
\usepackage{natbib}
\usepackage{graphicx}
\usepackage{multirow} 
\usepackage{amssymb}
\usepackage{graphicx}
\usepackage{tabularray}
\UseTblrLibrary{booktabs}

\usepackage{tcolorbox}
\usepackage{enumitem}
\usepackage{amsmath}
\usepackage{makecell}
\usepackage{listings}
\usepackage{threeparttable}
\usepackage{subcaption}
\usepackage{tcolorbox}
\usepackage{enumitem}
\usepackage{xcolor}
\usepackage{array}

\definecolor{myblue}{HTML}{00A0E9}
\definecolor{mygreen}{HTML}{009944}

\lstdefinestyle{datasetstyle}{
    basicstyle=\ttfamily\small,   
    frame=single,                 
    backgroundcolor=\color{gray!10}, 
    breaklines=true,              
    postbreak=\mbox{\textcolor{red}{$\hookrightarrow$}\space}, 
    showstringspaces=false,       
}
\usepackage{fontawesome5}

\title{RAM-H1200: A Unified Evaluation and Dataset on Hand Radiographs for Rheumatoid Arthritis}

\author{%
  \textbf{Songxiao Yang}$^1$\thanks{Equal Contribution.}\quad \textbf{Haolin Wang}$^{2\,*}$\quad \textbf{Yao Fu}$^{2\,*}$\quad \textbf{Junmu Peng}$^2$ \\ \textbf{Lin Fan}$^3$ \quad \textbf{Hongruixuan Chen}$^4$ \quad \textbf{Jian Song}$^4$ \quad \textbf{Masayuki Ikebe}$^2$\quad \\ \textbf{Shinya Takamaeda-Yamazaki}$^{4,5}$\quad \textbf{Masatoshi Okutomi}$^1$ \quad \textbf{Tamotsu Kamishima}$^2$ \quad \textbf{Yafei Ou}$^4$\thanks{Corresponding Author: Yafei Ou (\texttt{yafei.ou@riken.jp})}\\
  $^1$ Institute of Science Tokyo, Tokyo, Japan \\
  $^2$ Hokkaido University, Sapporo, Japan \\
  $^3$ Southwest Jiaotong University, Chengdu, China \\
  $^4$ RIKEN, Tokyo, Japan \\
  $^5$ The University of Tokyo, Tokyo, Japan
}

\begin{document}

\maketitle

\begin{abstract}
Rheumatoid arthritis (RA) assessment from hand radiographs requires multi-level analysis and modeling of anatomical structures and fine-grained local pathological changes. However, existing public resources do not support such multi-level analysis, often lacking full-hand coverage, fine-grained annotations, and consistent integration with clinical scoring systems. In particular, annotations that enable quantitative analysis of bone erosion (BE) remain scarce.
RAM-H1200 contains 1,200 hand radiographs from a regional, multi-institutional Japanese cohort collected across six medical centers in the Sapporo region, with multi-level annotations including (i) whole-hand bone structure instance segmentation, (ii) pixel-level BE masks, (iii) Sharp/van der Heijde (SvdH)-defined joint regions of interest, and (iv) joint-level SvdH scores for both BE and joint space narrowing (JSN). 
It provides a common evaluation resource covering anatomical structure modeling, localized erosive pathology analysis, and clinically standardized RA severity assessment from the same radiographs.
The proposed BE masks enable, for the first time, quantitative BE analysis beyond coarse categorical grading by providing explicit spatial supervision for lesion extent and morphology. 
To our knowledge, RAM-H1200 is the first public benchmark resource providing whole-hand bone structure instance segmentation, pixel-level BE delineation, and clinically grounded joint-level SvdH scoring for both BE and JSN. 
Results across benchmark tasks show that anatomical modeling is substantially more mature than quantitative BE analysis: whole-hand bone segmentation achieves strong performance, whereas BE segmentation remains a major open challenge. 
By providing multi-level annotations spanning anatomical structure modeling, quantitative lesion analysis, and clinically grounded SvdH scoring, RAM-H1200 provides a comprehensive benchmark resource for RA analysis on hand radiographs.
  \\
  \small \textbf{\mbox{\faGithub\hspace{.25em} Benchmark \& Code:}} \href{https://github.com/YSongxiao/RAM-H1200}{github.com/YSongxiao/RAM-H1200}\\
  \raisebox{-0.3\height}{\includegraphics[width=0.35cm]{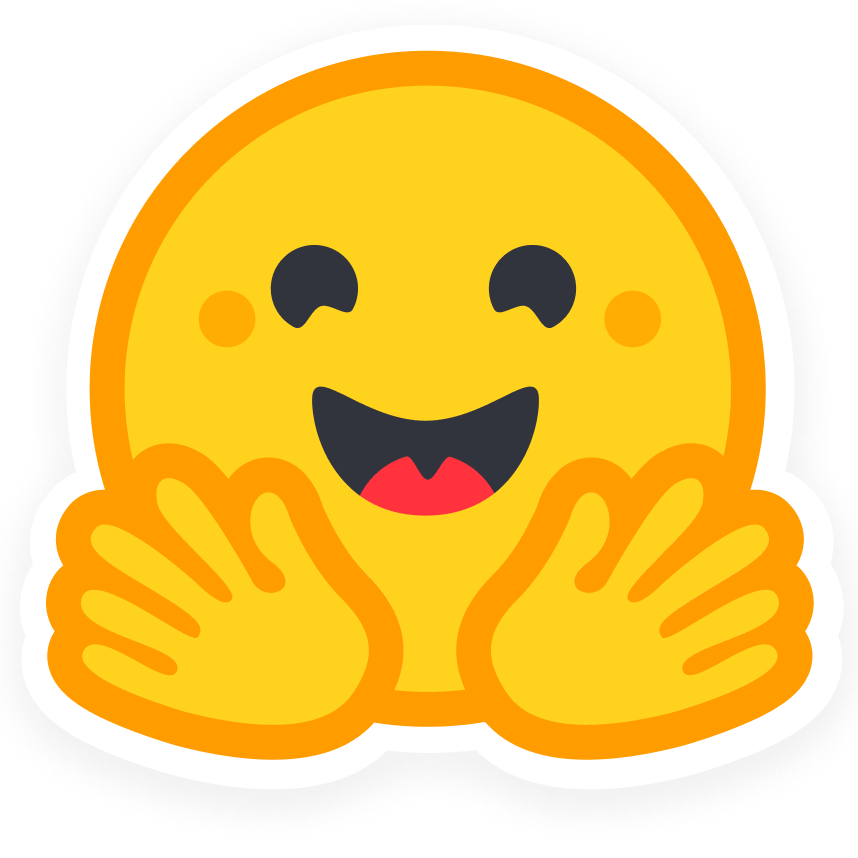}} \small \textbf{\mbox{Dataset Repository:}} \href{https://huggingface.co/datasets/TokyoTechMagicYang/RAM-H1200-v1}{huggingface.co/datasets/TokyoTechMagicYang/RAM-H1200-v1}
\end{abstract}

\section{Introduction}
\label{sec:intro}
Rheumatoid arthritis (RA) is a chronic autoimmune disease that frequently affects the small joints of the hands at an early stage, leading to progressive structural damage and functional impairment~\cite{sharif2018rheumatoid, komatsu2022mechanisms, smolen2018rheumatoid}. Hand radiographs remain a cornerstone for assessing disease severity and monitoring longitudinal progression in RA~\cite{aletaha2018diagnosis}, where bone erosion (BE) and joint space narrowing (JSN) are widely recognized as key radiographic manifestations of structural damage~\cite{schett2012bone, ponnusamy2023automatic}. Clinically, these changes are commonly evaluated using standardized scoring systems such as the Sharp/van der Heijde (SvdH) score~\cite{van2000read}. Nevertheless, conventional evaluation of hand radiographs largely depends on expert-driven visual inspection, which is inherently subjective, labor-intensive, and prone to inter-observer variability~\cite{sharp2004variability}, especially when identifying subtle or early-stage abnormalities. These limitations have motivated increasing efforts toward the development of automated computer-aided diagnosis (CAD) approaches~\cite{stoel2024deep, kingsmore2021introduction}, aiming to provide more consistent, efficient, and sensitive assessment of pathological changes in hand imaging~\cite{wang2025bls, wang2025layer, ou2019automatic, ou2022sub, wang2023deep, ou2025computer}.

Automated RA analysis from hand radiographs involves a hierarchical pipeline ranging from anatomical structure modeling to lesion quantification and clinically standardized scoring. At the anatomical level, accurate identification of individual hand bones provides the spatial foundation for localizing clinically relevant joints and cortical surfaces~\cite{filippucci2014progress}. At the pathological level, pixel-level BE segmentation enables quantitative characterization of erosive lesions, including their location, extent, and morphology~\cite{woodworth2017examining, borrero2011emerging}. At the clinical level, SvdH-based BE and JSN scoring summarize structural damage into standardized ordinal grades. These levels are closely connected: BE analysis depends on anatomical context because erosions predominantly occur near cortical margins and joint interfaces, while JSN assessment relies on the relative configuration of adjacent bones and joint geometry~\cite{schett2012bone, van1996plain, miyama2022deep}. Therefore, comprehensive RA radiograph understanding requires reasoning across anatomical structure, localized pathology, and clinically grounded scoring within a common evaluation framework~\cite{minopoulou2023imaging}.

However, existing hand and wrist radiograph datasets provide limited support for this comprehensive analysis paradigm~\cite{lin2022precision}. As summarized in Table~\ref{tab:dataset_compare}, large-scale hand or wrist radiograph datasets such as DHA~\cite{gertych2007bone}, RSNA Bone Age~\cite{halabi2019rsna}, and MURA~\cite{rajpurkar2017mura} mainly provide global labels or abnormality annotations, without fine-grained structural or pathological masks. Datasets with richer annotations, such as GRAZPEDWRI-DX~\cite{nagy2022pediatric}, focus on trauma rather than RA. RA-specific datasets such as RA2-DREAM~\cite{sun2022crowdsourcing} provide joint-level severity scores but lack pixel-wise annotations, while RAM-W600~\cite{yang2025ram} includes segmentation and scoring annotations but is limited to the wrist region. Consequently, datasets that simultaneously provide full-hand coverage, RA-specific annotations, pixel-level masks, SvdH-based scoring, and multi-center data remain scarce~\cite{lin2022precision}.

\begin{table*}[!t]
\centering
\caption{Comparison between RAM-H1200 and publicly available hand/wrist radiograph datasets. \textbf{Ann/Img}: Annotations per image; \textbf{F}, \textbf{C}, and \textbf{UR} denote segmentation annotations; \textbf{BE} and \textbf{JSN} denote score annotations. For RAM-H1200, $30+3$ denotes 30 bone structures and 3 BE mask classes, while 31 denotes 16 BE and 15 JSN joints.}
\label{tab:dataset_compare}
\begin{threeparttable}
\resizebox{\textwidth}{!}{
\begin{tabular}{cccccccccccc}
\toprule
\multirow{2.5}{*}{\textbf{Dataset}} &
\multicolumn{2}{c}{\textbf{Images (Ann/Img)}} &
\multirow{2.5}{*}{\textbf{\makecell{Age\\(Mean$\pm$SD)}}} &
\multirow{2.5}{*}{\textbf{Centers}} &
\multirow{2.5}{*}{\textbf{Patients}} &
\multirow{2.5}{*}{\textbf{Purpose}} &
\multicolumn{5}{c}{\textbf{RA-related Annotation}} \\
\cmidrule(lr){2-3}
\cmidrule(lr){8-10}
\cmidrule(lr){11-12}
& \textbf{Seg} & \textbf{Score} &
& & & &
\textbf{F} & \textbf{C} & \textbf{UR} &
\textbf{BE} & \textbf{JSN} \\
\midrule

DHA~\cite{gertych2007bone}
& -
& 1400 (1)
& -
& 1
& -
& BAA
&  &  & 
&  & \\

MURA~\cite{rajpurkar2017mura}
& -
& 40561 (1)
& -
& 1
& 12173
& Abnormality
&  &  & 
&  & \\

RSNA Bone Age~\cite{halabi2019rsna}
& -
& 14236 (1)
& 10.59
& 2
& -
& BAA
&  &  & 
&  & \\

GRAZPEDWRI-DX~\cite{nagy2022pediatric}
& 20327 (2)
& -
& 10.9
& 1
& 6091
& Trauma
&  &  &
&  &  \\

\midrule

RA2-DREAM~\cite{sun2022crowdsourcing}
& -
& 674 (31)
& -
& -
& 562
& RA
&  &  & 
& \checkmark & \checkmark \\

RAM-W600~\cite{yang2025ram}
& 618 (14)
& 800 (6)
& 49.86$\pm$20.26
& 6
& 388
& RA
&  & \checkmark & \checkmark
& $\triangle$ &  \\

RAM-H1200 (Ours)
& 1200 (30+3)
& 1200 (31)
& 57.70$\pm$13.76
& 6
& 291
& RA
& \checkmark & \checkmark & \checkmark
& \checkmark & \checkmark \\

\bottomrule
\end{tabular}}
\begin{tablenotes}
\item \textbf{F}: Finger bones; \textbf{C}: Carpal and metacarpal bones; \textbf{UR}: Radius and ulna.
\item \textbf{BAA}: Bone Age Assessment; $\triangle$: Partially available annotation.
\end{tablenotes}
\end{threeparttable}
\end{table*}

This lack of comprehensive data support further compounds the inherent difficulty of RA radiograph analysis across multiple levels~\cite{ejbjerg2004magnetic}. 
Hand bone segmentation requires distinguishing numerous small and overlapping anatomical structures in high-resolution radiographs. Building upon this, BE segmentation introduces additional complexity, as erosive lesions are often tiny, low-contrast, and easily confounded with normal anatomical variations, imaging noise, or projection artifacts~\cite{sharp1985many, wakefield2000value, schett2012bone, dohn2008detection}. 
Furthermore, SvdH scoring requires translating these subtle and heterogeneous image patterns into clinically meaningful ordinal grades, where BE and JSN follow distinct visual cues but share the same anatomical basis~\cite{boini2001radiographic}. In the absence of datasets that provide annotations for these interconnected tasks, models are forced to implicitly infer missing structural or pathological context, thereby amplifying error propagation across stages~\cite{pandit2020machine}. As a result, reliable RA assessment from hand radiographs remains challenging, particularly when attempting to bridge structure-aware modeling, lesion-level quantification, and clinically standardized scoring within a common framework.

\begin{figure}[!t]
    \centering
    \includegraphics[width=\textwidth]{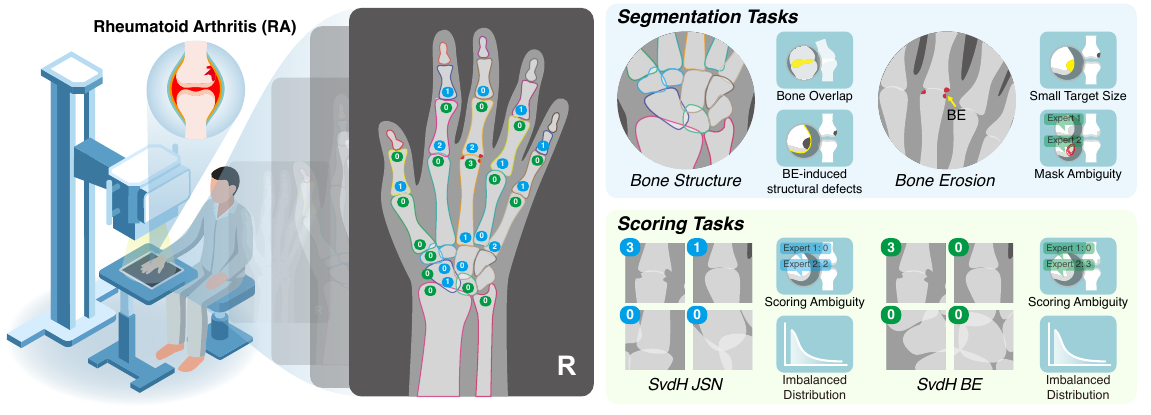}
    \caption{Overview of the proposed RAM-H1200 dataset. The dataset supports three hierarchically organized tasks: (i) hand bone structure modeling via instance-level segmentation, (ii) quantitative bone erosion (BE) analysis with pixel-level annotations, and (iii) standardized SvdH-based scoring for BE and joint space narrowing (JSN).}
    \label{fig:overview}
\end{figure}

In this paper, we introduce \textbf{R}heumatoid \textbf{A}rthritis \textbf{M}odeling-\textbf{H}and 1200 (RAM-H1200), a multi-center evaluation and dataset for comprehensive RA analysis in hand radiographs, as illustrated in Fig.~\ref{fig:overview}. RAM-H1200 contains 1,200 high-resolution hand radiographs collected from six medical centersin the Sapporo region of Japan, forming a regional, multi-institutional Japanese cohort with variability in imaging conditions. The dataset provides multi-level annotations, including instance-level segmentation masks for hand bone structures, pixel-level BE masks, and joint-level SvdH scores for both BE and JSN. This design enables systematic evaluation of RA radiograph analysis across anatomical, pathological, and clinical levels~\cite{smolen2015rheumatoid}.
\begin{itemize}
    \item \textbf{A comprehensive multi-task evaluation benchmark and dataset for RA analysis}: 
    RAM-H1200 establishes the first evaluation and dataset that systematically supports three hierarchically organized aspects of RA analysis: 
    (i) \textbf{quantitative BE analysis}, representing the first attempt to explicitly model and evaluate bone erosion in a quantitative manner rather than through coarse grading; 
    (ii) \textbf{hand bone structure modeling} via instance-level segmentation of the entire hand, providing the first systematic investigation of instance segmentation for all hand bone structures;
    and 
    (iii) \textbf{standardized SvdH-based scoring} for both BE and JSN, forming a publicly available dataset for clinically grounded scoring.
    This benchmark supports evaluation across lesion-level quantification, structure-aware modeling, and clinical assessment.
    
    \item \textbf{Fine-grained and anatomically consistent annotations}: We provide high-quality pixel-level annotations for both anatomical structures and erosion regions across the hand, together with joint-level SvdH scores. The annotations are carefully designed to preserve anatomical consistency and support both local lesion analysis and global structural modeling.
    \item \textbf{A multi-task benchmark for evaluating RA radiograph analysis}:
    We construct a multi-task benchmark covering segmentation and joint-level scoring, evaluating distinct capabilities from anatomical structure modeling to lesion-level BE quantification and joint-level SvdH scoring.
    For each sub-task in the pipeline, we establish standardized benchmarks with standardized evaluation metrics, enabling systematic evaluation and consistent cross-model comparison of intermediate task models.
\end{itemize}

\section{Overview of Dataset}
\paragraph{Ethical Considerations}
\label{sec:irb}
RAM-H1200 dataset is in compliance with the guidelines of the Declaration of Helsinki and obtained approval from the Ethics Committee of Hokkaido University (approval number: 24-104) and Institute of Science Tokyo (approval number: A24672). All radiographs included in this dataset were collected with informed consent for research use and public release.

\subsection{Image and Annotation}
\label{sec:image_and_annotation}
The dataset consists of 1,200 hand posteroanterior projection (PA) radiographs from 241 patients with RA and 50 non-RA patients. The images were collected from six institutions in the Sapporo region of Japan: Hokkaido Medical Center for Rheumatic Diseases (HMCRD), Sapporo City General Hospital (SCGH), Sagawa Akira Rheumatology Clinic (SARC), and three affiliated sites of Hokkaido University (HU1, HU2, and HU3). Each institution used its own CR system, and all data were managed in digital imaging and communications in medicine (DICOM) format. Imaging parameters are provided in Table~\ref{tab:medical_info_AP} of Appendix~\ref{sec:acquisition}. As end-stage cases exhibiting bony ankylosis were excluded during cohort construction, RAM-H1200 is intended primarily for fine-grained assessment of early-to-moderate RA structural changes rather than for evaluation of end-stage ankylosis.

We applied standardized preprocessing procedures to ensure consistency across the dataset. All radiographs were resampled to a uniform spatial resolution of 0.175 mm/pixel. Furthermore, left-hand images were horizontally flipped to match right-hand orientation, providing a unified anatomical coordinate system for subsequent analysis.
Annotation was performed by a dedicated team consisting of three radiological technologists and two clinically experienced experts, including a board-certified radiologist with 26 years of experience and an orthopedic doctor with 7 years of clinical practice. This multidisciplinary expertise ensured that the annotations were both medically accurate and clinically relevant. For segmentation tasks, initial contours were delineated by the radiological technologists. For score classification tasks, three radiological technologists and the orthopedic doctor participated in labeling. 
Each image was independently annotated by at least two annotators. The initial annotations were retained separately before consensus to enable assessment of inter-annotator variability. Discrepant annotations were subsequently reviewed jointly by the participating annotators and resolved through discussion-based consensus. The resulting consensus annotations were finally reviewed and
verified by the senior board-certified radiologist before inclusion in the released dataset. Further details of the annotation and disagreement-resolution protocol are provided in Appendix~\ref{sec:annotation_protocol_AP}.
Based on this protocol, the annotation comprised three principal components:

\begin{itemize}
    \item \textbf{Bone Structure Annotation}: 
    Precise contour delineation was performed for 30 anatomically defined structures spanning the entire hand, including the first to fifth proximal phalanges (PP1--5), middle phalanges (MP2--5), distal phalanges (DP1--5), metacarpals (MC1--5), and sesamoid bones (Ses). Carpal bones were annotated individually, including the trapezium (Tm), trapezoid (Td), scaphoid (Sca), lunate (Lu), capitate (Cap), hamate (Ham), and pisiform and triquetrum (Pis \& Tri). In addition, the distal radius (Radius) and distal ulna (Ulna) were included. Surrounding soft tissue regions were also delineated to provide additional anatomical context for structure-aware analysis. A multi-label annotation strategy was implemented to independently mark each structure.
    \item \textbf{BE Annotation}: 
    Pixel-level annotations were performed for BE regions across the hand. Following SvdH scoring principles, lesions were grouped into three categories by morphology and diagnostic certainty: high-confidence SvdH BE, moderate-confidence SvdH BE, and non-SvdH BE, referring to erosive patterns that are considered true bone erosion but do not fully meet the standard SvdH criteria.
    \item \textbf{SvdH-defined Joint ROIs for BE / JSN Scoring}: Anatomical locations for both BE and JSN were defined at the joint level following the SvdH scoring protocol, comprising 16 joints for BE assessment and 15 joints for JSN assessment. We performed ROI annotations on these areas.
    \item \textbf{SvdH BE / JSN Scoring Annotation}:
    Based on the predefined joint locations, SvdH scores for both BE and JSN were assigned at the joint level following the SvdH scoring protocol. BE scoring uses the label set $\{0,1,2,3,5\}$, where grade 5 denotes complete joint destruction. JSN scoring uses the label set $\{0,1,2,3,4\}$. Each joint was independently evaluated to quantify the severity of structural damage.
\end{itemize}

\begin{figure}[!t]
    \centering
    \includegraphics[width=\textwidth]{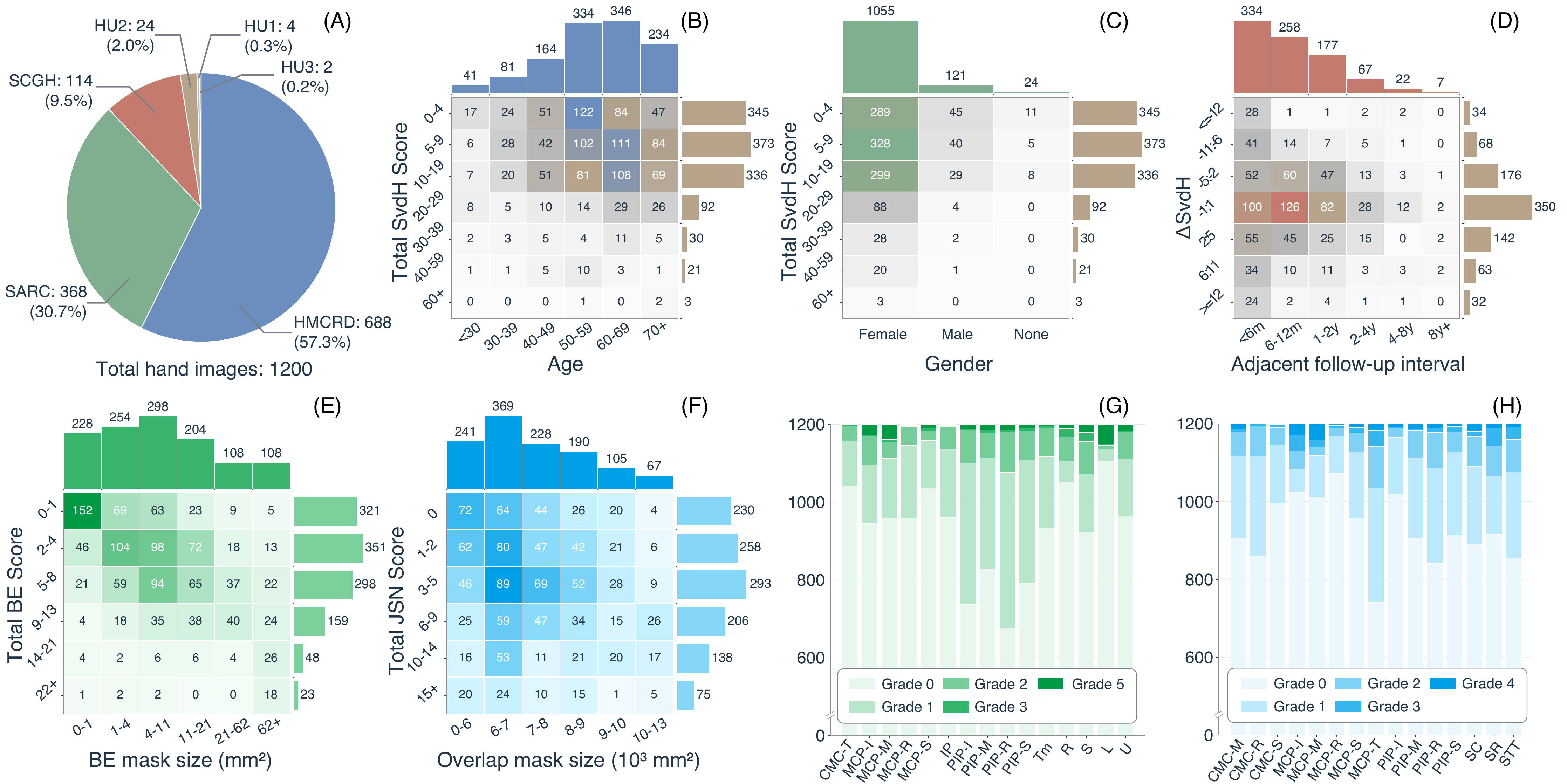}
    \caption{Distribution and Statistics for the RAM-H1200 dataset. (A) Data distribution by center. (B) Age vs. total SvdH score. (C) Gender vs. total SvdH score. (D) Adjacent follow-up interval vs. radiographic progression ($\Delta$SvdH). (E) BE mask size vs. total BE score (correlation: Spearman’s $\rho$ = 0.532, p < 0.001). (F) Overlap mask size vs. total JSN score (correlation: Spearman’s $\rho$ = 0.046, p = 0.109). (G) Joint-level BE score burden. (H) Joint-level JSN score burden. All analyses were performed at the hand-image level, treating left and right hands as independent samples.}
    \label{fig:statistics_overview}
\end{figure}

In addition, common external objects and imaging artifacts, such as rings, intravenous lines, and metallic implants, were annotated with pixel-level masks to reflect real-world clinical variability. These masks were excluded from the benchmark tasks and provided only for optional robust model development. Further details of data division and annotation are provided in Appendix~\ref{sec:detail_info}.

\subsection{Statistics of RAM-H1200}

We present statistical analyses of the RAM-H1200 dataset to characterize data sources, patient demographics, disease severity, longitudinal progression, and lesion distributions. Key attributes, including institutional distribution, age and gender, total SvdH scores, follow-up intervals, and lesion-level measurements, are summarized in Fig.~\ref{fig:statistics_overview}.
Fig.~\ref{fig:statistics_overview} (A) shows the institutional distribution of the dataset, which comprises 1,200 hand radiographs collected from six medical centers. 
Fig.~\ref{fig:statistics_overview} (B) and (C) illustrate the demographic and severity distributions. The cohort spans a wide age range, with most samples concentrated between 40 and 70 years old. 
The higher proportion of female patients is consistent with the known female predominance of RA, although this demographic imbalance should be considered when interpreting benchmark performance and generalization.
Total SvdH scores are dominated by low-to-moderate ranges, while high-severity cases are relatively limited.
Fig.~\ref{fig:statistics_overview} (D) shows the relationship between adjacent follow-up intervals and radiographic progression ($\Delta$SvdH). Most intervals fall within 1–4 years, and progression values are concentrated around small changes, indicating generally slow and incremental disease progression.
Fig.~\ref{fig:statistics_overview} (E) and (F) present lesion-level statistics. Larger BE mask sizes are associated with higher total BE scores, whereas hand-level bone-projection overlap area shows only a weak and non-significant correlation with total JSN score. We further assessed initial inter-annotator reliability before consensus discussion using the intraclass correlation coefficient (ICC), obtaining ICC(1,1) values of 0.5682 for BE, 0.4502 for JSN, and 0.5176 for the combined BE and JSN scores, broadly consistent with prior RA radiographic scoring studies~\cite{fujimori2018composite}. Discrepant cases were then reviewed and resolved through consensus discussion to produce the final annotations, with detailed analysis provided in Table~\ref{tab:icc_reliability_AP} in Appendix~\ref{app:icc_analysis_AP}.
Fig.~\ref{fig:statistics_overview} (G) and (H) show joint-level score distributions. Most joints are assigned grade 0, while higher grades appear only in a small subset, resulting in a highly imbalanced distribution. Such skewed distributions are commonly reported in clinical cohorts~\cite{bruynesteyn2002determination,jansen2001predictors}. This pattern is consistent with prior observations that advances in medical care and early intervention have reduced the prevalence of late-stage RA, making high severity scores increasingly rare in modern cohorts~\cite{yang2025ram}.

\section{Experiments and Benchmarks}
\label{sec:experiment}
\paragraph{Experimental Setup}
\label{sec:implement}
All experiments were conducted using patient-level train/validation/test splits to prevent data leakage across subsets. All models were trained using the AdamW optimizer with cosine annealing learning rate scheduling. Unless otherwise specified, the initial learning rate was set to $1\times10^{-4}$. Standard data augmentation techniques were applied during training. All experiments were conducted with a fixed random seed (2026) to ensure reproducibility. Training was performed on a server with four NVIDIA A100 40GB GPUs, while inference was carried out on an NVIDIA Quadro RTX 8000 GPU with 48GB memory. 
For segmentation tasks, we report Dice similarity coefficient (DSC), normalized surface Dice (NSD), volumetric overlap error (VOE), mean surface distance (MSD), recall (REC), and precision (PREC). For SvdH score classification, we report quadratic weighted kappa (QWK), mean absolute error (MAE), balanced accuracy (BACC), accuracy (ACC), within-one accuracy (W1-ACC), positive/negative sensitivity (P/N-SEN), and positive/negative accuracy (P/N-ACC). Task-specific implementation details are provided in Appendix~\ref{sec:implement_AP}.

\begin{table*}[!t]
\centering
\caption{Hand bone structure segmentation results obtained on the Test set. The best results in each column are highlighted in \textbf{bold}, and the second-best values are \underline{underlined}.}
\begin{threeparttable}
\resizebox{\textwidth}{!}{
\setlength{\tabcolsep}{2pt}
\begin{tabular}{lcccccccc}
\toprule
\textbf{Model} 
& \textbf{DSC (\%)} 
& \textbf{NSD (\%)} 
& \textbf{DSC$_O$ (\%)} 
& \textbf{NSD$_O$ (\%)} 
& \textbf{VOE (\%)} 
& \textbf{MSD (pix)} 
& \textbf{\#P (M)}
& \textbf{Time (ms)} \\
\midrule
\multicolumn{9}{c}{\textbf{Supervised Models}} \\
\midrule

Unet~\cite{ronneberger2015u} 
& 96.37$\pm$2.68 
& 90.41$\pm$5.82 
& 73.71$\pm$7.77 
& 73.12$\pm$10.78 
& 6.44$\pm$3.74 
& 2.85$\pm$2.98 
& 7.94 
& 325.21 \\

Unet++~\cite{zhou2018unet++} 
& 97.21$\pm$1.36 
& 92.54$\pm$4.05 
& 75.37$\pm$6.80 
& 75.66$\pm$10.04 
& 5.13$\pm$2.11 
& 2.85$\pm$3.79 
& 2.41 
& 842.49 \\

SegFormer~\cite{xie2021segformer}
& 96.65$\pm$1.48 
& 89.85$\pm$4.40
& 71.59$\pm$6.63 
& 69.39$\pm$9.70 
& 6.13$\pm$2.10
& 2.51$\pm$1.64 
& {21.88} 
& {272.00} \\

TransUnet~\cite{chen2021transunet} 
& 97.22$\pm$1.15 
& 92.87$\pm$3.88 
& 76.12$\pm$6.41 
& 76.59$\pm$10.08 
& 5.10$\pm$1.85 
& 2.10$\pm$1.15 
& 105.92 
& 893.21 \\

SwinUNETR~\cite{hatamizadeh2021swin} 
& \textbf{97.32$\pm$1.18}
& 93.07$\pm$3.96 
& \underline{76.27$\pm$7.07} 
& 77.04$\pm$10.49 
& 4.93$\pm$1.90
& \textbf{1.67$\pm$1.11}
& 25.14 
& 748.16 \\

UMambaEnc~\cite{ma2024u} 
& \underline{97.32$\pm$1.32}
& \underline{93.21$\pm$4.08} 
& 76.19$\pm$6.80 
& \underline{77.06$\pm$10.22} 
& \underline{4.93$\pm$2.08}
& \underline{1.90$\pm$3.26} 
& 4.59 
& 783.94 \\

SwinUMamba~\cite{liu2024swin} 
& 97.31$\pm$1.23 
& \textbf{93.44$\pm$3.86} 
& \textbf{76.32$\pm$6.83} 
& \textbf{77.68$\pm$10.09} 
& \textbf{4.91$\pm$1.93}
& 1.91$\pm$1.31 
& 59.89 
& 1261.96 \\

MambaVision~\cite{hatamizadeh2025mambavision} 
& 96.41$\pm$1.35 
& 87.10$\pm$4.48
& 66.43$\pm$5.93 
& 60.87$\pm$8.97 
& 6.61$\pm$2.02 
& 2.16$\pm$1.33 
& 62.43 
& 1140.82 \\

\midrule
\multicolumn{9}{c}{\textbf{Foundation Models}} \\
\midrule

SAM(Box)~\cite{kirillov2023segment} 
& 90.76$\pm$2.51
& 78.01$\pm$4.72
& 5.90$\pm$3.37 
& 4.24$\pm$3.03 
& 13.93$\pm$3.07
& 4.05$\pm$1.22 
& 641.09 
& 1478.79 \\

SAM(Point)~\cite{kirillov2023segment} 
& 75.45$\pm$9.21
& 59.18$\pm$8.81
& 3.44$\pm$1.72 
& 2.89$\pm$2.05 
& 34.34$\pm$9.91
& 39.68$\pm$29.82 
& 641.09 
& 1414.99 \\

MedSAM(Box)~\cite{MedSAM} 
& 80.61$\pm$4.42
& 33.52$\pm$8.37
& 10.81$\pm$4.74 
& 7.79$\pm$3.37 
& 30.95$\pm$5.99
& 10.33$\pm$2.10 
& 93.74
& 839.35 \\

\bottomrule
\end{tabular}
}
\raggedright
\end{threeparttable}
\label{tab:seg_results}
\end{table*}

\begin{figure}[!t]
    \centering
    \includegraphics[width=\textwidth]{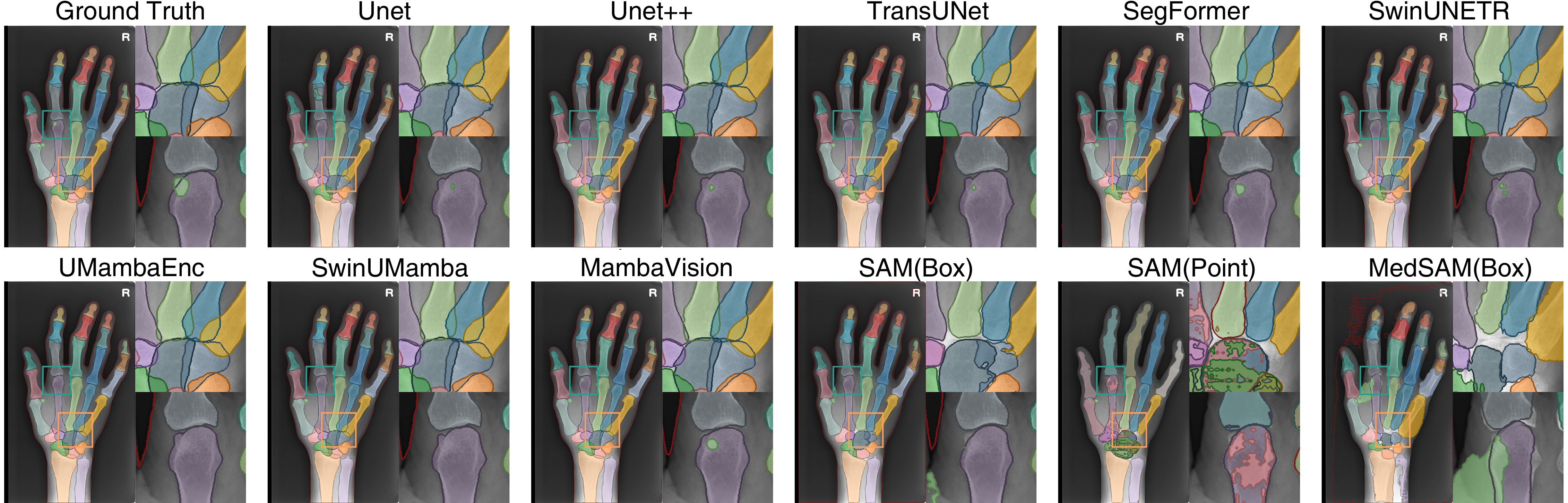}
    \caption{Hand bone structure segmentation visualization results.}
    \label{fig:bone_seg_results}
\end{figure}

\subsection{Hand Bone Structure Segmentation}

To better assess performance in anatomically overlapping regions, we additionally report overlap-aware metrics (DSC$_O$ and NSD$_O$) following~\cite{yang2025ram, yang2025ap}, where evaluation is restricted to projection-induced bone overlap regions. As shown in Table~\ref{tab:seg_results}, supervised models achieve consistently strong performance on hand bone structure segmentation, with DSC exceeding 96\% for most architectures. Among them, SwinUMamba achieves the best overall results, obtaining the highest NSD (93.44\%), DSC$_O$ (76.32\%), and NSD$_O$ (77.68\%), while SwinUNETR and UMambaEnc remain competitive across both region-based and boundary-aware metrics. For broader context, the overall DSC achieved by these supervised models falls within the high-performance range reported for comparatively well-delineated anatomical structures in established medical segmentation benchmarks; for example, large-organ tasks such as spleen segmentation in the Medical Segmentation Decathlon achieve DSC above 0.9 for many methods~\cite{antonelli2022medical}. However, despite similarly high overall DSC, clearer differences emerge in the overlap-aware metrics, with the best DSC$_O$ reaching only 76.32\%. This contrast indicates that the remaining difficulty is concentrated in anatomically complex overlap regions rather than global bone localization, where projection-induced boundary ambiguity makes precise separation of adjacent bones challenging. The degradation in overlap regions is more pronounced for MambaVision and SegFormer. In contrast, foundation models such as SAM and MedSAM perform substantially worse across all metrics, suggesting that prompt-based segmentation alone is insufficient for accurate anatomical delineation in this setting.
Qualitative results in Fig.~\ref{fig:bone_seg_results} are consistent with these observations. While most supervised models produce anatomically coherent segmentations in non-overlapping regions, errors concentrate around bone junctions and overlap areas. Models with stronger overlap-aware performance, such as SwinUMamba and SwinUNETR, better preserve boundary continuity, whereas weaker models exhibit boundary inconsistency and missing fine structures. Foundation models show more severe failures, including coarse and fragmented masks. Detailed bone-wise results, overlap-region analyses, qualitative examples, and statistical evaluations are provided in Appendix~\ref{sec:bone_seg_result_AP}.

\begin{table*}[!t]
\centering
\caption{Hand BE segmentation results obtained on the Test set. The best results in each column are highlighted in \textbf{bold}, and the second-best values are \underline{underlined}.}
\begin{threeparttable}
\resizebox{\textwidth}{!}{
\setlength{\tabcolsep}{2pt}
\begin{tabular}{lcccccccc}
\toprule
\textbf{Model} 
& \textbf{DSC (\%)} 
& \textbf{NSD (\%)} 
& \textbf{REC (\%)} 
& \textbf{PREC (\%)} 
& \textbf{VOE (\%)} 
& \textbf{MSD (pix)} 
& \textbf{\#P (M)}
& \textbf{Time (ms)} \\
\midrule

Unet~\cite{ronneberger2015u} 
& 17.25$\pm$12.09 
& 17.81$\pm$14.25 
& 23.68$\pm$23.47 
& 14.11$\pm$13.56 
& 90.07$\pm$7.45 
& \textbf{182.17$\pm$121.50} 
& 7.94 
& 365.42 \\

Unet++~\cite{zhou2018unet++} 
& 17.57$\pm$12.93 
& 17.61$\pm$14.30 
& 23.22$\pm$23.54 
& 14.37$\pm$13.94 
& 89.79$\pm$8.26 
& 196.17$\pm$116.29 
& 2.41 
& 772.82 \\

nnUnet~\cite{isensee2021nnu} 
& 12.64$\pm$17.80 
& 14.11$\pm$20.02 
& 8.79$\pm$15.74 
& \textbf{23.50$\pm$31.57} 
& 92.07$\pm$12.68 
& 212.00$\pm$143.98 
& 160.14
& 274.58 \\

TransUnet~\cite{chen2021transunet} 
& \underline{18.80$\pm$13.44} 
& 18.51$\pm$14.81 
& \underline{25.10$\pm$23.24} 
& 14.92$\pm$14.97 
& \underline{88.98$\pm$8.72} 
& 190.06$\pm$124.17 
& 105.32 
& 849.10 \\

SegFormer~\cite{xie2021segformer}
& 12.58$\pm$9.37 
& 12.42$\pm$10.38
& 23.93$\pm$23.49 
& 8.84$\pm$8.99 
& 93.01$\pm$5.57
& 213.47$\pm$118.76 
& 21.88 
& 272.00 \\

SwinUNETR~\cite{hatamizadeh2021swin} 
& 15.57$\pm$10.77 
& 15.34$\pm$12.46 
& 24.97$\pm$23.89 
& 11.69$\pm$11.48 
& 91.19$\pm$6.45 
& 211.72$\pm$118.84 
& 25.14 
& 826.32 \\

UMambaEnc~\cite{ma2024u} 
& 18.76$\pm$13.54 
& \underline{18.76$\pm$14.93} 
& 24.23$\pm$23.66 
& 15.53$\pm$14.72 
& 89.01$\pm$8.67 
& 196.61$\pm$111.87 
& 4.58 
& 780.26 \\

SwinUMamba~\cite{liu2024swin} 
& \textbf{19.59$\pm$13.40} 
& \textbf{19.93$\pm$15.28} 
& \textbf{25.30$\pm$23.95} 
& \underline{16.23$\pm$15.24} 
& \textbf{88.50$\pm$8.73} 
& \underline{183.68$\pm$112.02} 
& 59.89 
& 1361.45 \\

\bottomrule
\end{tabular}
}
\raggedright
\end{threeparttable}
\label{tab:beseg_results}
\end{table*}

\begin{figure}[!t]
    \centering
    \includegraphics[width=\textwidth]{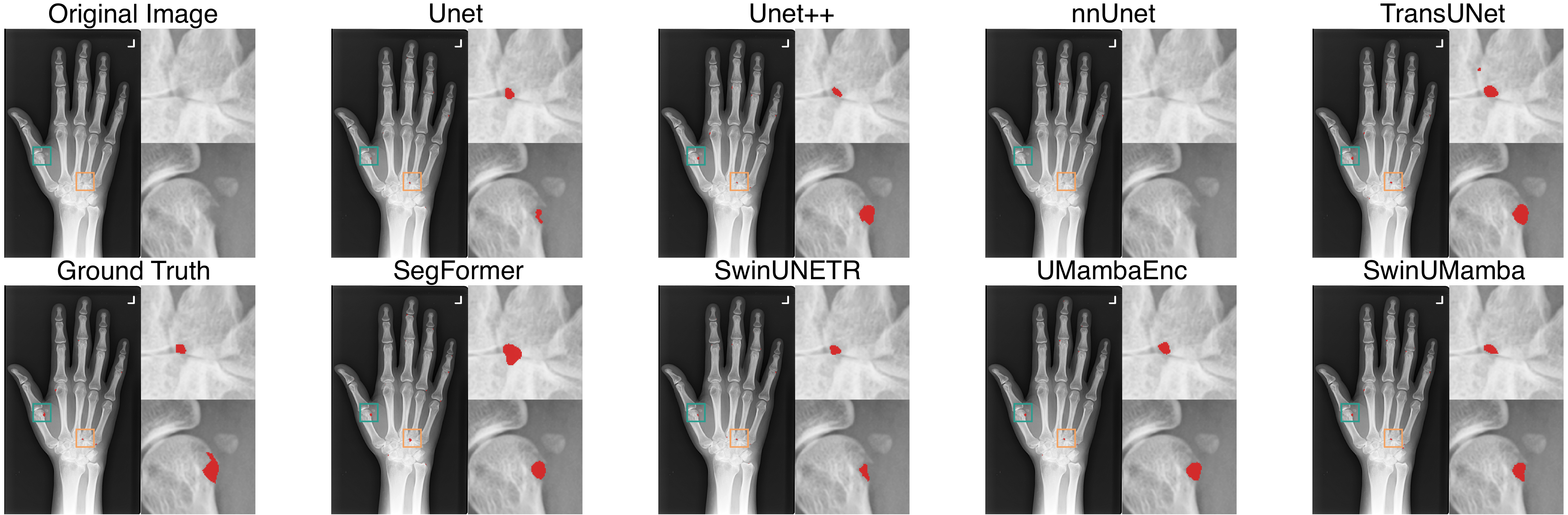}
    \caption{Hand BE segmentation visualization results.}
    \label{fig:be_seg_results}
\end{figure}

\subsection{Hand BE Segmentation}

As shown in Table~\ref{tab:beseg_results}, BE segmentation remains a highly challenging task, with all models achieving relatively low DSC scores (below 20\%). Among the evaluated methods, SwinUMamba achieves the best overall performance, obtaining the highest DSC (19.59\%), NSD (19.93\%), and REC (25.30\%), while TransUnet and UMambaEnc also show competitive results with comparable DSC and recall values. For broader context, similarly low DSC values have been reported for difficult lesion-segmentation tasks in established medical imaging benchmarks. In the Medical Segmentation Decathlon, the colon tumor task yielded a team-average median DSC as low as 0.16~\cite{antonelli2022medical}. On the MSD pancreas-tumor task, generic segmentation architectures have been reported in an approximately 0.19--0.30 DSC range, including UNETR (0.19), 3D U-Net (0.21), SegResNet (0.26), nnFormer (0.27), SwinUNETR (0.28), and Attention U-Net (0.30)~\cite{lu2025mdmu}. These results provide a numerical reference for contextualizing the low DSC observed in RAM-H1200, although direct comparison is limited by differences in imaging modality, anatomy, lesion characteristics, and evaluation settings.
Within RAM-H1200, performance remains consistently low across architectures, with a relatively limited gap between models compared with bone structure segmentation. Notably, nnU-Net achieves the highest precision (23.50\%) but substantially lower recall (8.79\%), indicating conservative predictions, whereas most other models achieve relatively higher recall at the cost of lower precision, reflecting more frequent false-positive predictions in uncertain regions. These results highlight a substantial trade-off between sensitivity and false positives in BE segmentation, with the evaluated architectures struggling to simultaneously achieve accurate lesion localization and reliable boundary delineation.

Qualitative analysis in Fig.~\ref{fig:be_seg_results} reveals three common failure patterns: (i) missed detections of small or low-contrast erosions, (ii) false positives around visually ambiguous cortical regions, and (iii) coarse or fragmented masks with imprecise boundaries. These errors reflect the difficulty of BE segmentation, where lesions are tiny and sparse, boundaries are often unclear, and appearances vary with projection and local bone morphology. Annotation can also be subjective because lesion extent is not always sharply defined. Stronger models such as SwinUMamba and TransUnet show better sensitivity to small lesions, but still suffer from boundary imprecision and occasional over-segmentation, while weaker models such as SegFormer and nnUnet often miss lesions or produce unstable fragments. Overall, even the best models struggle to achieve both accurate localization and clean boundary delineation. 
Additional multi-class BE segmentation results, independent-reader re-segmentation, joint- and lesion-level detection analyses, and statistical evaluations are provided in Appendix~\ref{sec:be_seg_result_AP}.

\begin{table*}[!t]
\centering
\caption{SvdH BE score classification results obtained on the Test set. The best results in each column are highlighted in \textbf{bold}, and the second-best values are \underline{underlined}.}
\label{tab:be_results}
\begin{threeparttable}
\resizebox{\textwidth}{!}{
\setlength{\tabcolsep}{2pt}
\begin{tabular}{lccccccccc}
\toprule
\textbf{Model} 
& \textbf{QWK} 
& \textbf{MAE} 
& \textbf{BACC (\%)} 
& \textbf{ACC (\%)} 
& \textbf{W1-ACC (\%)} 
& \textbf{P/N-SEN (\%)}
& \textbf{P/N-ACC (\%)} 
& \textbf{\#P (M)}
& \textbf{Time (ms)} \\
\midrule

ResNet~\cite{he2016deep} 
& \underline{0.4408} 
& 0.3904 
& \textbf{35.87} 
& 69.71 
& 92.81 
& \underline{37.00}
& 73.78 
& 21.28 
& 0.25 \\

DenseNet~\cite{huang2017densely} 
& 0.3905 
& \textbf{0.3701} 
& 33.06 
& \textbf{73.08} 
& 91.60 
& 15.77
& \underline{75.12} 
& 6.95 
& 0.61 \\

EfficientNetV2~\cite{tan2021efficientnetv2} 
& 0.3358 
& 0.3937 
& 30.67 
& 71.30 
& 91.71 
& 23.48
& 74.11 
& 20.18 
& 0.63 \\

MobileViT~\cite{mehta2021mobilevit} 
& 0.3920 
& 0.4050 
& 31.88 
& 69.41 
& \textbf{92.25} 
& 35.53
& 73.50 
& 4.94 
& 0.56 \\

LeViT~\cite{graham2021levit} 
& 0.2346 
& 0.4239 
& 25.66 
& 70.25 
& 90.94 
& 23.22
& 72.92 
& 7.01 
& 2.65 \\

EfficientFormer~\cite{li2022efficientformer} 
& 0.3504 
& 0.3951 
& 27.32 
& 70.46 
& \underline{92.13} 
& 25.22
& 73.97 
& 3.25 
& 0.54 \\

ConvNeXtV2~\cite{woo2023convnext} 
& 0.3058 
& 0.4127 
& 28.15 
& 70.04 
& 91.39 
& 22.62
& 72.87 
& 27.87 
& 1.13 \\

MedMamba~\cite{yue2024medmamba} 
& \textbf{0.4522} 
& 0.3961 
& \underline{34.91} 
& 70.13 
& 92.11 
& \textbf{38.73}
& 74.72 
& 14.45 
& 1.39 \\

MambaVision~\cite{hatamizadeh2025mambavision} 
& 0.3667 
& \underline{0.3804} 
& 30.59 
& \underline{72.85} 
& 91.15 
& 17.94
& \textbf{75.14} 
& 31.16 
& 0.73 \\

\bottomrule
\end{tabular}
}
\raggedright
\end{threeparttable}
\end{table*}

\begin{figure}[!t]
    \centering
    \includegraphics[width=\textwidth]{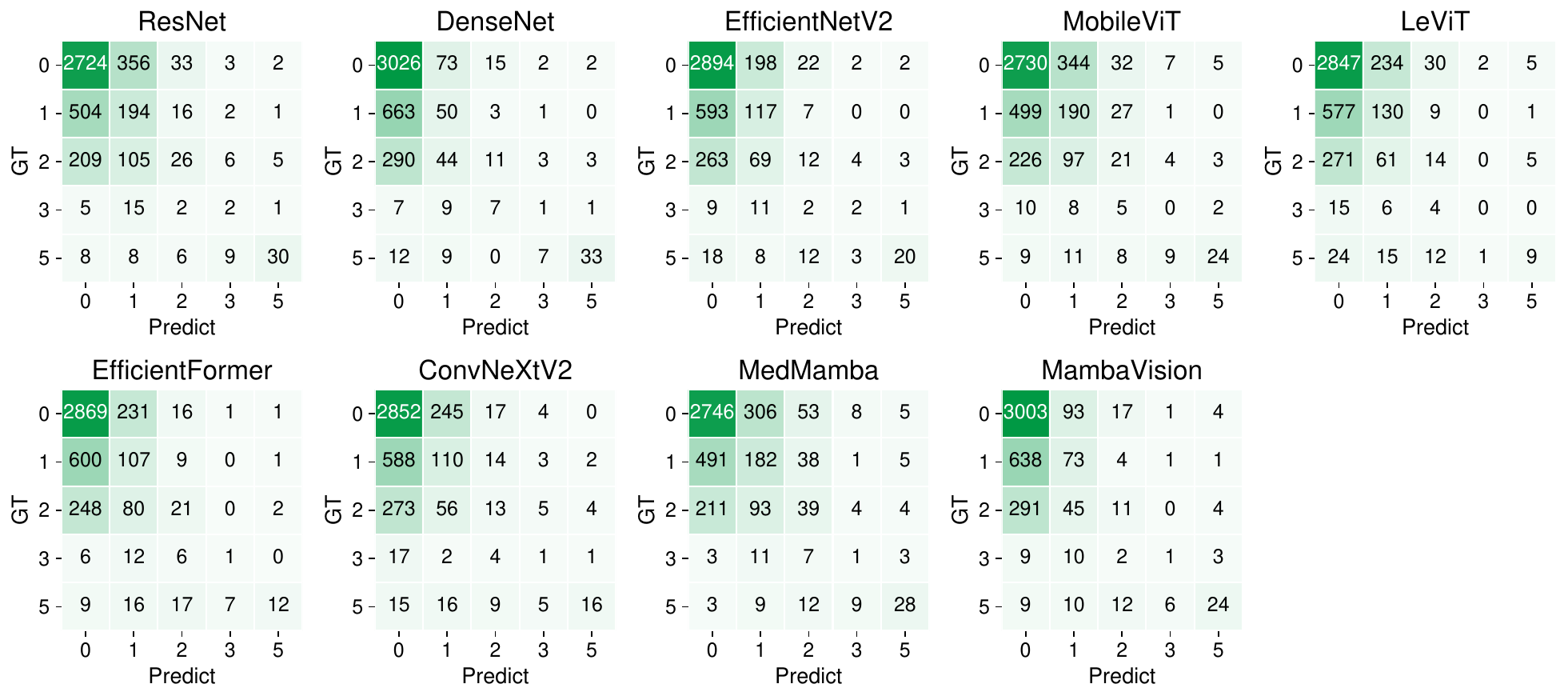}
    \caption{Confusion matrices for SvdH BE scoring across models.}
    \label{fig:be_score_results}
\end{figure}

\subsection{Scoring of SvdH BE}
As shown in Table~\ref{tab:be_results}, SvdH BE score classification remains challenging, with all models showing limited ordinal agreement and balanced accuracy. MedMamba achieves the best overall agreement with the ground truth, obtaining the highest QWK (0.4522) and the strongest positive sensitivity (38.73\%), while ResNet shows competitive performance with the best balanced accuracy (35.87\%) and high within-one accuracy. DenseNet obtains the lowest MAE (0.3701) and the highest ACC (73.08\%), but its relatively low positive sensitivity indicates a tendency to favor negative or low-score predictions. This discrepancy suggests that ACC alone can be misleading for BE scoring, since the label distribution is highly skewed toward low grades.
The confusion matrices in Fig.~\ref{fig:be_score_results} further show that predictions are concentrated at lower BE scores, with only partial diagonal alignment across severity levels. Higher-grade erosions are rarely predicted correctly, and many positive cases are shifted toward lower scores, reflecting the small, sparse, and visually ambiguous nature of BE lesions and the subjective boundaries between adjacent grades. Thus, beyond detecting erosion presence, estimating its progression level remains difficult. Overall, although several models achieve reasonable ACC and W1-ACC, their inconsistent BACC and P/N-SEN indicate that reliable BE scoring still requires better handling of subtle lesion patterns, ordinal severity, and severe class imbalance. Detailed joint-wise results, confusion-matrix analyses, and statistical evaluations are provided in Appendix~\ref{sec:be_score_result_AP}. An end-to-end evaluation using automatically detected joint ROIs is presented in Appendix~\ref{sec:auto_roi_be_AP}.

\begin{table*}[!t]
\centering
\caption{SvdH JSN score classification results obtained on the Test set. The best results in each column are highlighted in \textbf{bold}, and the second-best values are \underline{underlined}.}
\label{tab:jsn_results}
\begin{threeparttable}
\resizebox{\textwidth}{!}{
\setlength{\tabcolsep}{2pt}
\begin{tabular}{lccccccccc}
\toprule
\textbf{Model} 
& \textbf{QWK} 
& \textbf{MAE} 
& \textbf{BACC (\%)} 
& \textbf{ACC (\%)} 
& \textbf{W1-ACC (\%)}
& \textbf{P/N-SEN (\%)}
& \textbf{P/N-ACC (\%)}
& \textbf{\#P (M)}
& \textbf{Time (ms)} \\
\midrule

ResNet~\cite{he2016deep} 
& 0.5884 
& \underline{0.2824} 
& 39.51 
& 76.03 
& \textbf{96.33} 
& 46.32
& 80.37
& 21.28 
& 0.31 \\

DenseNet~\cite{huang2017densely} 
& 0.5829 
& 0.3031 
& 36.44 
& 74.08 
& 96.13
& \textbf{48.13}
& 78.88
& 6.95 
& 0.67 \\

EfficientNetV2~\cite{tan2021efficientnetv2} 
& 0.5393 
& 0.3011 
& 32.83 
& 75.43 
& 95.18 
& 41.45
& 80.15
& 20.18 
& 0.71 \\

MobileViT~\cite{mehta2021mobilevit} 
& \textbf{0.5967} 
& 0.2871 
& \textbf{42.80} 
& 76.01 
& 95.98 
& \underline{46.89}
& 80.25
& 4.94 
& 0.63 \\

LeViT~\cite{graham2021levit} 
& 0.5445 
& 0.3006 
& \underline{40.62} 
& 75.53 
& 95.48 
& 41.90
& 79.35
& 7.01 
& 2.65 \\

EfficientFormer~\cite{li2022efficientformer} 
& \underline{0.5919} 
& \textbf{0.2769} 
& 39.25 
& \textbf{76.75} 
& \underline{96.23} 
& 45.53
& \textbf{80.97}
& 3.25 
& 0.56 \\

ConvNeXtV2~\cite{woo2023convnext} 
& 0.5151 
& 0.2941 
& 32.51 
& \underline{76.13} 
& 95.36 
& 32.50
& 79.95
& 27.87 
& 1.16 \\

MedMamba~\cite{yue2024medmamba} 
& 0.5738 
& 0.2934 
& 38.66 
& 75.46 
& 95.86 
& 43.49
& 79.53
& 14.45 
& 1.47 \\

MambaVision~\cite{hatamizadeh2025mambavision} 
& 0.5457 
& 0.2899 
& 34.62 
& 76.01 
& 95.88
& 44.05
& \underline{80.45}
& 31.16 
& 0.80 \\

\bottomrule
\end{tabular}
}
\raggedright
\end{threeparttable}
\end{table*}

\begin{figure}[!t]
    \centering
    \includegraphics[width=\textwidth]{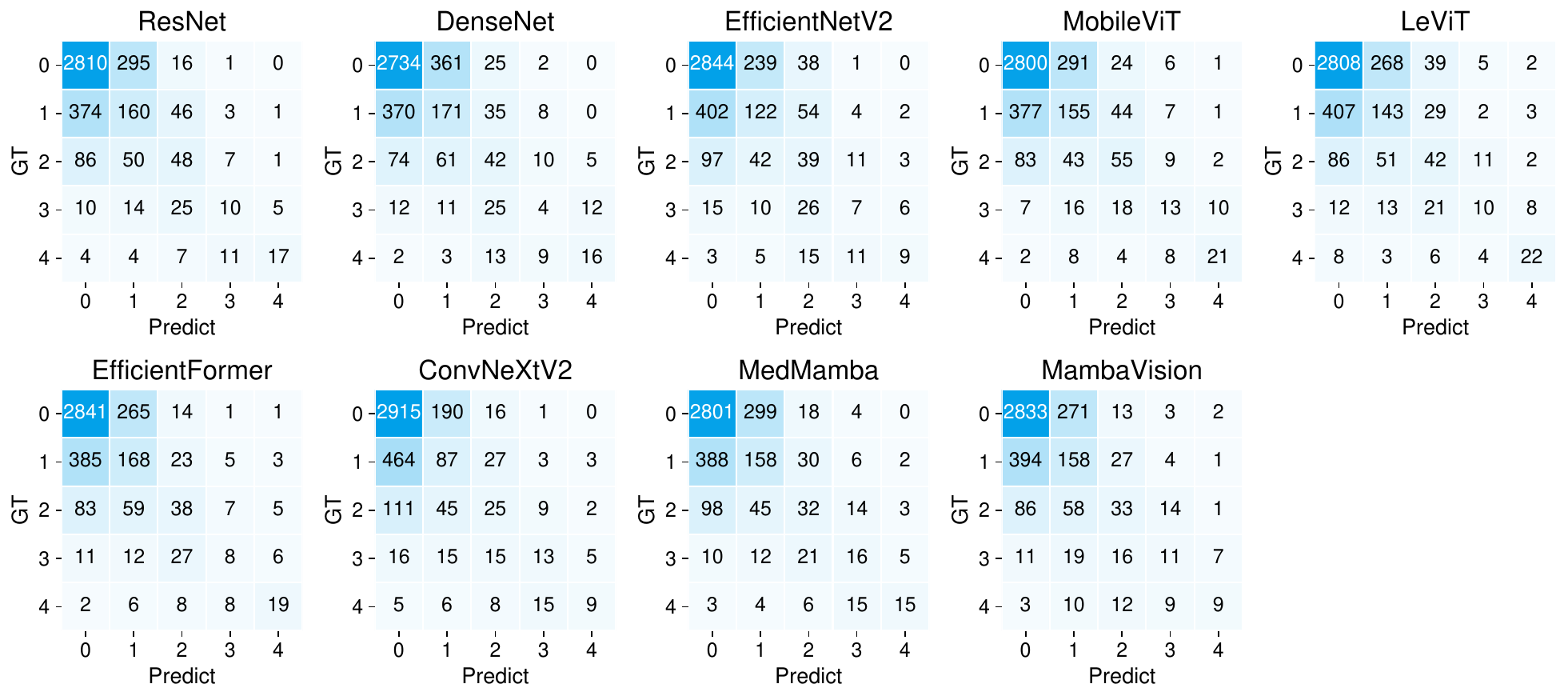}
    \caption{Confusion matrices for SvdH JSN scoring across models.}
    \label{fig:jsn_score_results}
\end{figure}

\subsection{Scoring of SvdH JSN}
As shown in Table~\ref{tab:jsn_results}, SvdH JSN score classification is relatively easier than BE scoring, but the overall performance is still not fully satisfactory. MobileViT achieves the best ordinal agreement and balanced classification performance, with the highest QWK (0.5967) and BACC (42.80\%). EfficientFormer obtains the lowest MAE (0.2769), the highest ACC (76.75\%), and the best P/N-ACC (80.97\%), while ResNet also shows strong within-one accuracy (96.33\%). These results suggest that the models can capture many coarse JSN patterns and often make predictions close to the ground truth. However, the gap between high ACC/W1-ACC and more moderate QWK/BACC indicates that correct fine-grained severity classification remains difficult. In addition, the variation in P/N-SEN across models shows that identifying positive JSN cases is still unstable under the imbalanced label distribution.
The confusion matrices in Fig.~\ref{fig:jsn_score_results} further support this observation. Most predictions are concentrated in the lower JSN scores, and many errors occur between neighboring grades rather than across distant classes. This explains the high W1-ACC, but it also shows that the models often struggle to make precise score-level decisions. Compared with BE, JSN changes are more structurally visible, so the overall agreement is better. Nevertheless, early narrowing remains difficult to grade: mild joint-space narrowing at scores 1--2 can be hard to distinguish by visual inspection, especially when the joint margins show slight irregularity. Moreover, the SvdH system requires fine-grained grading of narrowing severity from 0 to 4, which depends heavily on clinical reading experience and can introduce inter-observer variability. These factors make JSN progression difficult to model reliably, particularly for positive and higher-grade cases under class imbalance. 
Detailed joint-wise results, confusion-matrix analyses, and statistical evaluations are provided in Appendix~\ref{sec:jsn_score_result_AP}. The joint-level BE and JSN predictions are further aggregated into hand-level SvdH scores in Appendix~\ref{app:hand_level_aggregation}, while an end-to-end evaluation using automatically detected joint ROIs is presented in Appendix~\ref{sec:auto_roi_jsn_AP}.

\section{Conclusion and Limitations}
\label{sec:conclusion_limitation}
In this paper, we introduced RAM-H1200, a multi-task benchmark for RA assessment from hand radiographs, extending prior wrist-focused studies to a comprehensive whole-hand setting. The dataset covers three clinically relevant tasks and provides high-quality annotations at multiple levels, including hand bone structure segmentation, BE segmentation, and standardized SvdH-based scoring, enabling evaluation of both anatomical structure and pathological severity. Notably, this work represents one of the first attempts to explicitly formulate and benchmark quantitative BE analysis at scale, moving beyond coarse severity grading toward more fine-grained characterization of erosive progression.
Benchmark results demonstrate that, although state-of-the-art models achieve strong performance in global hand bone segmentation, their performance degrades substantially in regions affected by bone overlap. In addition, the detection and segmentation of small, sparsely distributed erosive lesions remain particularly challenging~\cite{lin2017focal}. Similarly, SvdH-based scoring continues to present significant difficulties, with overall performance still falling short of clinical applicability.
These limitations are fundamentally associated with the intrinsic characteristics of RA pathology and radiographic imaging. In bone structure segmentation, projection-induced overlap in two-dimensional radiographs obscures critical anatomical boundaries, thereby increasing task complexity. For BE segmentation, lesions are typically small, irregular, and locally ambiguous, making consistent and precise annotation difficult even for experienced clinicians. Furthermore, the SvdH scoring system inherently relies on subjective assessment and involves relatively ambiguous grading boundaries, which further constrain annotation consistency and model robustness. These challenges highlight the need for future research to incorporate structural priors, multi-scale modeling, and uncertainty-aware learning in order to improve robustness in subtle lesion scenarios while mitigating the impact of subjective annotation variability.

\section{Acknowledgments}
This work was supported by JST BOOST and JST SPRING, Japan Grant Number JPMJBS2426 and JPMJSP2180.

\bibliographystyle{plain}
\bibliography{refs}

\newpage
\appendix

\section{RAM-H1200 Data Access and Format} \label{sec:data_access}

The data can be accessed on HuggingFace at \url{https://huggingface.co/datasets/TokyoTechMagicYang/RAM-H1200-v1}. 
The benchmark and code can be accessed on GitHub at \url{https://github.com/YSongxiao/RAM-H1200}.

The dataset is organised into two main components (\texttt{Segmentation/} and \texttt{SvdH\_Scoring/}), corresponding to segmentation and SvdH scoring related tasks, respectively. 
The dataset structure is shown as follows:

\begin{lstlisting}[style=datasetstyle]
RAM-H1200-v1/
|-- Segmentation/
|   |-- train/
|   |   |-- JP_HMCRD_P0001_20210203_6791_L.bmp
|   |   |-- JP_HMCRD_P0001_20210203_6791_R.bmp
|   |   |-- ...
|   |   |-- _annotations_bone_rle.coco.json   # COCO annotations for bone segmentation
|   |   |-- _annotations_be_rle.coco.json     # COCO annotations for bone erosion segmentation
|   |-- val/
|   |   |-- JP_HMCRD_P0026_20201122_5221_L.bmp
|   |   |-- JP_HMCRD_P0026_20201122_5221_R.bmp
|   |   |-- ...
|   |   |-- _annotations_bone_rle.coco.json
|   |   |-- _annotations_be_rle.coco.json
|   |-- test/
|   |   |-- JP_HMCRD_P0007_20140629_6951_L.bmp
|   |   |-- JP_HMCRD_P0007_20140629_6951_R.bmp
|   |   |-- ...
|   |   |-- _annotations_bone_rle.coco.json
|   |   |-- _annotations_be_rle.coco.json
|-- SvdH_Scoring/
|   |-- SvdH_BE_Scoring/
|   |   |-- train/
|   |   |   |-- JP_HMCRD_P0001_20210203_6791_L/
|   |   |   |   |-- CMC-T.bmp
|   |   |   |   |-- IP.bmp
|   |   |   |   |-- L.bmp
|   |   |   |   |-- MCP-I.bmp
|   |   |   |   |-- ...
|   |   |   |-- ...
|   |   |   |-- _annotations_be_joint_detection.coco.json
|   |   |   |-- _annotation_be_scores.json
|   |   |-- val/
|   |   |   |-- ...
|   |   |   |-- _annotations_be_joint_detection.coco.json
|   |   |   |-- _annotation_be_scores.json
|   |   |-- test/
|   |   |   |-- ...
|   |   |   |-- _annotations_be_joint_detection.coco.json
|   |   |   |-- _annotation_be_scores.json
|   |-- SvdH_JSN_Scoring/
|   |   |-- train/
|   |   |   |-- JP_HMCRD_P0001_20210203_6791_L/
|   |   |   |   |-- CMC-M.bmp
|   |   |   |   |-- CMC-R.bmp
|   |   |   |   |-- CMC-S.bmp
|   |   |   |   |-- MCP-I.bmp
|   |   |   |   |-- ...
|   |   |   |-- ...
|   |   |   |-- _annotations_jsn_joint_detection.coco.json
|   |   |   |-- _annotation_jsn_scores.json
|   |   |-- val/
|   |   |   |-- ...
|   |   |   |-- _annotations_jsn_joint_detection.coco.json
|   |   |   |-- _annotation_jsn_scores.json
|   |   |-- test/
|   |   |   |-- ...
|   |   |   |-- _annotations_jsn_joint_detection.coco.json
|   |   |   |-- _annotation_jsn_scores.json
|-- Metadata.xlsx
\end{lstlisting}

\begin{itemize}
    \item \texttt{Segmentation/train/val/test/}: Contains hand radiographs in BMP format. 
    Each image file follows a naming convention similar to \texttt{JP\_[Center]\_P[PatientID]\_[StudyDate]\_[ImageID]\_[L/R].bmp}, 
    where \texttt{L} and \texttt{R} indicate the left or right hand, respectively. 
    The \texttt{StudyDate} field is de-identified via a consistent temporal offset applied per patient.

    \item \texttt{Segmentation/\_annotations\_bone\_rle.coco.json}: COCO-format annotations for hand bone structure segmentation. 
    Masks are stored using run-length encoding (RLE) in the \texttt{segmentation} field. 
    The categories correspond to anatomical structures and related objects, such as \texttt{Capitate}, \texttt{Lunate}, \texttt{Scaphoid}, \texttt{Radius}, \texttt{Ulna}, \texttt{MC1--MC5}, \texttt{PP1--PP5}, \texttt{DP1--DP5}, as well as some non-bone objects such as \texttt{Metal Implant}, \texttt{Ring}, and \texttt{SoftTissue}. 
    The format of entries in the JSON file is shown as follows:
    \begin{samepage}
    \begin{lstlisting}[style=datasetstyle]
{
  "images": [
    {
      "id": 0,
      "file_name": "JP_SCGH_P0024_20130727_1661_L.bmp",
      "height": 1431,
      "width": 893
    },
    ...
  ],
  "annotations": [
    {
      "id": 1,
      "image_id": 0,
      "category_id": 30,
      "bbox": [14.0, 198.0, 852.0, 1233.0],
      "area": 515212.0,
      "segmentation": {
        "size": [1431, 893],
        "counts": "..."
      }
    },
    ...
  ],
  "categories": [
    {
      "id": 1,
      "name": "Capitate",
      "supercategory": "bone"
    },
    {
      "id": 9,
      "name": "Lunate",
      "supercategory": "bone"
    },
    ...
  ]
}
    \end{lstlisting}
    \end{samepage}

    \item \texttt{Segmentation/\_annotations\_be\_rle.coco.json}: COCO-format annotations for bone erosion related segmentation. 
    The masks are also stored in RLE format. 
    The categories include \texttt{Non-SvdH-BE}, \texttt{SvdH-BE-50}, and \texttt{SvdH-BE-90}.

    \item \texttt{SvdH\_Scoring/SvdH\_BE\_Scoring/train/val/test/}: Each subset contains folders named by case identifiers, e.g., \texttt{JP\_HMCRD\_P0001\_20210203\_6791\_L}. 
    Inside each folder are ROI images in BMP format for bone erosion scoring. 
    A typical folder contains 16 ROI images corresponding to joints/surfaces such as \texttt{CMC-T}, \texttt{IP}, \texttt{L}, \texttt{Tm}, \texttt{R}, \texttt{U}, \texttt{MCP-T}, \texttt{MCP-I}, \texttt{MCP-M}, \texttt{MCP-R}, \texttt{MCP-S}, \texttt{PIP-I}, \texttt{PIP-M}, \texttt{PIP-R}, and \texttt{PIP-S}.

    \item \texttt{SvdH\_Scoring/SvdH\_BE\_Scoring/\_annotations\_be\_joint\_detection.coco.json}: A COCO-format JSON file containing joint detection annotations for BE-related ROIs. 
    Each image entry represents a hand radiograph, and the annotations provide bounding boxes for the corresponding joints. 
    The \texttt{categories} section maps category IDs to joint names such as \texttt{R}, \texttt{U}, \texttt{L}, \texttt{CMC-T}, \texttt{S}, \texttt{Tm}, \texttt{PIP-S}, and \texttt{MCP-T}.

    \item \texttt{SvdH\_Scoring/SvdH\_BE\_Scoring/\_annotation\_be\_scores.json}: A JSON file containing ground-truth BE scores, indexed by full image filename. 
    The format of entries in the JSON file is shown as follows:
    \begin{samepage}
    \begin{lstlisting}[style=datasetstyle]
{
  "JP_HMCRD_P0167_20111230_3497_L.bmp": {
    "BE_MCP-T": 0,
    "BE_MCP-I": 1,
    "BE_MCP-M": 0,
    "BE_MCP-R": 0,
    "BE_MCP-S": 0,
    "BE_IP": 0,
    "BE_PIP-I": 0,
    "BE_PIP-M": 0,
    "BE_PIP-R": 1,
    "BE_PIP-S": 1,
    "BE_CMC-T": 0,
    "BE_Tm": 1,
    "BE_S": 0,
    "BE_L": 0,
    "BE_U": 0,
    "BE_R": 0
  }
}
    \end{lstlisting}
    \end{samepage}

    \item \texttt{SvdH\_Scoring/SvdH\_JSN\_Scoring/train/val/test/}: Each subset contains folders named by case identifiers. 
    Inside each folder are ROI images in BMP format for joint space narrowing (JSN) scoring. 
    A typical folder contains 15 ROI images corresponding to \texttt{CMC-M}, \texttt{CMC-R}, \texttt{CMC-S}, \texttt{SC}, \texttt{SR}, \texttt{STT}, \texttt{MCP-T}, \texttt{MCP-I}, \texttt{MCP-M}, \texttt{MCP-R}, \texttt{MCP-S}, \texttt{PIP-I}, \texttt{PIP-M}, \texttt{PIP-R}, and \texttt{PIP-S}.

    \item \texttt{SvdH\_Scoring/SvdH\_JSN\_Scoring/\_annotations\_jsn\_joint\_detection.coco.json}: A COCO-format JSON file containing joint detection annotations for JSN-related ROIs. 
    The categories include carpal joints such as \texttt{CMC-M}, \texttt{CMC-R}, \texttt{CMC-S}, \texttt{SC}, \texttt{SR}, \texttt{STT}, as well as finger joints such as \texttt{MCP-T}, \texttt{MCP-I}, \texttt{MCP-M}, \texttt{MCP-R}, \texttt{MCP-S}, \texttt{PIP-I}, \texttt{PIP-M}, \texttt{PIP-R}, and \texttt{PIP-S}.

    \item \texttt{SvdH\_Scoring/SvdH\_JSN\_Scoring/\_annotation\_jsn\_scores.json}: A JSON file containing ground-truth JSN scores, indexed by full image filename. 
    The format of entries in the JSON file is shown as follows:
    \begin{samepage}
    \begin{lstlisting}[style=datasetstyle]
{
  "JP_HMCRD_P0167_20111230_3497_L.bmp": {
    "JSN_MCP-T": 2,
    "JSN_MCP-I": 0,
    "JSN_MCP-M": 0,
    "JSN_MCP-R": 0,
    "JSN_MCP-S": 0,
    "JSN_PIP-I": 0,
    "JSN_PIP-M": 0,
    "JSN_PIP-R": 0,
    "JSN_PIP-S": 0,
    "JSN_STT": 0,
    "JSN_SC": 0,
    "JSN_SR": 0,
    "JSN_CMC-M": 0,
    "JSN_CMC-R": 0,
    "JSN_CMC-S": 0
  }
}
    \end{lstlisting}
    \end{samepage}

    \item \texttt{Metadata.xlsx}: An Excel file containing study-level metadata. 
    The main columns include:
    \begin{itemize}
        \item \texttt{Mapped Image Stem}: A normalized image or study identifier.
        \item \texttt{StudyID}: A patient-specific study index indicating the chronological order of examinations for the same individual. It is used to distinguish different follow-up time points within a patient.
        \item \texttt{Normalized PatientID}: Normalized anonymized patient identifier.
        \item \texttt{isRA}: Binary indicator of rheumatoid arthritis status.
        \item \texttt{Sex}: Patient sex.
        \item \texttt{Age}: Patient age.
        \item \texttt{Center}: Source center or institution.
        \item \texttt{PixelSpacing}: In-plane image resolution.
        \item \texttt{ImageSize}: Image size in pixels.
        \item \texttt{LR}: Hand side indicator.
    \end{itemize}
\end{itemize}

\section{Related Works}
\label{sec:related}

\subsection{Hand Bone Structure Segmentation}

\begin{table*}[!h]
\centering
\caption{Summary of representative works on hand/wrist bone structure segmentation. Ann/Img: Annotations per image.}
\label{tab:hand_structure_seg}
\begin{threeparttable}
\resizebox{\textwidth}{!}{
\begin{tabular}{cccccccccc}
\toprule
\multirow{2.5}{*}{\textbf{Works}} & 
\multirow{2.5}{*}{\textbf{Year}} & 
\multirow{2.5}{*}{\textbf{Method}} & 
\multirow{2.5}{*}{\textbf{Dataset}} & 
\multirow{2.5}{*}{\textbf{\makecell{Images\\(Ann/Img)}}} & 
\multirow{2.5}{*}{\textbf{Patients}} &
\multicolumn{4}{c}{\textbf{Objects}} \\
\cmidrule(lr){7-10}
& & & & & & \textbf{F} & \textbf{MC} & \textbf{C} & \textbf{UR} \\
\midrule
Thodberg et al.~\cite{thodberg2002hands} & 2002 & AAM & Private & 99 (-) & - & \checkmark & \checkmark &  & \checkmark \\
Kauffman et al.~\cite{kauffman2004segmentation} & 2004 & AAM & Private & 50 (20) & - & \checkmark & \checkmark &  &  \\
Yang et al.~\cite{yang2021deep} & 2021 & ResNet & Private & 720 (2) & 360 &  &  &  & \checkmark \\
Kang et al.~\cite{kang2022automatic} & 2022 & CNN & Private & 702 (10) & 702 &  &  & \checkmark & \checkmark \\
Lee et al.~\cite{lee2023osteoporosis} & 2023 & SAM & Private & 192 (7) & 192 &  & \checkmark &  & \checkmark \\
Du et al.~\cite{du2024hand} & 2024 & GRU-Unet & Private & 2000 (13) & - & \checkmark & \checkmark &  & \checkmark \\
Yang et al.~\cite{yang2025ap} & 2025 & SwinUMamba & \cite{yang2025ram} & 618 (14) & 388 &  & \checkmark & \checkmark & \checkmark \\
\bottomrule
\end{tabular}
}
\begin{tablenotes}
\item \textbf{F}: Finger bones; \textbf{MC}: Metacarpals; \textbf{C}: Carpal bones; \textbf{UR}: Radius and ulna.
\item \textbf{AAM}: Active Appearance Model.
\end{tablenotes}
\end{threeparttable}
\end{table*}

As summarized in Table~\ref{tab:hand_structure_seg}, early studies on hand/wrist radiographs mainly relied on model-driven approaches such as active appearance models (AAMs), which explicitly encode shape and appearance priors for bone localization and segmentation~\cite{thodberg2002hands,kauffman2004segmentation}. However, these methods were typically evaluated on small private datasets and focused on limited anatomical structures.

Recent work has shifted toward deep learning-based segmentation. Methods based on ResNet, CNN, SAM, GRU-Unet, and SwinUMamba have been applied to segment different hand/wrist structures, including the radius/ulna, carpal bones, metacarpals, and phalanges~\cite{yang2021deep,kang2022automatic,lee2023osteoporosis,du2024hand,yang2025ap}. Although these approaches improve segmentation performance, most were developed on private datasets, and many remain restricted to partial anatomical regions, especially the wrist.

Large-scale hand/wrist radiograph datasets such as RSNA Bone Age~\cite{halabi2019rsna} and DHA~\cite{gertych2007bone} provide substantial data volume but only global labels, limiting their use for pixel-wise structural learning. Other datasets, such as GRAZPEDWRI-DX~\cite{nagy2022pediatric}, provide detection-oriented annotations rather than segmentation. More recent RA-oriented datasets, such as RAM-W600~\cite{yang2025ram}, introduce bone-level annotations but remain limited to the wrist region.

\subsection{Hand BE Segmentation}
Bone erosion (BE) in radiographs remains challenging to characterize due to its typically small lesion size, low contrast, and the difficulty of obtaining precise annotations. Existing studies have predominantly focused on quantitative analyses of BE, whereas pixel-wise segmentation has received comparatively limited attention. Most prior work emphasizes clinically relevant tasks such as detection, grading, or severity assessment, rather than dense delineation of erosion regions. Consequently, BE is commonly analyzed at the image-, joint-, or region-level, rather than being explicitly formulated as a segmentation problem. To the best of our knowledge, there remains a paucity of research directly addressing BE segmentation in radiographs, which in turn hinders the development of anatomically consistent and interpretable models for pixel-level BE assessment.
\subsection{Qualitative Evaluation of BE and JSN}
\begin{table*}[!h]
\centering
\caption{Summary of representative works related to SvdH-based joint space narrowing (JSN) and bone erosion (BE) analysis. Ann/Img: Annotations per image.}
\label{tab:svdh_be_summary}
\begin{threeparttable}
\resizebox{\textwidth}{!}{
\begin{tabular}{ccccccccc}
\toprule
\multirow{2.5}{*}{\textbf{Works}} & 
\multirow{2.5}{*}{\textbf{Year}} & 
\multirow{2.5}{*}{\textbf{Method}} & 
\multirow{2.5}{*}{\textbf{Dataset}} & 
\multirow{2.5}{*}{\textbf{\makecell{Images\\(Ann/Img)}}} & 
\multirow{2.5}{*}{\textbf{Patients}} &
\multirow{2.5}{*}{\textbf{Scoring}} & 
\multicolumn{2}{c}{\textbf{Task}} \\
\cmidrule(lr){8-9}
& & & & & & & \textbf{BE} & \textbf{JSN} \\
\midrule
Langs et al.~\cite{langs2007model} & 2007 & Appearance model & Private & 17 (-) & 8 & Binary & \checkmark &  \\
Langs et al.~\cite{langs2008automatic} & 2008 & ASM & Private & 57 (-) & 28 & Modified Sharp & \checkmark & \checkmark \\
Murakami et al.~\cite{murakami2018automatic} & 2018 & DCNN & Private & 159 (-) & 159 & Binary & \checkmark &  \\
Rohrbach et al.~\cite{rohrbach2019bone} & 2019 & VGG & Private & - & - & Ratingen & \checkmark &  \\
Hirano et al.~\cite{hirano2019development} & 2019 & CNN & Private & 216 (-) & 108 & SvdH & \checkmark & \checkmark \\
Maziarz et al.~\cite{maziarz2021deep} & 2021 & CNN & \cite{sun2022crowdsourcing} & 674 (-) & 562 & SvdH & \checkmark & \checkmark \\
Hioki et al.~\cite{hioki2021evaluation} & 2021 & YOLO V3 & Private & 50 (20) & - & SvdH & \checkmark &  \\
Miyama et al.~\cite{miyama2022deep} & 2022 & VGG & Private & 226 (31) & 40 & SvdH & \checkmark & \checkmark \\
Wang et al.~\cite{wang2022deep} & 2022 & EfficientNet & Private & 915 (30) & 400 & mTSS &  & \checkmark \\
Stolpovsky et al.~\cite{stolpovsky2023rheumavit} & 2023 & ViT & Public & 330 (42) & 330 & Modified Sharp & \checkmark & \checkmark \\ 
Bo et al.~\cite{bo2024deep} & 2024 & ResNet + MobileNetV2 & Private & 3818 (-) & - & SvdH & \checkmark & \checkmark \\
Bird et al.~\cite{bird2025ai} & 2025 & DenseNet & Private & 2059 (-) & 410 & SvdH & \checkmark & \checkmark \\
Lien et al.~\cite{lien2025deep} & 2025 & EfficientNetV2 & Private & 823 (30) & - & mTSS &  & \checkmark \\
\bottomrule
\end{tabular}
}
\end{threeparttable}
\begin{tablenotes}
\item \textbf{ASM}: Active Shape Model; \textbf{mTSS}: Modified Total Sharp Score.
\end{tablenotes}
\end{table*}

Qualitative evaluation of BE and JSN in RA is typically performed at predefined joint sites based on expert interpretation of radiographic abnormalities, and remains an important component of clinical image assessment. As summarized in Table~\ref{tab:svdh_be_summary}, most existing studies formulate these tasks as joint-level grading, classification, or score prediction, with the goal of estimating the severity of structural damage from local joint appearances rather than learning spatially explicit lesion representations.

Early studies mainly relied on model-based methods for BE analysis or combined BE / JSN assessment on small private datasets~\cite{langs2007model,langs2008automatic}. These approaches incorporated structural priors and handcrafted representations to capture radiographic changes, but their evaluation was limited by dataset scale and restricted clinical variability. More recent work has shifted toward deep learning-based frameworks. Representative studies use CNNs, VGG, EfficientNet, and ViT models to predict BE and/or JSN grades from cropped joints or regional image patches~\cite{murakami2018automatic,rohrbach2019bone,hirano2019development,maziarz2021deep,miyama2022deep,wang2022deep,stolpovsky2023rheumavit,bo2024deep,bird2025ai,lien2025deep}. These methods have improved automated radiographic evaluation and enabled more scalable qualitative assessment, but they generally focus on joint-level outputs and formulate the problem as grading or scoring rather than dense prediction.

\section{Detailed Information of Dataset}\label{sec:detail_info}
\subsection{License and Attribution}\label{sec:license}
The conventional radiographs and associated annotations (bone structure segmentation masks, BE segmentation masks, SvdH-defined joint ROIs for BE / JSN scoring, and SvdH BE / JSN scores) in the dataset are licensed under the Creative Commons Attribution-NonCommercial-ShareAlike 4.0 International License (CC BY-NC-SA 4.0).

For proper attribution when using this dataset in any publications or research outputs, please cite with the DOI. 

\textbf{\textit{Suggested Citation:}} Yang, S., Wang, H., Fu, Y., Peng, J., Fan, L., Chen, H., Song, J., Ikebe, M., Takamaeda-Yamazaki, S., Okutomi, M., Kamishima, T., \& Ou, Y.(2026). RAM-H1200: A Unified Evaluation and Dataset on Hand Radiographs for Rheumatoid Arthritis.

\subsection{Data Rights Compliance and Issue Reporting} \label{sec:data_protection}

We are committed to complying data protection rights in accordance with relevant regulations, including but not limited to the General Data Protection Regulation (GDPR). All personally identifiable information (PII) has been removed through anonymization techniques. If any individual represented in the dataset wishes to have their data removed, we provide a clear and accessible process for issue reporting and resolution via our GitHub repository. Concerned parties are encouraged to contact the authors directly through the contact form linked on the GitHub page. Upon receiving a request, we will engage with the individual to verify their identity and promptly remove the relevant data entries from the dataset.

\begin{table*}[!t]
\small
\caption{Radiographic imaging configuration parameters}
\label{tab:medical_info_AP}
\centering
\begin{threeparttable}
\setlength{\tabcolsep}{0mm}{
\begin{tabular}{p{3cm}<{\centering}p{1.8cm}<{\centering}p{1.8cm}<{\centering}p{1.8cm}<{\centering}p{1.8cm}<{\centering}p{1.8cm}<{\centering}p{1.8cm}<{\centering}}
\toprule
& \textbf{HMCRD} & \textbf{SCGH} & \textbf{SARC} & \textbf{HU1} & \textbf{HU2} & \textbf{HU3} \\
\midrule
Model & Radnext 32 & KXO-50G & CS-7 & - & - & - \\
Manufacturer & HITACHI & TOSHIBA & KONICA MINOLTA & FUJIFILM & FUJIFILM & FUJIFILM \\
Aluminum filter (mm) & 0.5 & NO & - & - & - & - \\
Tube voltage (kV) & 50 & 45 & - & - & - & - \\
Tube current (mA) & 100 & 250 & - & - & - & - \\
Exposure time (mSec) & 25 & 14 & - & - & - & - \\
Source to image (cm) & 100 & 100 & - & - & - & - \\
Resolution (mm/pixel) & 0.15 & 0.15 & 0.175 & 0.15 & 0.15 & 0.15 \\
Image size (pixel) & 2010$\times$1490 & 2010$\times$1490 & 1430$\times$1722 & 2010$\times$1670 & 2010$\times$1670 & 2010$\times$1670 \\
Bit depth (bit) & 10 & 10 & 12 & 10 & 12 & 12 \\
\bottomrule
\end{tabular}
}
\raggedright
\begin{tablenotes}
\item \textbf{HMCRD}: Hokkaido Medical Center for Rheumatic Diseases, Japan.
\item \textbf{SCGH}: Sapporo City General Hospital, Japan.
\item \textbf{SARC}: Sagawa Akira Rheumatology Clinic, Japan.
\item \textbf{HU1\textasciitilde3}: Faculty of Health Sciences, Hokkaido University, Japan.
\end{tablenotes}
\end{threeparttable}
\end{table*}

\subsection{Data Acquisition}
\label{sec:acquisition}
Radiographs were collected from six institutions with varying imaging configurations, including differences in equipment models, acquisition settings, and image resolutions, as shown in Table~\ref{tab:medical_info_AP}.

\subsection{Annotation and Disagreement-Resolution Protocol}
\label{sec:annotation_protocol_AP}
All annotations were produced through a multi-stage protocol involving three radiological technologists, an orthopedic doctor with 7 years of clinical experience, and a board-certified radiologist with 26 years of experience. The participating personnel differed according to annotation type. For the segmentation tasks, including hand bone structures and pixel-level BE lesions, initial contours were delineated by the radiological technologists. For the SvdH BE and JSN scoring tasks, the three radiological technologists and the orthopedic doctor participated in assigning joint-level scores. 

For each image, at least two annotators independently produced the initial annotations without consensus discussion. These initial annotations were kept separate before adjudication, allowing inter-annotator variability to be assessed prior to consensus. When discrepancies were identified, the corresponding annotations were jointly reviewed by the participating annotators. Differences in anatomical boundaries, BE lesion delineation, or SvdH grades were discussed with reference to the predefined annotation criteria and resolved through discussion-based consensus.

After consensus was reached, the resulting annotations underwent a final review by the senior board-certified radiologist. This final verification served as the last quality-control stage before the annotations were included in the released dataset. Therefore, the benchmark annotations represent consensus labels after independent annotation and disagreement resolution, rather than labels produced by a single reader.

\begin{figure}[!t]
    \centering
    \includegraphics[width=\textwidth]{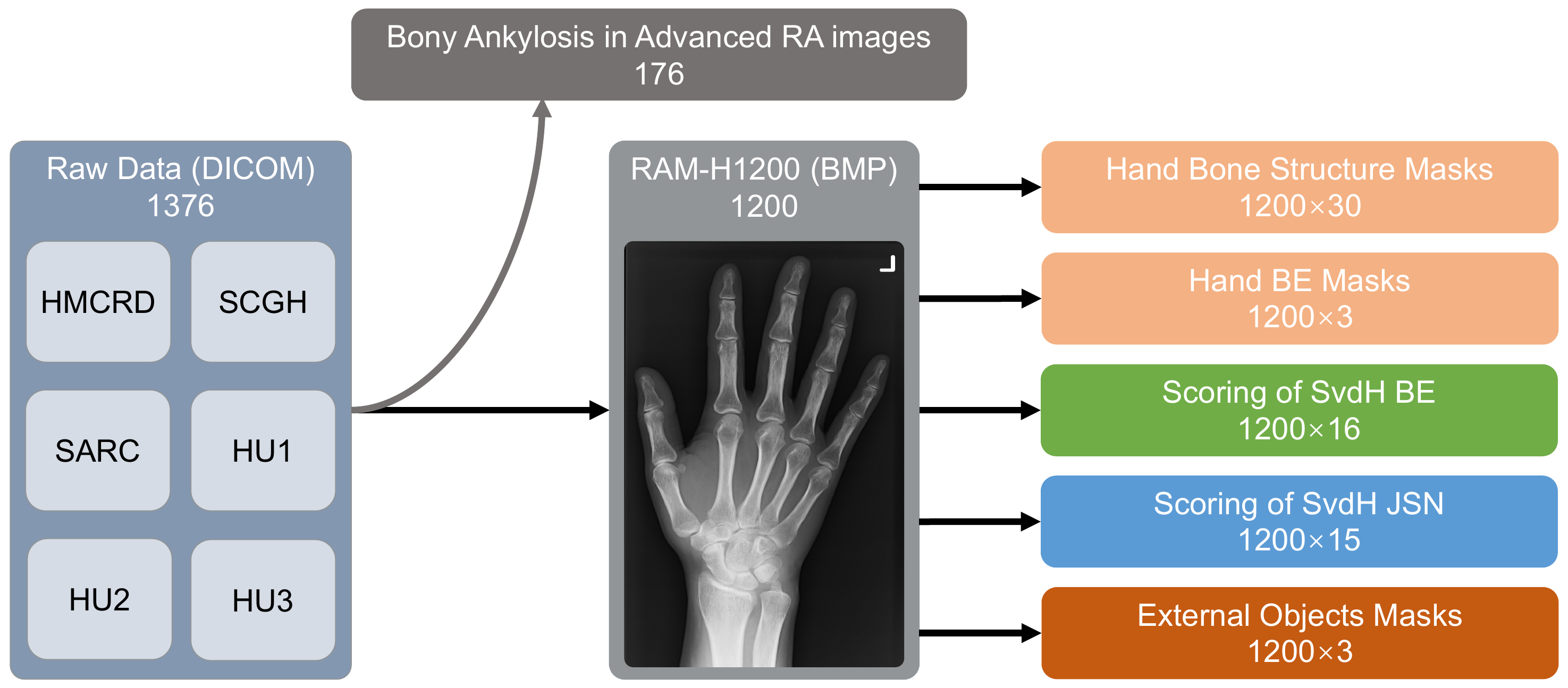}
    \caption{Overview of the data collection and processing pipeline for RAM-H1200. A total of 1376 DICOM-format hand radiographs were collected from six institutions. After excluding 176 end-stage RA radiographs exhibiting bony ankylosis, 1200 images were retained and converted to BMP format. The dataset supports multiple tasks, including hand bone structure segmentation (1200$\times$30 masks), bone erosion (BE) segmentation (1200$\times$3 masks), SvdH BE scoring (1200$\times$16 joints), SvdH JSN scoring (1200$\times$15 joints), and external object segmentation (1200$\times$3 masks, covering catheters, implants, and rings).}
    \label{fig:data_filter}
\end{figure}

\subsection{Data Pre-Processing}

As illustrated in Fig.~\ref{fig:data_filter}, a total of 1376 DICOM-format hand radiographs were initially collected from six institutions. Compared to \cite{yang2025ram}, the collected cohort includes a larger proportion of moderate-to-advanced RA cases, which increases annotation complexity due to severe anatomical deformation and structural overlap. 
However, 176 end-stage RA radiographs exhibiting bony ankylosis were excluded during preprocessing. These cases typically present radiographically conspicuous bone fusion and severe joint deformation and constitute a readily scored high-severity tail of the cohort. Their exclusion therefore reduces the representation of end-stage disease while retaining the early-to-moderate cases targeted for fine-grained assessment in the benchmark. After exclusion, 1,200 radiographs were retained for subsequent annotation and evaluation.
All selected radiographs were originally stored in DICOM format with 10- or 12-bit intensity depth, depending on the acquisition system. During conversion to 8-bit BMP format, each image underwent a white-balance step followed by linear tone mapping to the 8-bit intensity range. The radiographs were subsequently resampled to a uniform spatial resolution of 0.175~mm/pixel to standardize the spatial scale across acquisition systems. The resulting dataset forms the basis for multiple downstream tasks, including hand bone structure segmentation, BE segmentation, and SvdH-based scoring for both BE and JSN.

\begin{table*}[!t]
\caption{Score distributions across centers in train, valid, and test sets.}
\centering

\begin{subtable}[!t]{0.48\textwidth}
\centering
\caption{\textbf{\textit{Hand Bone Structure Segmentation}}}
\label{tab:center_distribution_hand_structure}
\resizebox{\linewidth}{!}{
\begin{tabular}{cccccccc}
\toprule
\textbf{SvdH} & \textbf{HMCRD} & \textbf{SARC} & \textbf{SCGH} & \textbf{HU1} & \textbf{HU2} & \textbf{HU3} \\
\midrule
\multicolumn{7}{c}{\textit{Train Set}} \\
\midrule
0--4   & 153 & 68  & 19 & 0 & 4 & 1 \\
5--9   & 106 & 86  & 40 & 2 & 1 & 1 \\
10--19 & 51  & 108 & 39 & 0 & 3 & 0 \\
20--29 & 33  & 28  & 7  & 0 & 0 & 0 \\
30--39 & 21  & 6   & 2  & 0 & 0 & 0 \\
40--59 & 13  & 1   & 0  & 0 & 0 & 0 \\
60+    & 0   & 0   & 0  & 0 & 0 & 0 \\
\midrule
\multicolumn{7}{c}{\textit{Valid Set}} \\
\midrule
0--4   & 28 & 1  & 0 & 1 & 5 & 0 \\
5--9   & 36 & 1  & 0 & 1 & 3 & 0 \\
10--19 & 37 & 13 & 1 & 0 & 0 & 0 \\
20--29 & 3  & 7  & 1 & 0 & 0 & 0 \\
30--39 & 1  & 0  & 0 & 0 & 0 & 0 \\
40--59 & 1  & 0  & 0 & 0 & 0 & 0 \\
60+    & 0  & 0  & 0 & 0 & 0 & 0 \\
\midrule
\multicolumn{7}{c}{\textit{Test Set}} \\
\midrule
0--4   & 54 & 0  & 0 & 0 & 1 & 0 \\
5--9   & 72 & 17 & 1 & 0 & 1 & 0 \\
10--19 & 58 & 22 & 2 & 0 & 6 & 0 \\
20--29 & 11 & 5  & 1 & 0 & 0 & 0 \\
30--39 & 7  & 0  & 0 & 0 & 0 & 0 \\
40--59 & 1  & 4  & 1 & 0 & 0 & 0 \\
60+    & 2  & 1  & 0 & 0 & 0 & 0 \\
\bottomrule
\end{tabular}}
\end{subtable}
\hfill
\begin{subtable}[!t]{0.506\textwidth}
\centering
\caption{\textbf{\textit{Hand BE Segmentation}}}
\label{tab:center_distribution_be_seg}
\resizebox{\linewidth}{!}{
\begin{tabular}{cccccccc}
\toprule
\textbf{SvdH BE} & \textbf{HMCRD} & \textbf{SARC} & \textbf{SCGH} & \textbf{HU1} & \textbf{HU2} & \textbf{HU3} \\
\midrule
\multicolumn{7}{c}{\textit{Train Set}} \\
\midrule
0--4   & 204 & 165 & 82 & 2 & 5 & 2 \\
5--9   & 104 & 84  & 19 & 0 & 3 & 0 \\
10--14 & 35  & 36  & 5  & 0 & 0 & 0 \\
15--19 & 17  & 8   & 1  & 0 & 0 & 0 \\
20--24 & 6   & 3   & 0  & 0 & 0 & 0 \\
25--29 & 7   & 1   & 0  & 0 & 0 & 0 \\
30+    & 4   & 0   & 0  & 0 & 0 & 0 \\
\midrule
\multicolumn{7}{c}{\textit{Valid Set}} \\
\midrule
0--4   & 53 & 6  & 1 & 2 & 8 & 0 \\
5--9   & 30 & 12 & 1 & 0 & 0 & 0 \\
10--14 & 21 & 4  & 0 & 0 & 0 & 0 \\
15--19 & 1  & 0  & 0 & 0 & 0 & 0 \\
20--24 & 1  & 0  & 0 & 0 & 0 & 0 \\
25--29 & 0  & 0  & 0 & 0 & 0 & 0 \\
30+    & 0  & 0  & 0 & 0 & 0 & 0 \\
\midrule
\multicolumn{7}{c}{\textit{Test Set}} \\
\midrule
0--4   & 110 & 0  & 3 & 0 & 2 & 0 \\
5--9   & 64  & 21 & 1 & 0 & 5 & 0 \\
10--14 & 17  & 21 & 0 & 0 & 1 & 0 \\
15--19 & 8   & 2  & 1 & 0 & 0 & 0 \\
20--24 & 3   & 1  & 0 & 0 & 0 & 0 \\
25--29 & 1   & 3  & 0 & 0 & 0 & 0 \\
30+    & 2   & 1  & 0 & 0 & 0 & 0 \\
\bottomrule
\end{tabular}}
\end{subtable}

\vspace{0.4cm}

\begin{subtable}[!t]{0.49\textwidth}
\centering
\caption{\textbf{\textit{Scoring of SvdH BE}}}
\label{tab:center_distribution_be_score}
\resizebox{\linewidth}{!}{
\begin{tabular}{cccccccc}
\toprule
\textbf{SvdH BE} & \textbf{HMCRD} & \textbf{SARC} & \textbf{SCGH} & \textbf{HU1} & \textbf{HU2} & \textbf{HU3} \\
\midrule
\multicolumn{7}{c}{\textit{Train Set}} \\
\midrule
0 & 4716 & 3681 & 1477 & 32 & 107 & 32 \\
1 & 887  & 796  & 168  & 0  & 18  & 0  \\
2 & 275  & 212  & 46   & 0  & 3   & 0  \\
3 & 21   & 34   & 18   & 0  & 0   & 0  \\
5 & 133  & 29   & 3    & 0  & 0   & 0  \\
\midrule
\multicolumn{7}{c}{\textit{Valid Set}} \\
\midrule
0 & 1307 & 242 & 24 & 30 & 113 & 0 \\
1 & 288  & 78  & 7  & 2  & 15  & 0 \\
2 & 69   & 29  & 1  & 0  & 0   & 0 \\
3 & 7    & 3   & 0  & 0  & 0   & 0 \\
5 & 25   & 0   & 0  & 0  & 0   & 0 \\
\midrule
\multicolumn{7}{c}{\textit{Test Set}} \\
\midrule
0 & 2540 & 426 & 63 & 0 & 89 & 0 \\
1 & 440  & 232 & 9  & 0 & 36 & 0 \\
2 & 251  & 94  & 3  & 0 & 3  & 0 \\
3 & 8    & 14  & 3  & 0 & 0  & 0 \\
5 & 41   & 18  & 2  & 0 & 0  & 0 \\
\bottomrule
\end{tabular}}

\end{subtable}
\hfill
\begin{subtable}[!t]{0.498\textwidth}
\centering
\caption{\textbf{\textit{Scoring of SvdH JSN}}}
\label{tab:center_distribution_jsn_score}
\resizebox{\linewidth}{!}{
\begin{tabular}{cccccccc}
\toprule
\textbf{SvdH JSN} & \textbf{HMCRD} & \textbf{SARC} & \textbf{SCGH} & \textbf{HU1} & \textbf{HU2} & \textbf{HU3} \\
\midrule
\multicolumn{7}{c}{\textit{Train Set}} \\
\midrule
0 & 4758 & 3277 & 992 & 21 & 92 & 22 \\
1 & 491  & 802  & 482 & 4  & 24 & 5  \\
2 & 202  & 286  & 112 & 5  & 1  & 3  \\
3 & 85   & 60   & 16  & 0  & 3  & 0  \\
4 & 119  & 30   & 3   & 0  & 0  & 0  \\
\midrule
\multicolumn{7}{c}{\textit{Valid Set}} \\
\midrule
0 & 1280 & 215 & 11 & 23 & 100 & 0 \\
1 & 248  & 52  & 5  & 6  & 20  & 0 \\
2 & 44   & 43  & 14 & 1  & 0   & 0 \\
3 & 13   & 20  & 0  & 0  & 0   & 0 \\
4 & 5    & 0   & 0  & 0  & 0   & 0 \\
\midrule
\multicolumn{7}{c}{\textit{Test Set}} \\
\midrule
0 & 2422 & 594 & 30 & 0 & 76 & 0 \\
1 & 426  & 84  & 32 & 0 & 42 & 0 \\
2 & 153  & 30  & 7  & 0 & 2  & 0 \\
3 & 53   & 7   & 4  & 0 & 0  & 0 \\
4 & 21   & 20  & 2  & 0 & 0  & 0 \\
\bottomrule
\end{tabular}}
\end{subtable}

\label{tab:dataset_distribution_institution_AP}
\end{table*}

\subsection{Dataset Split}
To avoid patient-level data leakage, the train, validation, and test sets were split at the patient level, ensuring that all radiographs from the same patient, including bilateral hands and longitudinal follow-up studies, were assigned to the same subset. 
Table~\ref{tab:dataset_distribution_institution_AP} details the institution-wise distribution across the four related tasks. Table~\ref{tab:center_distribution_hand_structure} summarizes the institution-wise distribution for the hand bone structure segmentation dataset, while Table~\ref{tab:center_distribution_be_seg} presents the corresponding distribution for the BE segmentation dataset. Table~\ref{tab:center_distribution_be_score} and Table~\ref{tab:center_distribution_jsn_score} further show the score distributions for the SvdH BE scoring and SvdH JSN scoring datasets, respectively. This breakdown highlights the contribution of each collaborating institution and illustrates how score imbalance manifests across different sources and subsets.

\begin{table*}[!t]
\centering
\caption{Detailed annotation reliability statistics for BE, JSN, and their combined scores. Reliability was quantified using the intraclass correlation coefficient under the one-way random-effects single-measure setting, ICC(1,1). MSB and MSW denote the mean square between subjects and the mean square within subjects, respectively.}
\label{tab:icc_reliability_AP}
\resizebox{0.55\textwidth}{!}{
\begin{tabular}{lccc}
\toprule
\textbf{Task} & \textbf{MSB} & \textbf{MSW} & \textbf{ICC(1,1)} \\
\midrule
SvdH BE Score & 0.9700 & 0.2671 & 0.5682 \\
SvdH JSN Score & 1.4466 & 0.5485 & 0.4502 \\
SvdH BE + JSN Score & 1.2686 & 0.4032 & 0.5176 \\
\bottomrule
\end{tabular}}
\end{table*}

\subsection{Annotation Reliability Analysis}
\label{app:icc_analysis_AP}

Following the annotation protocol described in Appendix~\ref{sec:annotation_protocol_AP}, we additionally quantified inter-annotator reliability using the initial independent scoring annotations obtained before consensus adjudication. We report the detailed annotation reliability statistics in Table~\ref{tab:icc_reliability_AP}.
The intraclass correlation coefficient (ICC) was calculated using the ICC(1,1) formulation, where MSB and MSW denote the mean square between subjects and the mean square within subjects, respectively. The ICC values were computed from the initial independent annotations before consensus discussion, and therefore reflect the baseline inter-annotator agreement of the scoring process.

The obtained ICC values indicate moderate initial agreement for BE and the combined BE/JSN annotations, while JSN shows relatively lower agreement. This trend is expected because JSN scoring depends on subtle differences in joint space width and alignment, which can be affected by projection variation, anatomical overlap, and adjacent-grade ambiguity. The observed initial agreement is broadly consistent with prior RA radiographic scoring studies, where SvdH-based intra- and interobserver ICC values were reported in a similar range~\cite{fujimori2018composite}. After this initial assessment, discrepant cases were reviewed and resolved through consensus discussion to produce the final annotations used in RAM-H1200. These results suggest that the annotation process provides a reasonable and clinically meaningful reference, while also reflecting the inherent subjectivity of fine-grained radiographic scoring. 

\subsection{Independent-Reader SvdH Scoring Baseline}
\label{app:human_scoring_baseline}

To contextualize model performance relative to clinical assessment, an independent radiologist re-scored the complete SvdH scoring test set of 267 hand radiographs. The independent-reader scores were evaluated against the same consensus reference used for all benchmark models, yielding 4,272 BE joint-level comparisons (267 hands $\times$ 16 joints) and 4,005 JSN joint-level comparisons (267 hands $\times$ 15 joints). No images or joints were excluded from this analysis. The independent
reader and all nine benchmark models were evaluated using the same QWK, MAE, ACC, and W1-ACC metrics.

Table~\ref{tab:human_scoring_baseline_AP} contextualizes model performance using an independent radiologist evaluated against the same consensus reference. For SvdH BE scoring, the independent reader achieves a QWK of
0.7334 and an MAE of 0.2285, while the evaluated models achieve QWK values up to 0.4521 and MAE values as low as 0.3701. For SvdH JSN scoring, the corresponding reader results are 0.8279 and 0.1740, compared with model results of up to 0.5967 in QWK and as low as 0.2769 in MAE.

The independent-reader results also provide context for the variability of the consensus-based scoring labels. Exact-grade accuracy is 80.66\% for BE and 83.15\% for JSN, while W1-ACC reaches 97.17\% and 99.45\%, respectively, indicating that most reader--consensus differences fall within one grade. These results provide a clinical reference for interpreting the benchmark results while illustrating the variability associated with fine-grained SvdH grading. This analysis represents an independent-reader-versus-consensus comparison rather than a direct measurement of inter-observer agreement or an absolute human performance ceiling. We further examine whether joint-level cases with reader--consensus disagreement are also more difficult for the evaluated models in Sec.~\ref{sec:agreement_stratified_AP}.

\begin{table*}[!t]
\centering
\caption{
Independent-reader clinical scoring baseline on the complete test set. The independent reader and all benchmark models are evaluated against the same consensus reference. The independent-reader row serves as a clinical reference and is not included in the model ranking.
}
\label{tab:human_scoring_baseline_AP}
\resizebox{\textwidth}{!}{
\begin{tabular}{lcccccccc}
\toprule
& \multicolumn{4}{c}{\textbf{SvdH BE Scoring}}
& \multicolumn{4}{c}{\textbf{SvdH JSN Scoring}} \\
\cmidrule(lr){2-5}\cmidrule(lr){6-9}

& \textbf{QWK} 
& \textbf{MAE} 
& \textbf{ACC (\%)} 
& \textbf{W1-ACC (\%)} 
& \textbf{QWK} 
& \textbf{MAE} 
& \textbf{ACC (\%)} 
& \textbf{W1-ACC (\%)} \\ 
\midrule

Independent reader
& 0.7334
& 0.2285
& 80.66
& 97.17
& 0.8279
& 0.1740
& 83.15
& 99.45 \\
\midrule

ResNet
& 0.4420 & 0.3909 & 69.66 & 92.81
& 0.5884 & 0.2824 & 76.03 & 96.33 \\

DenseNet
& 0.3899 & 0.3701 & 73.06 & 91.62
& 0.5829 & 0.3031 & 74.08 & 96.13 \\

EfficientNetV2
& 0.3356 & 0.3940 & 71.28 & 91.71
& 0.5393 & 0.3011 & 75.43 & 95.18 \\

MobileViT
& 0.3920 & 0.4050 & 69.41 & 92.25
& 0.5967 & 0.2871 & 76.00 & 95.98 \\

LeViT
& 0.2343 & 0.4242 & 70.22 & 90.94
& 0.5445 & 0.3006 & 75.53 & 95.48 \\

EfficientFormer
& 0.3504 & 0.3951 & 70.46 & 92.13
& 0.5919 & 0.2769 & 76.75 & 96.23 \\

ConvNeXtV2
& 0.3058 & 0.4127 & 70.04 & 91.39
& 0.5151 & 0.2941 & 76.13 & 95.36 \\

MedMamba
& 0.4521 & 0.3961 & 70.13 & 92.11
& 0.5738 & 0.2934 & 75.46 & 95.86 \\

MambaVision
& 0.3667 & 0.3804 & 72.85 & 91.15
& 0.5457 & 0.2899 & 76.00 & 95.88 \\

\bottomrule
\end{tabular}
}
\end{table*}

\subsection{Agreement-Stratified SvdH Scoring}
\label{sec:agreement_stratified_AP}

To characterize how scoring performance varies with reader--consensus consistency, we divide the complete test set according to whether an independent reader assigns the same exact SvdH grade as the consensus reference. The \emph{high-agreement} group contains joints for which the independent-reader and consensus grades are identical, whereas the \emph{low-agreement} group contains joints for which they differ. This results in 3,446 high-agreement and 826 low-agreement joints for BE, and 3,330 high-agreement and 675 low-agreement joints for JSN. In both groups, model predictions are evaluated against the same consensus reference; the independent reader is used only to define the groups.

Table~\ref{tab:agreement_stratified_AP} shows a substantial performance difference between the two groups across all evaluated models. For BE, the mean QWK across all models decreases from 0.4753 in the high-agreement group to 0.1012 in the low-agreement group, accompanied by an increase in MAE from 0.2589 to 0.9703, a decrease in accuracy from 80.90\% to 28.63\%, and a decrease in W1-ACC from 95.05\% to 78.18\%. A similar pattern is observed for JSN, where mean QWK decreases from 0.6234 to 0.3452, MAE increases from 0.1941 to 0.7756, accuracy decreases from 83.62\% to 36.71\%, and W1-ACC decreases from 97.41\% to 87.98\%. This direction is consistent across all models and all four metrics for both BE and JSN, indicating that joints associated with reader--consensus disagreement are also more difficult for automated scoring.

We further assess the robustness of this difference using patient-level cluster bootstrap analysis ($B=2000$), resampling patients while retaining all associated hands and joints. The 95\% confidence intervals of the high- and low-agreement groups are non-overlapping for MAE in all models for both BE and JSN. For QWK, the intervals are non-overlapping for all BE models and six of the nine JSN models. The performance separation is therefore consistently observed at the patient level, although it is less uniform for JSN QWK.

We also examine whether the relationship between reader disagreement and model difficulty varies across anatomical sites. Across joint types, the independent-reader disagreement rate is strongly associated with the mean model MAE, with Pearson $r=0.9528$ ($p=1.2\times10^{-8}$) for BE and $r=0.8937$ ($p=7.1\times10^{-6}$) for JSN. Joint types at which the independent reader more frequently differs from the consensus therefore also tend to exhibit larger model errors.

The performance gap between the two groups is accompanied by differences in case composition. In particular, the low-agreement group has a substantially different consensus-grade distribution and is unevenly distributed across anatomical sites. Grade-matched analyses further show that the MAE gap becomes considerably smaller for the most common grades, indicating that differences in grade composition contribute substantially to the aggregate performance gap. Taken together, these results show that reader--consensus disagreement serves as a useful marker of difficult scoring cases, while the observed performance difference also reflects differences in the composition of the two groups. This analysis therefore provides a complementary characterization of benchmark difficulty beyond performance averaged over the complete test set.

\begin{table*}[!t]
\centering
\caption{
Agreement-stratified SvdH scoring performance on the complete test set.
}
\label{tab:agreement_stratified_AP}
\resizebox{\textwidth}{!}{
\begin{tabular}{lcccccccc}
\toprule
& \multicolumn{4}{c}{\textbf{High-agreement subset}}
& \multicolumn{4}{c}{\textbf{Low-agreement subset}} \\
\cmidrule(lr){2-5}
\cmidrule(lr){6-9}
\textbf{Model}
& \textbf{QWK}
& \textbf{MAE}
& \textbf{ACC (\%)}
& \textbf{W1-ACC (\%)}
& \textbf{QWK}
& \textbf{MAE}
& \textbf{ACC (\%)}
& \textbf{W1-ACC (\%)} \\
\midrule

\multicolumn{9}{l}{\textbf{(a) SvdH BE scoring}} \\
\midrule
ResNet        & 0.5713 & 0.2571 & 79.66 & 95.85 & 0.0955 & 0.9492 & 27.97 & 80.15 \\
DenseNet        & 0.5271 & 0.2200 & 84.16 & 95.15 & 0.1332 & 0.9964 & 26.76 & 76.88 \\
EfficientNetV2  & 0.4548 & 0.2530 & 81.66 & 95.01 & 0.0763 & 0.9818 & 27.97 & 77.97 \\
MobileViT       & 0.4953 & 0.2786 & 78.82 & 95.15 & 0.0997 & 0.9322 & 30.15 & 80.15 \\
LeViT           & 0.3134 & 0.2896 & 79.83 & 94.25 & 0.0505 & 0.9855 & 30.15 & 77.12 \\
EfficientFormer & 0.4595 & 0.2586 & 80.67 & 95.27 & 0.1093 & 0.9649 & 27.85 & 79.06 \\
ConvNeXtV2      & 0.4069 & 0.2748 & 80.15 & 94.66 & 0.0774 & 0.9879 & 27.85 & 77.72 \\
MedMamba        & 0.5637 & 0.2644 & 79.63 & 95.36 & 0.1323 & 0.9455 & 30.51 & 78.57 \\
MambaVision   & 0.4856 & 0.2345 & 83.49 & 94.78 & 0.1367 & 0.9891 & 28.45 & 76.03 \\
\midrule
Mean            & 0.4753 & 0.2589 & 80.90 & 95.05 & 0.1012 & 0.9703 & 28.63 & 78.18 \\

\midrule
\multicolumn{9}{l}{\textbf{(b) SvdH JSN scoring}} \\ \midrule
ResNet        & 0.6558 & 0.1886 & 83.57 & 97.87 & 0.3453 & 0.7452 & 38.81 & 88.74 \\
DenseNet        & 0.6308 & 0.2102 & 81.47 & 97.78 & 0.3763 & 0.7615 & 37.63 & 88.00 \\
EfficientNetV2  & 0.5863 & 0.2036 & 83.33 & 96.79 & 0.3349 & 0.7822 & 36.44 & 87.26 \\
MobileViT       & 0.6568 & 0.1892 & 83.96 & 97.51 & 0.3635 & 0.7704 & 36.74 & 88.44 \\
LeViT           & 0.5848 & 0.2036 & 83.48 & 97.03 & 0.3592 & 0.7793 & 36.30 & 87.85 \\
EfficientFormer & 0.6481 & 0.1805 & 84.74 & 97.60 & 0.3797 & 0.7526 & 37.33 & 89.48 \\
ConvNeXtV2      & 0.6027 & 0.1850 & 84.59 & 97.27 & 0.2775 & 0.8326 & 34.37 & 85.93 \\
MedMamba        & 0.6365 & 0.1949 & 83.39 & 97.54 & 0.3450 & 0.7793 & 36.30 & 87.56 \\
MambaVision   & 0.6092 & 0.1910 & 84.02 & 97.36 & 0.3250 & 0.7778 & 36.44 & 88.59 \\
\midrule
Mean            & 0.6234 & 0.1941 & 83.62 & 97.41 & 0.3452 & 0.7756 & 36.71 & 87.98 \\
\bottomrule
\end{tabular}
}
\end{table*}

\subsection{Dataset Maintenance}
As the authors and maintainers of this dataset, we affirm that while the dataset is self-contained and does not depend on any external links or content, we may provide future updates, such as adding new cases or incorporating additional tasks. These potential updates aim to enhance the dataset's value while maintaining its long-term usability.

\section{Detailed Information of Tasks}
\begin{table*}[!t]
\centering
\caption{Summary of RAM-H1200 benchmark tasks. For SvdH BE scoring, the adopted grading scheme uses $\{0,1,2,3,5\}$; grade 4 is not defined, and grade 5 denotes complete joint destruction. SvdH JSN scoring uses $\{0,1,2,3,4\}$.}
\label{tab:task_summary_AP}
\small
\setlength{\tabcolsep}{3pt}
\renewcommand{\arraystretch}{1.15}

\begin{threeparttable}
\begin{tabular}{
>{\raggedright\arraybackslash}p{0.14\textwidth}
>{\raggedright\arraybackslash}p{0.15\textwidth}
>{\raggedright\arraybackslash}p{0.10\textwidth}
>{\raggedright\arraybackslash}p{0.12\textwidth}
>{\raggedright\arraybackslash}p{0.14\textwidth}
>{\raggedright\arraybackslash}p{0.24\textwidth}
}
\toprule
\textbf{Task} &
\textbf{Input} &
\textbf{Output} &
\textbf{Primary metric} &
\textbf{What it evaluates} &
\textbf{Challenges} \\
\midrule
Hand bone structure segmentation &
Whole-hand radiograph &
30 bone structure masks &
DSC / NSD / DSC$_O$ / NSD$_O$ &
Anatomical structure modeling &
Projection-induced overlap ambiguity \\
\midrule
Hand BE segmentation &
Whole-hand radiograph / local patch &
Pixel-level BE mask &
DSC / REC / PREC &
Quantitative lesion localization &
Tiny, sparse, and low-contrast lesions \\
\midrule
Scoring of SvdH BE &
SvdH-defined joint ROI &
Ordinal score 0/1/2/3/5 &
QWK / MAE / BACC &
Clinical erosion severity &
Class imbalance and subjective adjacent grades \\
\midrule
Scoring of SvdH JSN &
SvdH-defined joint ROI &
Ordinal score 0/1/2/3/4 &
QWK / MAE / BACC &
Structural narrowing severity &
Adjacent-grade ambiguity and joint-dependent variation \\
\bottomrule
\end{tabular}

\begin{tablenotes}[flushleft]
\item \textbf{BE}: bone erosion; \textbf{JSN}: joint space narrowing; \textbf{SvdH}: Sharp/van der Heijde; 
\item \textbf{DSC}: Dice similarity coefficient; \textbf{NSD}: normalized surface Dice;
\item \textbf{REC}: recall; \textbf{PREC}: precision; \textbf{QWK}: quadratic weighted kappa;
\item \textbf{MAE}: mean absolute error; \textbf{BACC}: balanced accuracy.
\item Subscript $O$ denotes overlap-aware evaluation.
\end{tablenotes}
\end{threeparttable}
\end{table*}

\begin{figure}[!t]
    \centering
    \includegraphics[width=\textwidth]{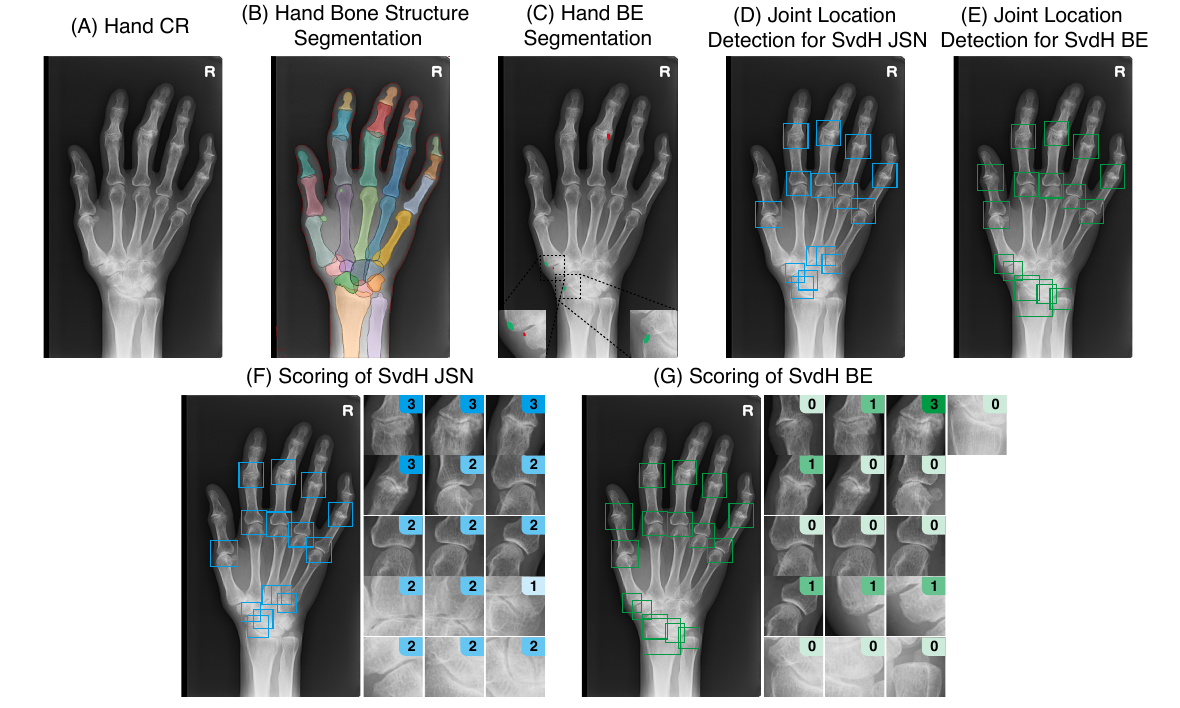}
    \caption{
        Overview of the tasks supported in RAM-H1200. 
        (A) Original hand radiograph (CR). 
        (B) Instance-level segmentation of hand bone structures across the entire hand. 
        (C) Pixel-wise segmentation of BE regions. 
        (D,E) Joint localization for SvdH scoring, where predefined joint regions are detected for JSN and BE assessment, respectively. 
        (F,G) Joint-level SvdH scoring for JSN and BE, where each detected joint is assigned an ordinal severity score based on cropped regions. 
        This figure illustrates the multi-level nature of the benchmark, spanning structure-level modeling, lesion-level analysis, and clinically grounded scoring.
        }
    \label{fig:dataset_tasks_AP}
\end{figure}

RAM-H1200 is designed as a multi-task benchmark for comprehensive analysis of RA from hand radiographs. 
As illustrated in Fig.~\ref{fig:dataset_tasks_AP} and Table~\ref{tab:task_summary_AP}, the dataset supports multiple levels of analysis. 
At the structure level, it provides instance-level annotations for all hand bone structures, enabling hand bone structure segmentation. 
At the lesion level, pixel-wise annotations of BE are provided to facilitate fine-grained pathological analysis. 
At the clinical level, the dataset follows the standardized SvdH protocol, where predefined joint regions are first localized and then assigned ordinal severity scores for both BE and JSN. 
These tasks are included within a common benchmark, allowing the study of relationships between anatomical structures, localized lesions, and clinically interpretable scoring outcomes.

\subsection{Hand Bone Structure Segmentation}

\begin{figure}[!t]
    \centering
    \includegraphics[width=\textwidth]{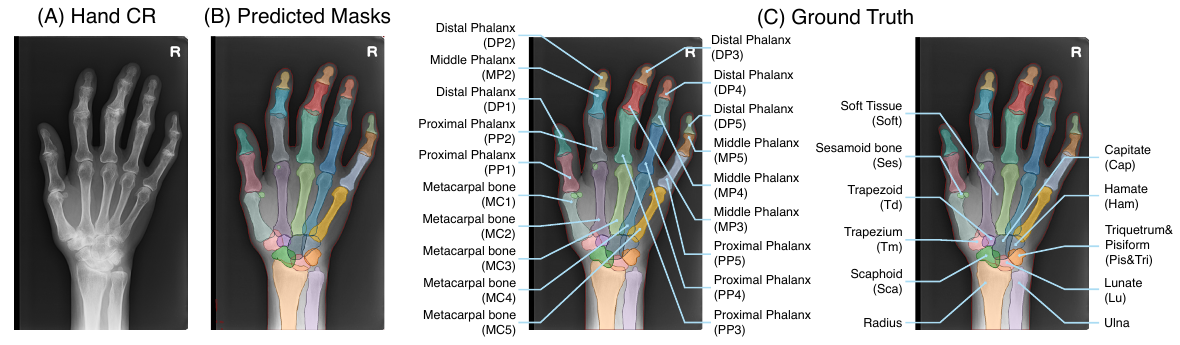}
    \caption{Hand bone structure segmentation task on radiographs. 
        (A) Input hand CR image. 
        (B) Model predictions. 
        (C) Ground truth annotations with fine-grained instance-level labels. 
        The task requires comprehensive segmentation of all hand bones, including phalanges, metacarpals, and carpal structures, as well as surrounding tissues, under strong projection-induced overlap and complex anatomical configurations.}
    \label{fig:seg_task_intro_AP}
\end{figure}

Hand bone structure segmentation aims to delineate all bone instances across the entire hand from radiographs, as illustrated in Fig.~\ref{fig:seg_task_intro_AP}. This task serves as the structural foundation for subsequent analysis, including joint localization, morphological assessment, and clinically grounded scoring.

Compared to wrist-focused settings, hand bone structure segmentation introduces increased complexity due to the larger number of anatomical structures, diverse bone shapes, and extensive projection overlap, particularly in the carpal region. The task requires distinguishing fine-grained boundaries between adjacent bones while maintaining global anatomical consistency across the hand.

Accurate segmentation enables explicit modeling of bone geometry and spatial relationships, which are essential for both structure-oriented tasks such as JSN assessment and lesion-oriented tasks such as BE analysis. The provided annotations support instance-level segmentation of all hand bones, allowing models to
capture both local details and global structural organization.

\subsection{Hand BE Segmentation}

\begin{figure}[!t]
    \centering
    \includegraphics[width=\textwidth]{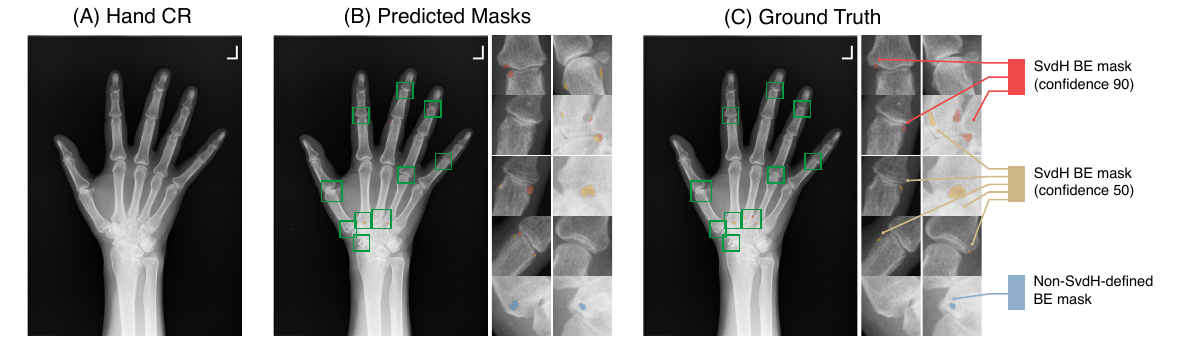}
    \caption{Illustration of the hand BE segmentation task on hand radiographs. 
        (A) Input hand CR image. 
        (B) Representative predicted BE masks with corresponding regions of interest (green boxes), highlighting localized erosion patterns. 
        (C) Ground truth annotations, including SvdH-defined BE regions with different confidence levels (e.g., 90 and 50) and additional non-SvdH-defined erosion regions. 
        The task focuses on detecting and delineating small and subtle erosive lesions under severe class imbalance and ambiguous appearance, particularly in early-stage RA.}
    \label{fig:be_seg_task_intro_AP}
\end{figure}

Hand BE segmentation focuses on pixel-level delineation of erosive lesions in hand radiographs (Fig.~\ref{fig:be_seg_task_intro_AP}). This task enables direct and quantitative assessment of structural damage, providing spatially explicit information beyond traditional grading-based approaches.

Hand BE segmentation is particularly challenging due to the extremely small size of lesions, severe class imbalance, and low contrast between erosion regions and surrounding bone structures. In early-stage RA, erosive changes often manifest as subtle cortical irregularities, making them difficult to distinguish from normal anatomical variations and imaging artifacts.

Furthermore, erosion patterns are closely associated with anatomical structures, typically occurring along bone surfaces and near joint interfaces. As a result, accurate BE segmentation inherently relies on structural context and may benefit from joint modeling of anatomy and pathology.

The dataset provides pixel-level BE annotations categorized according to SvdH principles, enabling fine-grained and anatomically consistent evaluation of erosion detection and delineation.

\subsection{Scoring of SvdH BE}

\begin{figure}[!t]
    \centering
    \includegraphics[width=\textwidth]{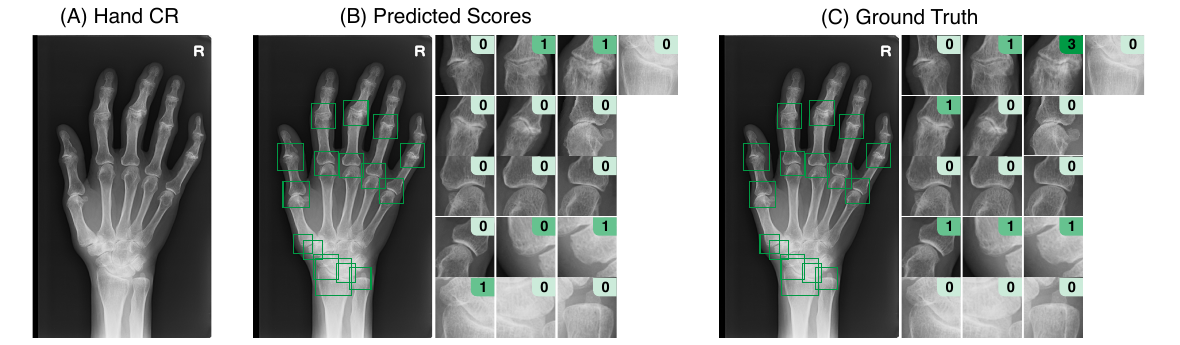}
    \caption{SvdH-based BE scoring task on hand radiographs. 
        (A) Input CR image. 
        (B) Predicted ordinal scores at predefined joint regions, shown with corresponding joint crops. 
        (C) Ground truth scores. 
        Each joint is assigned a discrete SvdH score reflecting the severity of erosion. 
        The task is formulated as an ordinal classification problem, focusing on joint-level assessment rather than explicit delineation of lesion boundaries.}
    \label{fig:be_scoring_task_intro_AP}
\end{figure}
SvdH BE scoring is a key component of the SvdH scoring system, widely adopted in CAD systems for evaluating the severity of joint damage in RA. As illustrated in Fig.~\ref{fig:be_scoring_task_intro_AP}, this task focuses on assessing the degree of bone erosion at predefined joint locations from radiographs, rather than delineating the exact lesion boundaries. Each joint is assigned a discrete severity level based on the extent of structural damage, reflecting progressive stages of erosion.

In our setting, BE is evaluated at \textbf{16 predefined joint locations per hand}, following the SvdH standard. These include \textit{four proximal interphalangeal (PIP) joints (digits 2–5), four metacarpophalangeal (MCP) joints (digits 2–5), the interphalangeal (IP) joint of the thumb, and seven wrist-related joint regions (including radiocarpal and intercarpal articulations)}. Each joint surface is annotated with one of five ordinal classes corresponding to the raw SvdH BE scores of $(0, 1, 2, 3, 5)$. Accordingly, the task is formulated as a joint-level severity prediction problem.

Unlike segmentation-based approaches that aim to localize erosion regions, SvdH BE scoring requires the model to infer the overall severity of damage from the radiographic appearance of a given joint ROI. Since the score labels are inherently ordered, this task is more appropriately modeled as an ordinal classification problem rather than a standard multi-class classification task.

In our implementation, we adopt an ordinal classification strategy, where the model predicts a series of ordered binary thresholds. Specifically, for a 5-class problem, the model outputs 4 binary decisions, each indicating whether the true score exceeds a given threshold. The final score is obtained by counting the number of positive predictions, allowing the model to preserve the ordinal relationships among severity levels while maintaining a simple and effective formulation.

The task is particularly challenging due to the subtle visual differences between adjacent severity levels. Early-stage erosion often manifests as mild cortical irregularities that are difficult to distinguish from normal anatomical variation. In addition, factors such as image quality, projection differences, and complex anatomical overlap in the wrist further hinder consistent assessment. As a result, SvdH BE scoring constitutes a clinically meaningful yet challenging fine-grained ordinal prediction task.

\subsection{Scoring of SvdH JSN}
\begin{figure}[!t]
    \centering
    \includegraphics[width=\textwidth]{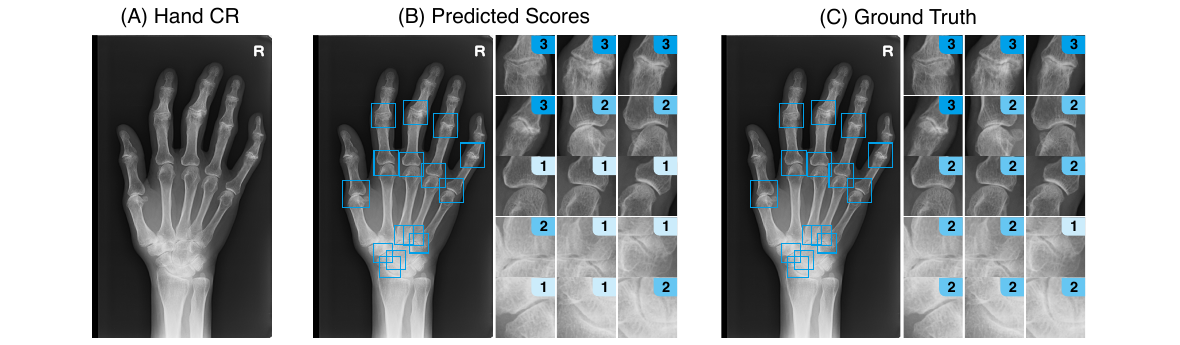}
    \caption{Illustration of the SvdH JSN scoring task. 
        (A) Input hand radiograph. 
        (B) Predicted JSN scores at predefined joint locations, with corresponding cropped regions of interest. 
        (C) Ground truth annotations. 
        Each joint is assigned an ordinal score according to the SvdH system, reflecting the degree of joint space narrowing. 
        Unlike lesion-based analysis, this task focuses on structural assessment of inter-bone spacing at the joint level.}
    \label{fig:jsn_scoring_task_intro_AP}
\end{figure}
SvdH JSN scoring is a key component of the SvdH scoring system, widely adopted in CAD systems for evaluating the progression of joint space narrowing in RA. As illustrated in Fig.~\ref{fig:jsn_scoring_task_intro_AP}, this task focuses on assessing the degree of joint space narrowing progression at predefined joint locations from radiographs, reflecting structural changes associated with disease development over time.

In our setting, JSN is evaluated at \textbf{15 predefined joint locations per hand}, following the SvdH standard. These include:
\textit{four proximal interphalangeal (PIP) joints (digits 2–5), four metacarpophalangeal (MCP) joints (digits 2–5), the interphalangeal (IP) joint of the thumb, and six wrist joint regions (including the radiocarpal and intercarpal articulations)}.
Each joint is assigned one of five ordinal classes corresponding to the raw SvdH JSN scores of $(0, 1, 2, 3, 4)$. Accordingly, the task is formulated as a joint-level severity prediction problem rather than lesion localization.

Compared with BE scoring, SvdH JSN scoring is more structure-oriented, as it primarily depends on the relative spacing and configuration between adjacent bones instead of localized erosive patterns. The model must therefore capture differences in joint space width and joint geometry from radiographic appearance.

Since the score labels are inherently ordered, this task is also formulated as a 5-level ordinal classification problem. We adopt the same ordinal classification formulation as in SvdH BE scoring, where the model predicts a series of ordered binary thresholds and derives the final score accordingly.

The task is particularly challenging due to the subtle visual differences between adjacent severity levels. JSN often manifests as gradual narrowing of joint space, which can be difficult to quantify under projection effects, anatomical overlap, and variations in imaging quality. As a result, reliable scoring requires the model to capture fine-grained structural relationships between adjacent bones in a consistent and clinically meaningful manner, making SvdH JSN scoring a challenging ordinal prediction task.

\section{Implementation Details}
\label{sec:implement_AP}
\subsection{Hand Bone Structure Segmentation}
\label{sec:implement_boneseg_AP}
All radiographs were processed in a patch-wise manner using random cropping with a patch size of $512 \times 512$. For each radiograph, 16 patches were randomly sampled per epoch, with a foreground sampling probability of 0.7. The augmentation pipeline included Resize, RandomScale, CenterCrop, ShiftScaleRotate, and RandomBrightnessContrast transformations.
Since projection-induced bone overlap allows multiple anatomical structures to occupy the same image region, hand bone structure segmentation was formulated as a multi-label segmentation task. Each anatomical structure was predicted by an independent output channel with sigmoid activation, rather than a mutually exclusive softmax output. The training objective was the sum
of mean squared error (MSE), binary cross-entropy with logits (BCEWithLogits), and Dice loss with sigmoid activation:
\begin{equation}
    \mathcal{L}_{\mathrm{seg}}
    =
    \mathcal{L}_{\mathrm{MSE}}
    +
    \mathcal{L}_{\mathrm{BCEWithLogits}}
    +
    \mathcal{L}_{\mathrm{Dice}},
\end{equation}
where all three loss terms were assigned a weight of 1.0. For evaluation, DSC was computed independently for each anatomical structure and then averaged across structures.
Model training employed the AdamW optimizer with a weight decay of $1\times10^{-2}$. The initial learning rate was set to $1\times10^{-4}$ and decayed using a cosine annealing schedule (CosineAnnealingLR). Training was conducted for up to 200 epochs with a batch size of 8 and standard data augmentation techniques. All experiments were conducted with a fixed random seed for reproducibility. Early stopping was applied based on the validation DSC with a patience of 15 epochs. 
During test-time inference, we used MONAI's \path{sliding\_window\_inference} with a window size of $512\times512$, an overlap ratio of 0.5, Gaussian weighting for overlapping predictions, and a sliding-window batch size of 4. No test-time augmentation was applied during validation or testing.

\subsection{Hand BE Segmentation}
\label{sec:implement_beseg_AP}
All radiographs were processed in a patch-wise manner using random cropping with a patch size of $256 \times 256$. For each radiograph, 24 patches were randomly sampled per epoch, with 70\% of the patches sampled from foreground regions and the remaining 30\% from background regions. For foreground sampling, patches were centered around annotated BE lesions with a random spatial displacement of up to $\pm 96$ pixels to increase spatial diversity while retaining lesion content. The augmentation pipeline consisted of Resize, RandomScale, CenterCrop, ShiftScaleRotate, and RandomBrightnessContrast transformations.
Although three-class BE annotations are available, we restrict the benchmark to the SvdH-defined BE category (i.e., SvdH-BE-90) in this setting. Benchmark results using all classes are provided in Appendix~\ref{sec:benchmark_be_all_AP}.
Model training employed the AdamW optimizer with a weight decay of $1\times10^{-2}$. The initial learning rate was set to $1\times10^{-4}$ and decayed using a cosine annealing schedule (CosineAnnealingLR). Training was conducted for up to 200 epochs with a batch size of 16 and standard data augmentation techniques. All experiments were conducted with a fixed random seed for reproducibility. Early stopping was applied based on the validation DSC with a patience of 20 epochs. 
During test-time inference, we used MONAI's \path{sliding\_window\_inference} with a window size of $256\times256$, an overlap ratio of 0.5, Gaussian weighting for overlapping predictions, and a sliding-window batch size of 4. No test-time augmentation was applied during validation or testing.

For nnUnet, we adopted its default experimental planning and training pipeline, where preprocessing, network architecture, and hyperparameters are automatically configured based on the dataset. The model was trained for 500 epochs following the standard nnUnet setting, and all other configurations, including data sampling strategies, were kept unchanged.

\subsection{Scoring of SvdH BE}
\label{sec:implement_be_score_AP}
All joint crops were processed in an image-wise manner and resized to $224 \times 224$. During training, the input images were augmented using ShiftScaleRotate with rotations of up to $\pm10^\circ$, followed by RandomBrightnessContrast. Model training employed an ordinal classification formulation with four binary thresholds optimized using BCEWithLogitsLoss. No label smoothing was applied. The models were optimized using AdamW optimizer with an initial learning rate of $1\times10^{-4}$ and a cosine annealing learning rate schedule (CosineAnnealingLR).
Training was conducted for 200 epochs with a batch size of 32. All experiments were conducted with a fixed random seed for reproducibility. 

\subsection{Scoring of SvdH JSN}
\label{sec:implement_jsn_score_AP}
All joint crops were processed in an image-wise manner and resized to $224 \times 224$. During training, the input images were augmented using ShiftScaleRotate with rotations of up to $\pm10^\circ$, followed by RandomBrightnessContrast. Model training employed an ordinal classification formulation with four binary thresholds optimized using BCEWithLogitsLoss. No label smoothing was applied. The models were optimized using AdamW with an initial learning rate of $1\times10^{-4}$ and a cosine annealing learning rate schedule (CosineAnnealingLR).
Training was conducted for 200 epochs with a batch size of 32. All experiments were conducted with a fixed random seed for reproducibility. 

\section{Detailed Analysis of Experimental Results}
\subsection{Hand Bone Structure Segmentation}
\label{sec:bone_seg_result_AP}
\begin{table*}[!t]
\centering
\caption{DSC for hand bone structure segmentation on all anatomical structures. Bone Mean denotes the average over all bone categories except soft tissue, while All Mean denotes the average over all categories. The best result in each column is highlighted in \textbf{bold}, and the second-best result is \underline{underlined}.}
\label{tab:all_bone_dsc}
\begin{minipage}{\textwidth}
\centering
\resizebox{\textwidth}{!}{%
\setlength{\tabcolsep}{4pt}
\begin{tabular}{lcccccccc}
\toprule
\textbf{Model} & \textbf{Cap} & \textbf{Radius} & \textbf{Ulna} & \textbf{Ham} & \textbf{Lu} & \textbf{Pis\&Tri} & \textbf{Sca} & \textbf{Tm} \\
\midrule
\multicolumn{9}{c}{\textbf{Supervised Models}} \\
\midrule
Unet & 96.26$\pm$7.44 & 98.73$\pm$0.76 & 98.73$\pm$1.21 & 96.44$\pm$2.27 & 95.27$\pm$7.23 & 96.67$\pm$2.96 & 96.70$\pm$3.00 & 96.06$\pm$5.07 \\
Unet++ & 97.13$\pm$2.65 & 98.83$\pm$0.71 & 98.82$\pm$1.22 & 96.81$\pm$1.66 & 95.86$\pm$4.60 & 96.99$\pm$2.41 & 97.13$\pm$3.19 & 96.44$\pm$2.60 \\
SegFormer & 96.28$\pm$3.63 & 98.64$\pm$0.63 & 98.55$\pm$1.18 & 96.07$\pm$1.91 & 95.29$\pm$3.86 & 96.06$\pm$2.32 & 96.10$\pm$3.18 & 95.52$\pm$4.17 \\
TransUnet & \underline{97.19$\pm$2.23} & \textbf{98.92$\pm$0.60} & 98.84$\pm$1.14 & 96.97$\pm$1.47 & \underline{96.33$\pm$4.22} & 97.09$\pm$2.30 & 97.24$\pm$2.41 & 96.48$\pm$2.99 \\
SwinUNETR & 97.10$\pm$3.12 & \textbf{98.92$\pm$0.61} & \textbf{98.87$\pm$1.11} & \underline{97.02$\pm$1.61} & 96.18$\pm$4.14 & 97.19$\pm$1.82 & \underline{97.30$\pm$2.36} & \underline{96.62$\pm$3.01} \\
UMambaEnc & \textbf{97.34$\pm$2.18} & 98.88$\pm$0.63 & \underline{98.85$\pm$1.24} & 96.99$\pm$1.61 & 96.25$\pm$4.02 & \underline{97.20$\pm$2.08} & 97.22$\pm$3.60 & 96.56$\pm$3.42 \\
SwinUMamba & 97.10$\pm$3.02 & \textbf{98.92$\pm$0.59} & \underline{98.85$\pm$1.16} & \textbf{97.14$\pm$1.62} & \textbf{96.42$\pm$4.23} & \textbf{97.23$\pm$1.84} & \textbf{97.38$\pm$2.12} & \textbf{96.74$\pm$2.16} \\
MambaVision & 95.56$\pm$3.82 & 98.16$\pm$0.73 & 98.21$\pm$1.16 & 95.32$\pm$1.92 & 94.52$\pm$3.96 & 95.51$\pm$2.81 & 95.11$\pm$3.36 & 94.98$\pm$3.22 \\
\midrule
\multicolumn{9}{c}{\textbf{Foundation Models}} \\
\midrule
SAM(Box) & 91.50$\pm$4.91 & 92.22$\pm$15.45 & 97.66$\pm$6.41 & 85.50$\pm$7.23 & 85.91$\pm$5.53 & 94.33$\pm$3.63 & 89.79$\pm$4.50 & 81.54$\pm$4.86 \\
SAM(Point) & 75.54$\pm$29.15 & 90.80$\pm$13.66 & 96.15$\pm$12.13 & 69.69$\pm$27.91 & 74.37$\pm$27.16 & 89.03$\pm$20.23 & 73.14$\pm$28.31 & 62.06$\pm$25.67 \\
MedSAM(Box) & 82.08$\pm$7.97 & 83.49$\pm$7.87 & 87.14$\pm$11.83 & 77.22$\pm$7.92 & 76.25$\pm$7.65 & 84.38$\pm$6.76 & 78.11$\pm$8.10 & 81.24$\pm$6.86 \\
\bottomrule
\end{tabular}%
}
\end{minipage}

\vspace{0.6cm}

\begin{minipage}{\textwidth}
\centering
\resizebox{\textwidth}{!}{%
\setlength{\tabcolsep}{4pt}
\begin{tabular}{lcccccccc}
\toprule
\textbf{Model} & \textbf{Td} & \textbf{MC1} & \textbf{MC2} & \textbf{MC3} & \textbf{MC4} & \textbf{MC5} & \textbf{PP1} & \textbf{PP2} \\
\midrule
\multicolumn{9}{c}{\textbf{Supervised Models}} \\
\midrule
Unet & 94.14$\pm$8.22 & 98.34$\pm$2.11 & 97.95$\pm$3.78 & 97.65$\pm$4.52 & 97.93$\pm$2.80 & 98.39$\pm$1.10 & 97.94$\pm$4.93 & 97.61$\pm$4.43 \\
Unet++ & 94.88$\pm$3.51 & 98.68$\pm$0.74 & 98.45$\pm$1.48 & 98.22$\pm$1.40 & 98.35$\pm$1.06 & 98.56$\pm$0.70 & 98.28$\pm$2.30 & 98.58$\pm$1.18 \\
SegFormer & 94.12$\pm$6.29 & 98.35$\pm$0.52 & 98.01$\pm$1.69 & 97.88$\pm$1.36 & 97.93$\pm$1.62 & 98.17$\pm$0.87 & 97.82$\pm$4.13 & 98.27$\pm$1.54 \\
TransUnet & 94.92$\pm$2.86 & 98.69$\pm$0.62 & 98.53$\pm$1.01 & 98.24$\pm$1.43 & 98.31$\pm$1.24 & 98.55$\pm$0.72 & \underline{98.44$\pm$0.97} & \underline{98.70$\pm$0.70} \\
SwinUNETR & \underline{94.96$\pm$3.43} & 98.67$\pm$1.18 & 98.56$\pm$1.08 & \underline{98.26$\pm$1.29} & \textbf{98.44$\pm$0.81} & \underline{98.57$\pm$0.72} & \underline{98.44$\pm$1.30} & \textbf{98.71$\pm$0.86} \\
UMambaEnc & \textbf{95.12$\pm$2.64} & \underline{98.75$\pm$0.56} & \textbf{98.58$\pm$1.03} & 98.23$\pm$1.97 & 98.40$\pm$1.01 & \underline{98.57$\pm$0.76} & 98.33$\pm$2.32 & 98.65$\pm$1.61 \\
SwinUMamba & \underline{94.96$\pm$4.85} & \textbf{98.79$\pm$0.56} & \textbf{98.58$\pm$1.06} & \textbf{98.33$\pm$1.37} & \underline{98.41$\pm$1.19} & \textbf{98.62$\pm$0.84} & \textbf{98.53$\pm$0.76} & 98.68$\pm$1.21 \\
MambaVision & 93.46$\pm$3.02 & 98.15$\pm$0.57 & 97.94$\pm$1.17 & 97.61$\pm$1.22 & 97.74$\pm$0.85 & 97.95$\pm$0.69 & 97.65$\pm$2.90 & 98.24$\pm$1.30 \\
\midrule
\multicolumn{9}{c}{\textbf{Foundation Models}} \\
\midrule
SAM(Box) & 84.77$\pm$6.09 & 97.66$\pm$0.87 & 95.98$\pm$8.10 & 94.77$\pm$7.80 & 93.85$\pm$3.87 & 96.64$\pm$3.09 & 97.27$\pm$1.42 & 97.51$\pm$1.25 \\
SAM(Point) & 49.54$\pm$22.38 & 76.81$\pm$13.64 & 85.68$\pm$16.44 & 90.38$\pm$12.84 & 91.32$\pm$11.65 & 90.15$\pm$17.01 & 72.04$\pm$14.17 & 74.12$\pm$10.69 \\
MedSAM(Box) & 74.93$\pm$12.71 & 89.16$\pm$8.93 & 87.82$\pm$8.22 & 89.52$\pm$7.03 & 84.85$\pm$9.22 & 80.96$\pm$9.60 & 83.31$\pm$8.35 & 89.69$\pm$7.06 \\
\bottomrule
\end{tabular}%
}
\end{minipage}

\vspace{0.6cm}

\begin{minipage}{\textwidth}
\centering
\resizebox{\textwidth}{!}{%
\setlength{\tabcolsep}{4pt}
\begin{tabular}{lcccccccc}
\toprule
\textbf{Model} & \textbf{PP3} & \textbf{PP4} & \textbf{PP5} & \textbf{MP2} & \textbf{MP3} & \textbf{MP4} & \textbf{MP5} & \textbf{DP1} \\
\midrule
\multicolumn{9}{c}{\textbf{Supervised Models}} \\
\midrule
Unet & 97.23$\pm$6.24 & 97.38$\pm$4.31 & 97.94$\pm$2.00 & 96.67$\pm$5.40 & 95.59$\pm$9.54 & 96.19$\pm$6.68 & 96.39$\pm$7.60 & 97.34$\pm$2.54 \\
Unet++ & 98.63$\pm$1.52 & 98.35$\pm$2.61 & 98.34$\pm$0.85 & 98.34$\pm$1.08 & 98.37$\pm$1.22 & 98.22$\pm$2.17 & 97.64$\pm$4.19 & 97.67$\pm$1.08 \\
SegFormer & 98.34$\pm$2.02 & 98.16$\pm$2.09 & 97.95$\pm$1.07 & 97.97$\pm$0.81 & 97.98$\pm$1.53 & 97.95$\pm$0.96 & 96.85$\pm$7.18 & 97.16$\pm$1.03 \\
TransUnet & 98.59$\pm$1.45 & 98.40$\pm$2.21 & 98.28$\pm$0.97 & 98.10$\pm$1.04 & 98.21$\pm$2.07 & 98.22$\pm$1.50 & 97.61$\pm$4.09 & 97.65$\pm$1.07 \\
SwinUNETR & \underline{98.67$\pm$1.30} & 98.42$\pm$1.95 & \underline{98.45$\pm$0.81} & 98.40$\pm$0.90 & 98.26$\pm$1.89 & 98.23$\pm$1.70 & \textbf{97.72$\pm$4.04} & \underline{97.68$\pm$1.40} \\
UMambaEnc & 98.60$\pm$1.67 & \underline{98.47$\pm$1.93} & 98.35$\pm$1.20 & \textbf{98.42$\pm$1.03} & \underline{98.38$\pm$1.95} & \underline{98.38$\pm$1.04} & 97.55$\pm$4.34 & 97.57$\pm$1.24 \\
SwinUMamba & \textbf{98.70$\pm$1.26} & \textbf{98.62$\pm$1.12} & \textbf{98.46$\pm$1.08} & \textbf{98.42$\pm$0.86} & \textbf{98.47$\pm$1.51} & \textbf{98.39$\pm$0.98} & \underline{97.71$\pm$5.19} & \textbf{97.79$\pm$1.24} \\
MambaVision & 98.26$\pm$1.43 & 98.17$\pm$1.42 & 97.86$\pm$1.20 & 97.79$\pm$0.70 & 97.85$\pm$1.53 & 97.80$\pm$1.41 & 96.60$\pm$6.28 & 96.47$\pm$3.08 \\
\midrule
\multicolumn{9}{c}{\textbf{Foundation Models}} \\
\midrule
SAM(Box) & 97.63$\pm$1.61 & 97.59$\pm$1.18 & 97.26$\pm$1.08 & 96.29$\pm$1.64 & 96.50$\pm$1.60 & 96.32$\pm$2.10 & 94.84$\pm$2.62 & 95.96$\pm$1.69 \\
SAM(Point) & 72.25$\pm$9.03 & 71.94$\pm$7.55 & 73.46$\pm$7.42 & 70.98$\pm$21.11 & 68.20$\pm$19.83 & 74.48$\pm$18.72 & 83.29$\pm$19.09 & 58.98$\pm$15.39 \\
MedSAM(Box) & 88.91$\pm$8.07 & 86.74$\pm$8.87 & 80.02$\pm$7.15 & 82.93$\pm$9.01 & 84.97$\pm$8.11 & 83.23$\pm$8.05 & 75.95$\pm$8.81 & 79.31$\pm$7.92 \\
\bottomrule
\end{tabular}%
}
\end{minipage}

\vspace{0.6cm}

\begin{minipage}{\textwidth}
\centering
\resizebox{\textwidth}{!}{%
\setlength{\tabcolsep}{4pt}
\begin{tabular}{lcccccccc}
\toprule
\textbf{Model} & \textbf{DP2} & \textbf{DP3} & \textbf{DP4} & \textbf{DP5} & \textbf{Ses} & \textbf{Soft} & \textbf{Bone Mean} & \textbf{All Mean} \\
\midrule
\multicolumn{9}{c}{\textbf{Supervised Models}} \\
\midrule
Unet & 95.85$\pm$6.05 & 95.50$\pm$5.70 & 96.23$\pm$6.42 & 96.17$\pm$4.47 & 78.26$\pm$13.19 & 99.50$\pm$0.19 & 96.26$\pm$2.77 & 96.37$\pm$2.68 \\
Unet++ & 97.23$\pm$4.86 & 97.33$\pm$2.23 & 97.45$\pm$1.43 & \underline{96.75$\pm$4.28} & 80.47$\pm$12.92 & 99.37$\pm$0.84 & 97.13$\pm$1.40 & 97.21$\pm$1.36 \\
SegFormer & 96.73$\pm$1.29 & 97.01$\pm$1.07 & 96.98$\pm$1.45 & 95.94$\pm$3.56 & 78.32$\pm$12.63 & 99.25$\pm$0.39 & 96.56$\pm$1.53 & 96.65$\pm$1.48 \\
TransUnet & 97.40$\pm$1.38 & 97.40$\pm$2.19 & 97.37$\pm$3.30 & 96.71$\pm$3.63 & 79.82$\pm$12.91 & 99.51$\pm$0.20 & 97.14$\pm$1.19 & 97.22$\pm$1.15 \\
SwinUNETR & \underline{97.52$\pm$1.42} & 97.52$\pm$1.46 & 97.44$\pm$1.55 & 96.43$\pm$5.04 & \textbf{81.62$\pm$13.29} & \textbf{99.52$\pm$0.23} & \textbf{97.25$\pm$1.22} & \textbf{97.32$\pm$1.18} \\
UMambaEnc & 97.49$\pm$1.91 & \underline{97.68$\pm$1.35} & \underline{97.51$\pm$1.30} & 96.45$\pm$6.06 & \underline{81.27$\pm$13.64} & 99.49$\pm$0.34 & \underline{97.24$\pm$1.36} & \textbf{97.32$\pm$1.32} \\
SwinUMamba & \textbf{97.61$\pm$1.73} & \textbf{97.73$\pm$1.25} & \textbf{97.54$\pm$1.22} & \textbf{96.78$\pm$5.37} & 78.82$\pm$13.78 & \textbf{99.52$\pm$0.28} & 97.23$\pm$1.27 & 97.31$\pm$1.23 \\
MambaVision & 96.66$\pm$1.50 & 96.85$\pm$1.53 & 96.66$\pm$1.57 & 95.43$\pm$3.90 & 80.46$\pm$13.75 & 99.38$\pm$0.20 & 96.31$\pm$1.40 & 96.41$\pm$1.35 \\
\midrule
\multicolumn{9}{c}{\textbf{Foundation Models}} \\
\midrule
SAM(Box) & 95.98$\pm$2.44 & 96.25$\pm$1.92 & 96.26$\pm$1.73 & 95.33$\pm$2.17 & 71.73$\pm$31.65 & 17.81$\pm$37.99 & 93.27$\pm$2.28 & 90.76$\pm$2.51 \\
SAM(Point) & 75.66$\pm$21.10 & 77.54$\pm$20.23 & 79.55$\pm$20.46 & 84.29$\pm$20.90 & 65.01$\pm$36.68 & 47.05$\pm$44.90 & 76.43$\pm$9.35 & 75.45$\pm$9.21 \\
MedSAM(Box) & 75.21$\pm$11.00 & 78.58$\pm$9.05 & 77.96$\pm$9.14 & 70.51$\pm$10.12 & 58.00$\pm$28.70 & 65.70$\pm$6.48 & 81.12$\pm$4.58 & 80.61$\pm$4.42 \\
\bottomrule
\end{tabular}%
}
\end{minipage}
\end{table*}

\begin{table*}[!t]
\centering
\caption{NSD for bone segmentation on all hand bone structures. Bone Mean denotes the average over all bone categories except soft tissue, while All Mean denotes the average over all categories. The best result in each column is highlighted in \textbf{bold}, and the second-best result is \underline{underlined}.}
\label{tab:all_bone_nsd}
\begin{minipage}{\textwidth}
\centering
\resizebox{\textwidth}{!}{%
\setlength{\tabcolsep}{3pt}
\begin{tabular}{lcccccccc}
\toprule
\textbf{Model} & \textbf{Cap} & \textbf{Radius} & \textbf{Ulna} & \textbf{Ham} & \textbf{Lu} & \textbf{Pis\&Tri} & \textbf{Sca} & \textbf{Tm} \\
\midrule
\multicolumn{9}{c}{\textbf{Supervised Models}} \\
\midrule
Unet & 81.38$\pm$14.19 & 90.71$\pm$6.63 & 97.76$\pm$4.47 & 81.61$\pm$13.33 & 78.77$\pm$15.69 & 86.98$\pm$12.57 & 85.00$\pm$14.02 & 80.69$\pm$12.47 \\
Unet++ & 84.55$\pm$13.38 & 91.89$\pm$6.24 & 97.98$\pm$4.54 & 83.93$\pm$11.48 & 80.59$\pm$15.40 & 88.48$\pm$11.75 & 87.88$\pm$12.96 & 82.06$\pm$12.25 \\
SegFormer & 77.55$\pm$14.00 & 89.75$\pm$6.36 & 97.14$\pm$4.70 & 77.29$\pm$12.34 & 75.89$\pm$16.68 & 81.11$\pm$13.82 & 79.59$\pm$14.16 & 74.84$\pm$11.95 \\
TransUnet & \underline{84.75$\pm$12.57} & \textbf{93.10$\pm$5.50} & \underline{98.39$\pm$3.86} & 84.99$\pm$11.18 & \underline{83.46$\pm$14.87} & \underline{89.50$\pm$11.06} & 88.55$\pm$12.91 & 82.59$\pm$11.97 \\
SwinUNETR & 84.60$\pm$13.52 & 92.89$\pm$5.80 & 98.38$\pm$3.87 & \underline{85.55$\pm$11.64} & 81.82$\pm$15.19 & 89.00$\pm$11.46 & 88.66$\pm$12.50 & \underline{83.74$\pm$12.18} \\
UMambaEnc & \textbf{85.83$\pm$12.88} & 92.67$\pm$5.86 & \underline{98.39$\pm$4.14} & 85.10$\pm$11.88 & 82.73$\pm$15.71 & 89.44$\pm$11.75 & \underline{88.91$\pm$11.83} & 83.35$\pm$11.94 \\
SwinUMamba & 84.46$\pm$12.91 & \underline{92.94$\pm$5.58} & \textbf{98.49$\pm$3.81} & \textbf{86.72$\pm$11.58} & \textbf{83.49$\pm$15.02} & \textbf{89.68$\pm$10.93} & \textbf{89.30$\pm$13.00} & \textbf{84.10$\pm$11.85} \\
MambaVision & 69.01$\pm$14.41 & 83.87$\pm$7.17 & 94.30$\pm$5.85 & 69.39$\pm$12.86 & 70.48$\pm$18.40 & 78.00$\pm$13.55 & 71.50$\pm$16.87 & 69.95$\pm$11.81 \\
\midrule
\multicolumn{9}{c}{\textbf{Foundation Models}} \\
\midrule
SAM(Box) & 60.98$\pm$15.95 & 75.02$\pm$11.28 & 95.95$\pm$9.85 & 47.07$\pm$15.69 & 60.51$\pm$13.81 & 75.70$\pm$14.67 & 66.29$\pm$14.32 & 47.16$\pm$10.43 \\
SAM(Point) & 47.26$\pm$28.07 & 69.53$\pm$17.68 & 93.59$\pm$15.42 & 35.58$\pm$22.76 & 53.46$\pm$23.64 & 71.51$\pm$22.38 & 51.34$\pm$27.66 & 32.88$\pm$19.14 \\
MedSAM(Box) & 24.74$\pm$12.82 & 29.13$\pm$13.37 & 44.97$\pm$20.50 & 22.28$\pm$11.13 & 24.38$\pm$11.36 & 32.74$\pm$14.74 & 26.20$\pm$12.34 & 29.69$\pm$11.66 \\
\bottomrule
\end{tabular}%
}
\end{minipage}

\vspace{0.6cm}

\begin{minipage}{\textwidth}
\centering
\resizebox{\textwidth}{!}{%
\setlength{\tabcolsep}{3pt}
\begin{tabular}{lcccccccc}
\toprule
\textbf{Model} & \textbf{Td} & \textbf{MC1} & \textbf{MC2} & \textbf{MC3} & \textbf{MC4} & \textbf{MC5} & \textbf{PP1} & \textbf{PP2} \\
\midrule
\multicolumn{9}{c}{\textbf{Supervised Models}} \\
\midrule
Unet & 74.96$\pm$16.05 & 95.33$\pm$6.97 & 94.54$\pm$8.74 & 91.81$\pm$8.88 & 95.26$\pm$7.17 & 96.84$\pm$4.57 & 95.40$\pm$7.75 & 95.38$\pm$10.68 \\
Unet++ & 76.87$\pm$15.48 & 96.38$\pm$4.78 & 95.74$\pm$5.18 & 93.29$\pm$5.53 & 96.25$\pm$4.95 & 97.52$\pm$3.29 & 96.05$\pm$6.24 & 97.43$\pm$5.86 \\
SegFormer & 72.03$\pm$16.24 & 94.57$\pm$4.20 & 93.84$\pm$5.11 & 91.31$\pm$6.35 & 94.00$\pm$6.64 & 94.94$\pm$4.85 & 94.88$\pm$6.99 & 97.08$\pm$5.93 \\
TransUnet & 76.23$\pm$15.25 & 96.64$\pm$4.11 & 96.19$\pm$4.35 & \underline{93.82$\pm$5.45} & 96.37$\pm$5.14 & \underline{97.71$\pm$3.53} & 96.90$\pm$4.68 & 98.09$\pm$3.86 \\
SwinUNETR & 76.98$\pm$15.44 & 96.74$\pm$4.57 & 96.19$\pm$4.83 & 93.70$\pm$5.46 & 96.71$\pm$4.37 & 97.69$\pm$3.28 & \underline{96.98$\pm$4.70} & 98.09$\pm$4.30 \\
UMambaEnc & \underline{77.64$\pm$15.51} & \underline{97.13$\pm$3.82} & \underline{96.26$\pm$4.77} & 93.68$\pm$5.80 & \underline{96.72$\pm$4.63} & 97.57$\pm$3.44 & 96.68$\pm$6.03 & \textbf{98.26$\pm$4.51} \\
SwinUMamba & \textbf{78.00$\pm$15.32} & \textbf{97.34$\pm$3.45} & \textbf{96.41$\pm$4.76} & \textbf{94.28$\pm$5.38} & \textbf{96.95$\pm$4.34} & \textbf{98.08$\pm$3.24} & \textbf{97.22$\pm$4.28} & \underline{98.22$\pm$4.64} \\
MambaVision & 64.94$\pm$15.64 & 91.76$\pm$5.13 & 91.75$\pm$5.29 & 89.26$\pm$6.41 & 91.91$\pm$6.30 & 92.77$\pm$5.16 & 91.58$\pm$6.67 & 96.12$\pm$5.77 \\
\midrule
\multicolumn{9}{c}{\textbf{Foundation Models}} \\
\midrule
SAM(Box) & 41.82$\pm$17.90 & 90.26$\pm$5.22 & 86.01$\pm$7.75 & 79.02$\pm$5.95 & 79.05$\pm$7.91 & 86.80$\pm$8.63 & 89.70$\pm$6.87 & 91.12$\pm$5.83 \\
SAM(Point) & 19.96$\pm$14.74 & 59.40$\pm$15.70 & 69.88$\pm$18.72 & 71.82$\pm$15.08 & 76.01$\pm$13.92 & 77.15$\pm$20.69 & 57.32$\pm$14.45 & 60.74$\pm$10.48 \\
MedSAM(Box) & 24.99$\pm$13.70 & 41.61$\pm$16.76 & 42.74$\pm$16.23 & 50.46$\pm$17.44 & 39.53$\pm$17.58 & 29.32$\pm$14.70 & 31.42$\pm$16.93 & 49.56$\pm$21.37 \\
\bottomrule
\end{tabular}%
}
\end{minipage}

\vspace{0.6cm}

\begin{minipage}{\textwidth}
\centering
\resizebox{\textwidth}{!}{%
\setlength{\tabcolsep}{3pt}
\begin{tabular}{lcccccccc}
\toprule
\textbf{Model} & \textbf{PP3} & \textbf{PP4} & \textbf{PP5} & \textbf{MP2} & \textbf{MP3} & \textbf{MP4} & \textbf{MP5} & \textbf{DP1} \\
\midrule
\multicolumn{9}{c}{\textbf{Supervised Models}} \\
\midrule
Unet & 94.78$\pm$12.30 & 94.75$\pm$9.85 & 96.40$\pm$6.21 & 94.43$\pm$9.31 & 91.39$\pm$13.89 & 92.44$\pm$12.53 & 94.15$\pm$10.63 & 94.40$\pm$6.87 \\
Unet++ & 98.18$\pm$4.91 & 97.99$\pm$4.74 & 98.09$\pm$3.17 & 97.97$\pm$4.03 & 97.27$\pm$5.47 & 97.92$\pm$5.01 & \underline{96.91$\pm$7.67} & 95.35$\pm$5.45 \\
SegFormer & 97.00$\pm$6.43 & 97.34$\pm$4.94 & 96.27$\pm$4.33 & 97.48$\pm$3.89 & 96.22$\pm$5.89 & 96.70$\pm$4.72 & 94.87$\pm$9.66 & 94.26$\pm$5.74 \\
TransUnet & 98.03$\pm$4.70 & 98.03$\pm$3.96 & 97.63$\pm$3.13 & 96.47$\pm$5.02 & 96.74$\pm$5.43 & 97.43$\pm$5.02 & 96.55$\pm$7.39 & 95.25$\pm$5.43 \\
SwinUNETR & 98.14$\pm$4.73 & 98.04$\pm$4.28 & \underline{98.29$\pm$3.33} & 98.26$\pm$3.31 & 97.17$\pm$5.41 & 97.64$\pm$4.99 & \underline{96.91$\pm$6.65} & \underline{95.64$\pm$5.68} \\
UMambaEnc & \underline{98.26$\pm$4.81} & \underline{98.39$\pm$3.92} & 98.13$\pm$3.22 & \textbf{98.60$\pm$3.74} & \underline{97.82$\pm$5.28} & \underline{97.95$\pm$5.17} & 96.61$\pm$7.47 & 95.11$\pm$5.56 \\
SwinUMamba & \textbf{98.38$\pm$4.53} & \textbf{98.48$\pm$3.84} & \textbf{98.58$\pm$3.38} & \underline{98.42$\pm$3.18} & \textbf{97.96$\pm$4.30} & \textbf{98.37$\pm$3.75} & \textbf{97.25$\pm$7.84} & \textbf{95.88$\pm$5.18} \\
MambaVision & 96.47$\pm$5.67 & 96.76$\pm$5.27 & 95.52$\pm$4.90 & 95.44$\pm$4.97 & 94.87$\pm$6.37 & 95.64$\pm$5.97 & 91.79$\pm$9.51 & 90.88$\pm$8.15 \\
\midrule
\multicolumn{9}{c}{\textbf{Foundation Models}} \\
\midrule
SAM(Box) & 91.61$\pm$7.10 & 92.41$\pm$6.16 & 91.79$\pm$7.08 & 87.18$\pm$7.18 & 85.91$\pm$6.73 & 86.59$\pm$7.30 & 83.99$\pm$8.42 & 89.98$\pm$5.84 \\
SAM(Point) & 59.93$\pm$9.17 & 60.20$\pm$8.60 & 61.81$\pm$7.81 & 53.00$\pm$20.49 & 48.63$\pm$18.56 & 56.96$\pm$19.69 & 69.91$\pm$21.77 & 50.14$\pm$13.53 \\
MedSAM(Box) & 48.09$\pm$22.10 & 42.77$\pm$22.58 & 27.67$\pm$12.66 & 38.34$\pm$17.33 & 39.05$\pm$18.09 & 34.06$\pm$16.35 & 25.99$\pm$11.22 & 28.89$\pm$11.45 \\
\bottomrule
\end{tabular}%
}
\end{minipage}

\vspace{0.6cm}

\begin{minipage}{\textwidth}
\centering
\resizebox{\textwidth}{!}{%
\setlength{\tabcolsep}{3pt}
\begin{tabular}{lcccccccc}
\toprule
\textbf{Model} & \textbf{DP2} & \textbf{DP3} & \textbf{DP4} & \textbf{DP5} & \textbf{Ses} & \textbf{Soft} & \textbf{Bone Mean} & \textbf{All Mean} \\
\midrule
\multicolumn{9}{c}{\textbf{Supervised Models}} \\
\midrule
Unet & 93.30$\pm$11.04 & 91.92$\pm$12.15 & 93.45$\pm$10.60 & 95.27$\pm$8.49 & 70.37$\pm$16.41 & 92.89$\pm$6.95 & 90.33$\pm$6.02 & 90.41$\pm$5.82 \\
Unet++ & 96.45$\pm$7.72 & 95.81$\pm$6.73 & 96.16$\pm$5.95 & \underline{96.53$\pm$7.66} & 73.10$\pm$15.55 & 91.44$\pm$9.45 & 92.57$\pm$4.19 & 92.54$\pm$4.05 \\
SegFormer & 95.77$\pm$6.13 & 95.26$\pm$6.33 & 95.35$\pm$6.64 & 94.53$\pm$8.95 & 68.43$\pm$16.37 & 90.16$\pm$6.60 & 89.84$\pm$4.55 & 89.85$\pm$4.40 \\
TransUnet & 96.55$\pm$6.15 & 96.62$\pm$5.65 & 96.23$\pm$6.06 & 96.52$\pm$7.22 & 73.38$\pm$15.61 & 93.28$\pm$7.50 & 92.85$\pm$4.02 & 92.87$\pm$3.88 \\
SwinUNETR & 96.97$\pm$5.91 & 96.35$\pm$6.20 & 96.38$\pm$5.39 & 96.13$\pm$8.13 & \underline{74.75$\pm$16.67} & \underline{93.72$\pm$7.01} & 93.05$\pm$4.10 & 93.07$\pm$3.96 \\
UMambaEnc & \underline{97.05$\pm$5.53} & \underline{96.83$\pm$5.44} & \underline{96.41$\pm$5.63} & 95.86$\pm$8.52 & \textbf{75.91$\pm$15.99} & 92.96$\pm$7.69 & \underline{93.22$\pm$4.22} & \underline{93.21$\pm$4.08} \\
SwinUMamba & \textbf{97.16$\pm$5.69} & \textbf{96.98$\pm$5.24} & \textbf{96.48$\pm$5.59} & \textbf{96.89$\pm$7.44} & 72.70$\pm$16.16 & \textbf{94.06$\pm$7.08} & \textbf{93.42$\pm$3.99} & \textbf{93.44$\pm$3.86} \\
MambaVision & 94.62$\pm$7.59 & 94.41$\pm$7.30 & 93.66$\pm$7.35 & 92.58$\pm$8.77 & 74.48$\pm$17.63 & 89.43$\pm$7.54 & 87.02$\pm$4.63 & 87.10$\pm$4.48 \\
\midrule
\multicolumn{9}{c}{\textbf{Foundation Models}} \\
\midrule
SAM(Box) & 92.75$\pm$8.24 & 91.76$\pm$7.70 & 92.04$\pm$7.73 & 92.29$\pm$7.43 & 68.89$\pm$32.58 & 20.55$\pm$30.86 & 79.99$\pm$4.87 & 78.01$\pm$4.72 \\
SAM(Point) & 60.96$\pm$21.75 & 61.51$\pm$22.31 & 67.32$\pm$22.91 & 80.32$\pm$22.82 & 62.55$\pm$36.80 & 34.74$\pm$42.51 & 60.02$\pm$8.92 & 59.18$\pm$8.81 \\
MedSAM(Box) & 35.23$\pm$14.68 & 32.85$\pm$13.87 & 34.83$\pm$13.47 & 27.63$\pm$11.21 & 37.42$\pm$23.10 & 9.13$\pm$5.13 & 34.36$\pm$8.69 & 33.52$\pm$8.37 \\
\bottomrule
\end{tabular}%
}
\par\smallskip
\end{minipage}
\end{table*}

\begin{figure}[!t]
    \centering
    \includegraphics[width=\textwidth]{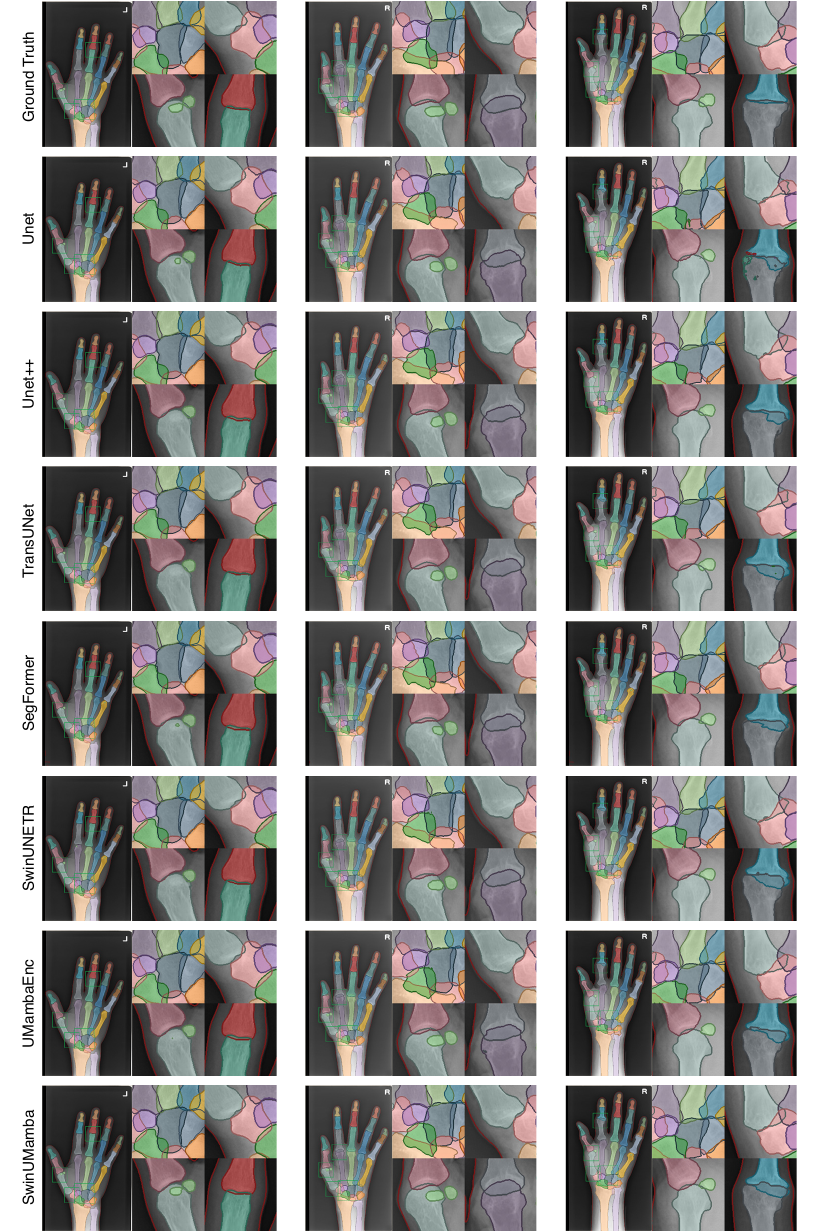}
    \caption{Hand bone structure segmentation results (A).}
    \label{fig:bone_seg_result1_AP}
\end{figure}

\begin{figure}[!t]
    \centering
    \includegraphics[width=\textwidth]{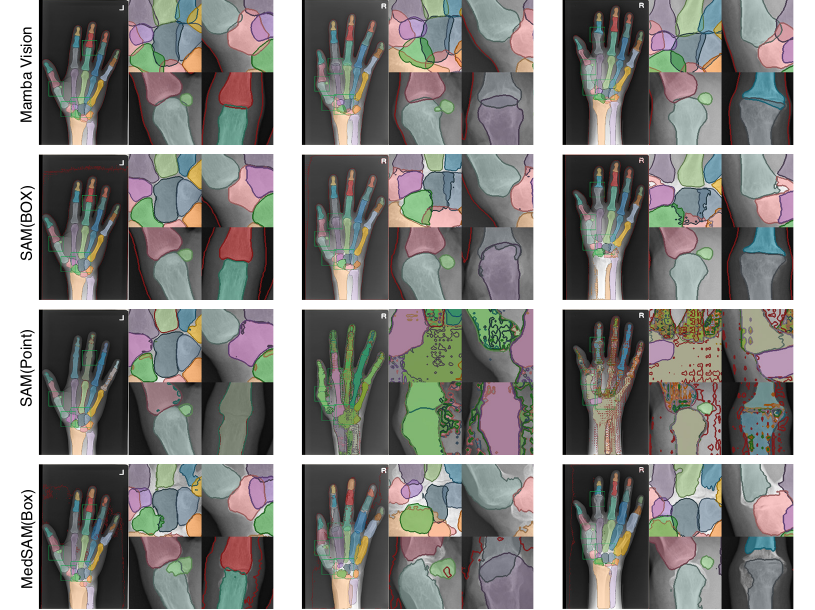}
    \caption{Hand bone structure segmentation results (B).}
    \label{fig:bone_seg_result2_AP}
\end{figure}

\subsubsection{Overall and Bone-wise Results}

Quantitative results for hand bone structure segmentation are summarized in Table~\ref{tab:all_bone_dsc} and Table~\ref{tab:all_bone_nsd}, with qualitative comparisons shown in Fig.~\ref{fig:bone_seg_result1_AP} and Fig.~\ref{fig:bone_seg_result2_AP}. Overall, supervised models achieve strong segmentation performance across most anatomical structures. In Table~\ref{tab:all_bone_dsc}, SwinUNETR obtains the highest Bone Mean DSC of 97.25\%, followed closely by UMambaEnc with 97.24\% and SwinUMamba with 97.23\%. The corresponding All Mean DSC values are also highly similar, with SwinUNETR and UMambaEnc both reaching 97.32\%, and SwinUMamba reaching 97.31\%. These small differences indicate that the general whole-bone segmentation task is already relatively well handled by supervised models.

The bone-wise results show that large and clearly visible bones are segmented with particularly high accuracy. The radius and ulna reach approximately 98--99\% DSC for the best supervised models, and most metacarpal, proximal phalangeal, middle phalangeal, and distal phalangeal bones also achieve DSC values around or above 98\%. In contrast, smaller or anatomically crowded structures remain more difficult. The sesamoid bones show the lowest DSC among all categories, with the best result only reaching 81.62\% for SwinUNETR. Several carpal bones, including the trapezoid, lunate, trapezium, scaphoid, and pisiform/triquetrum, also show lower performance than long bones, reflecting the effect of small structure size, low contrast, and dense anatomical overlap around the wrist.

The NSD results in Table~\ref{tab:all_bone_nsd} make the boundary-level differences more apparent. SwinUMamba achieves the best Bone Mean NSD of 93.42\% and All Mean NSD of 93.44\%, while UMambaEnc ranks second with 93.22\% Bone Mean NSD and 93.21\% All Mean NSD. This suggests that although SwinUNETR slightly leads in Bone Mean DSC, SwinUMamba provides more accurate boundary alignment overall. The advantage is especially meaningful for small bones and joint-adjacent structures, where minor contour shifts can substantially affect anatomical consistency.

Qualitative results in Fig.~\ref{fig:bone_seg_result1_AP} and Fig.~\ref{fig:bone_seg_result2_AP} support these quantitative findings. Supervised models generally preserve the global hand skeleton and produce coherent masks for large bones and phalanges. However, visible differences appear around the wrist and metacarpal bases, where boundaries between adjacent bones are weak or partially overlapped. Models with stronger NSD performance produce smoother and more anatomically consistent contours, whereas weaker models show more fragmented masks, boundary leakage, or missing small structures. Foundation models are less reliable in these qualitative examples, especially in Fig.~\ref{fig:bone_seg_result2_AP}, where predictions often become coarse, fragmented, or misaligned in small and overlapping bones.

\begin{table*}[!t]
\centering
\caption{Overlap DSC performance on overlapping regions. The best results in each column are highlighted in \textbf{bold}, and the second-best values are \underline{underlined}.}
\label{tab:overlap_dsc}
\begin{minipage}{\textwidth}
\centering
\resizebox{\textwidth}{!}{%
\setlength{\tabcolsep}{8pt}
\begin{tabular}{lcccccccc}
\toprule
\textbf{Model} & \textbf{Cap-Sca} & \textbf{Cap-Td} & \textbf{Cap-MC3} & \textbf{Radius-Lu} & \textbf{Radius-Sca} & \textbf{Ham-MC4} & \textbf{Ham-MC5} \\
\midrule
\multicolumn{8}{c}{\textbf{Supervised Models}} \\
\midrule
Unet        & 87.31$\pm$10.90 & 62.96$\pm$23.94 & 49.57$\pm$24.83 & 85.80$\pm$12.88 & 83.25$\pm$11.98 & 64.95$\pm$22.20 & 87.20$\pm$6.99 \\
Unet++      & 88.70$\pm$10.91 & 64.52$\pm$23.23 & 55.41$\pm$24.21 & 87.75$\pm$12.80 & 84.54$\pm$12.22 & 65.31$\pm$22.37 & 87.51$\pm$6.39 \\
SegFormer   & 85.33$\pm$10.83 & 61.65$\pm$22.59 & 48.49$\pm$24.80 & 84.53$\pm$12.99 & 81.21$\pm$12.07 & 61.57$\pm$21.51 & 84.73$\pm$6.95 \\
TransUnet   & 89.25$\pm$10.80 & \underline{65.64$\pm$23.37} & \textbf{57.34$\pm$23.73} & \underline{88.20$\pm$11.62} & 85.41$\pm$11.17 & 64.71$\pm$22.23 & 88.11$\pm$6.34 \\
SwinUNETR   & 89.26$\pm$10.28 & 64.81$\pm$24.40 & \underline{57.03$\pm$24.18} & \textbf{88.24$\pm$12.55} & 85.32$\pm$11.50 & \underline{65.73$\pm$22.66} & \underline{88.50$\pm$6.28} \\
UMambaEnc   & \underline{89.37$\pm$10.27} & \textbf{65.69$\pm$24.54} & 56.35$\pm$23.43 & 88.17$\pm$11.95 & \underline{85.67$\pm$10.97} & \textbf{67.21$\pm$21.34} & 87.89$\pm$7.47 \\
SwinUMamba  & \textbf{89.55$\pm$10.82} & 64.60$\pm$22.95 & 54.63$\pm$24.60 & 87.24$\pm$11.90 & \textbf{86.19$\pm$10.82} & 63.46$\pm$22.96 & \textbf{89.10$\pm$5.91} \\
MambaVision & 81.15$\pm$11.35 & 57.98$\pm$24.41 & 41.88$\pm$21.92 & 76.29$\pm$13.03 & 74.88$\pm$11.55 & 53.89$\pm$19.36 & 81.73$\pm$7.68 \\
\midrule
\multicolumn{8}{c}{\textbf{Foundation Models}} \\
\midrule
SAM(Box)    & 8.08$\pm$21.36 & 1.06$\pm$6.35 & 0.00$\pm$0.00 & 0.03$\pm$0.46 & 0.89$\pm$5.02 & 0.00$\pm$0.03 & 1.38$\pm$6.62 \\
SAM(Point)  & 2.41$\pm$6.57 & 0.54$\pm$1.85 & 0.12$\pm$0.80 & 0.11$\pm$0.52 & 1.20$\pm$4.48 & 0.21$\pm$1.58 & 2.92$\pm$8.72 \\
MedSAM(Box) & 13.53$\pm$25.04 & 2.50$\pm$9.72 & 0.88$\pm$4.46 & 32.57$\pm$24.80 & 24.45$\pm$22.48 & 0.10$\pm$1.16 & 2.60$\pm$9.52 \\
\bottomrule
\end{tabular}}
\end{minipage}

\vspace{0.6cm}

\begin{minipage}{\textwidth}
\centering
\resizebox{\textwidth}{!}{%
\setlength{\tabcolsep}{8pt}
\begin{tabular}{lcccccccc}
\toprule
\textbf{Model} & \textbf{Lu-Sca} & \textbf{Sca-Tm} & \textbf{Tm-Td} & \textbf{Tm-MC1} & \textbf{Tm-MC2} & \textbf{Td-MC2} & \textbf{MC2-MC3} \\
\midrule
\multicolumn{8}{c}{\textbf{Supervised Models}} \\
\midrule
Unet        & 72.15$\pm$20.51 & 72.49$\pm$22.39 & 90.27$\pm$9.04 & 75.92$\pm$17.86 & 81.30$\pm$13.94 & 40.83$\pm$23.72 & 71.15$\pm$15.33 \\
Unet++      & 73.04$\pm$21.30 & 75.95$\pm$19.06 & 90.32$\pm$8.79 & 77.96$\pm$17.93 & 82.67$\pm$13.42 & 41.30$\pm$23.68 & 73.39$\pm$15.34 \\
SegFormer   & 69.51$\pm$21.94 & 71.14$\pm$20.24 & 89.26$\pm$7.54 & 74.61$\pm$16.82 & 78.26$\pm$14.18 & 36.47$\pm$22.59 & 68.50$\pm$15.32 \\
TransUnet   & \underline{75.13$\pm$20.68} & 76.08$\pm$19.16 & \underline{90.97$\pm$6.85} & 78.40$\pm$16.34 & 83.03$\pm$12.63 & 42.01$\pm$23.34 & \underline{74.59$\pm$15.48} \\
SwinUNETR   & 74.61$\pm$20.01 & \textbf{77.36$\pm$19.68} & 90.95$\pm$8.88 & 78.20$\pm$18.11 & 83.43$\pm$13.22 & \underline{43.09$\pm$24.21} & 74.53$\pm$14.63 \\
UMambaEnc   & 74.18$\pm$20.32 & \underline{76.50$\pm$19.72} & \underline{90.97$\pm$7.75} & \underline{78.62$\pm$17.59} & \underline{83.59$\pm$11.86} & 42.21$\pm$23.77 & 73.54$\pm$15.22 \\
SwinUMamba  & \textbf{75.51$\pm$21.41} & 75.36$\pm$19.76 & \textbf{91.29$\pm$7.39} & \textbf{79.19$\pm$16.90} & \textbf{84.30$\pm$12.25} & \textbf{46.11$\pm$25.04} & \textbf{75.22$\pm$15.12} \\
MambaVision & 62.40$\pm$23.29 & 66.34$\pm$20.10 & 87.66$\pm$8.88 & 65.21$\pm$18.48 & 76.81$\pm$13.97 & 33.08$\pm$22.12 & 63.01$\pm$14.20 \\
\midrule
\multicolumn{8}{c}{\textbf{Foundation Models}} \\
\midrule
SAM(Box)    & 7.01$\pm$16.36 & 1.60$\pm$7.68 & 52.86$\pm$14.47 & 1.00$\pm$6.49 & 5.14$\pm$10.60 & 1.72$\pm$5.63 & 0.00$\pm$0.00 \\
SAM(Point)  & 1.80$\pm$4.36 & 0.99$\pm$2.84 & 27.22$\pm$14.68 & 0.62$\pm$3.51 & 6.93$\pm$10.89 & 1.34$\pm$3.21 & 0.01$\pm$0.06 \\
MedSAM(Box) & 3.29$\pm$11.78 & 1.87$\pm$8.95 & 49.71$\pm$29.48 & 7.20$\pm$12.84 & 4.61$\pm$12.89 & 2.73$\pm$7.15 & 0.49$\pm$3.74 \\
\bottomrule
\end{tabular}}
\par\smallskip
\end{minipage}
\end{table*}

\begin{table*}[!t]
\centering
\caption{Overlap NSD performance on overlapping regions. The best results in each column are highlighted in \textbf{bold}, and the second-best values are \underline{underlined}.}
\label{tab:overlap_nsd}
\begin{minipage}{\textwidth}
\centering
\resizebox{\textwidth}{!}{%
\setlength{\tabcolsep}{8pt}
\begin{tabular}{lcccccccc}
\toprule
\textbf{Model} & \textbf{Cap-Sca} & \textbf{Cap-Td} & \textbf{Cap-MC3} & \textbf{Radius-Lu} & \textbf{Radius-Sca} & \textbf{Ham-MC4} & \textbf{Ham-MC5} \\
\midrule
\multicolumn{8}{c}{\textbf{Supervised Models}} \\
\midrule
Unet        & 80.24$\pm$17.63 & 70.53$\pm$26.21 & 67.18$\pm$25.00 & 77.34$\pm$19.49 & 77.69$\pm$18.74 & 70.97$\pm$27.15 & 80.27$\pm$15.83 \\
Unet++      & 84.39$\pm$16.05 & 73.21$\pm$24.91 & 70.79$\pm$24.41 & 81.41$\pm$18.85 & 79.38$\pm$17.31 & 72.04$\pm$26.62 & 80.99$\pm$15.38 \\
SegFormer   & 74.55$\pm$18.24 & 67.72$\pm$24.68 & 67.21$\pm$23.89 & 74.76$\pm$18.86 & 74.06$\pm$17.29 & 67.20$\pm$25.79 & 73.83$\pm$16.81 \\
TransUnet   & \underline{85.34$\pm$16.45} & 73.18$\pm$26.09 & 71.51$\pm$24.29 & 82.21$\pm$18.28 & 82.00$\pm$17.99 & 72.55$\pm$25.98 & 82.81$\pm$15.13 \\
SwinUNETR   & 84.87$\pm$16.83 & 72.91$\pm$25.43 & 71.21$\pm$25.32 & \textbf{82.52$\pm$18.78} & 81.81$\pm$16.53 & \underline{72.70$\pm$27.02} & \underline{83.79$\pm$14.87} \\
UMambaEnc   & 85.16$\pm$16.02 & \textbf{74.25$\pm$25.93} & \textbf{72.26$\pm$24.39} & \underline{82.48$\pm$18.29} & \underline{81.82$\pm$16.56} & \textbf{73.65$\pm$26.93} & 82.26$\pm$16.08 \\
SwinUMamba  & \textbf{86.21$\pm$16.43} & \underline{73.50$\pm$25.03} & \underline{71.86$\pm$24.44} & 78.72$\pm$19.22 & \textbf{82.51$\pm$18.71} & 72.11$\pm$27.05 & \textbf{85.91$\pm$14.02} \\
MambaVision & 63.42$\pm$19.83 & 63.95$\pm$26.13 & 60.39$\pm$24.54 & 59.27$\pm$19.73 & 63.08$\pm$19.05 & 58.41$\pm$25.32 & 64.87$\pm$18.52 \\
\midrule
\multicolumn{8}{c}{\textbf{Foundation Models}} \\
\midrule
SAM(Box)    & 6.67$\pm$15.26 & 1.71$\pm$7.70 & 0.02$\pm$0.34 & 0.07$\pm$1.20 & 1.53$\pm$4.89 & 0.00$\pm$0.00 & 2.78$\pm$9.44 \\
SAM(Point)  & 3.12$\pm$8.42 & 1.69$\pm$4.55 & 0.80$\pm$3.34 & 0.25$\pm$1.30 & 2.47$\pm$6.44 & 0.47$\pm$2.06 & 3.86$\pm$8.73 \\
MedSAM(Box) & 11.54$\pm$15.79 & 5.59$\pm$13.10 & 1.87$\pm$7.78 & 16.85$\pm$12.86 & 16.04$\pm$12.69 & 0.53$\pm$3.57 & 3.55$\pm$9.03 \\
\bottomrule
\end{tabular}}
\end{minipage}

\vspace{0.6cm}

\begin{minipage}{\textwidth}
\centering
\resizebox{\textwidth}{!}{%
\setlength{\tabcolsep}{8pt}
\begin{tabular}{lcccccccc}
\toprule
\textbf{Model} & \textbf{Lu-Sca} & \textbf{Sca-Tm} & \textbf{Tm-Td} & \textbf{Tm-MC1} & \textbf{Tm-MC2} & \textbf{Td-MC2} & \textbf{MC2-MC3} \\
\midrule
\multicolumn{8}{c}{\textbf{Supervised Models}} \\
\midrule
Unet        & 66.02$\pm$26.99 & 72.09$\pm$26.71 & 74.54$\pm$17.65 & 79.94$\pm$20.52 & 69.66$\pm$21.88 & 59.20$\pm$31.21 & 75.55$\pm$19.23 \\
Unet++      & 68.97$\pm$27.56 & 77.07$\pm$23.58 & 74.89$\pm$17.54 & 82.66$\pm$20.64 & 71.85$\pm$21.36 & 59.95$\pm$32.04 & 79.90$\pm$18.33 \\
SegFormer   & 62.74$\pm$26.44 & 70.72$\pm$24.50 & 67.47$\pm$17.33 & 77.10$\pm$19.72 & 61.76$\pm$20.63 & 57.24$\pm$31.40 & 73.87$\pm$17.97 \\
TransUnet   & \underline{72.75$\pm$26.21} & 75.34$\pm$24.99 & 75.67$\pm$17.09 & \underline{83.59$\pm$18.55} & 73.04$\pm$20.46 & 59.65$\pm$31.42 & 80.80$\pm$18.09 \\
SwinUNETR   & 70.66$\pm$26.42 & \textbf{78.71$\pm$23.34} & \underline{77.04$\pm$17.76} & 83.14$\pm$19.68 & 74.59$\pm$21.53 & \underline{61.43$\pm$31.32} & \underline{80.93$\pm$18.75} \\
UMambaEnc   & 71.64$\pm$26.16 & \underline{78.40$\pm$23.95} & 76.15$\pm$17.08 & \textbf{83.61$\pm$19.53} & \underline{74.61$\pm$20.17} & 60.61$\pm$31.77 & 79.97$\pm$18.58 \\
SwinUMamba  & \textbf{73.91$\pm$25.96} & 78.10$\pm$23.87 & \textbf{77.54$\pm$16.93} & \textbf{84.77$\pm$18.78} & \textbf{76.82$\pm$20.65} & \textbf{61.82$\pm$32.34} & \textbf{81.75$\pm$19.06} \\
MambaVision & 54.59$\pm$25.07 & 60.66$\pm$25.57 & 60.97$\pm$16.95 & 64.36$\pm$20.78 & 56.63$\pm$21.64 & 51.65$\pm$30.93 & 68.36$\pm$19.40 \\
\midrule
\multicolumn{8}{c}{\textbf{Foundation Models}} \\
\midrule
SAM(Box)    & 8.20$\pm$14.51 & 2.13$\pm$7.78 & 23.74$\pm$13.54 & 1.30$\pm$6.46 & 6.95$\pm$12.42 & 3.77$\pm$10.89 & 0.00$\pm$0.00 \\
SAM(Point)  & 3.73$\pm$7.47 & 2.81$\pm$6.82 & 7.19$\pm$7.57 & 1.22$\pm$5.38 & 7.64$\pm$11.34 & 4.81$\pm$8.77 & 0.05$\pm$0.46 \\
MedSAM(Box) & 2.63$\pm$8.02 & 2.81$\pm$8.43 & 22.04$\pm$16.66 & 11.03$\pm$14.27 & 4.59$\pm$10.24 & 6.18$\pm$12.02 & 1.16$\pm$6.40 \\
\bottomrule
\end{tabular}}
\par\smallskip
\end{minipage}
\end{table*}

\subsubsection{Segmentation of Overlapping Regions}

Quantitative results for anatomically overlapping regions are shown in Table~\ref{tab:overlap_dsc} and Table~\ref{tab:overlap_nsd}, with qualitative examples provided in Fig.~\ref{fig:bone_seg_result1_AP} and Fig.~\ref{fig:bone_seg_result2_AP}. Compared with the overall bone-wise results in Table~\ref{tab:all_bone_dsc} and Table~\ref{tab:all_bone_nsd}, overlap-specific performance drops substantially, showing that projection overlap is one of the main bottlenecks in hand bone structure segmentation. While supervised models reach more than 97\% Bone Mean DSC on all bones, their overlap DSC values are much lower, and several difficult bone pairs fall below 60\%.

The difficulty varies across anatomical pairs. Relatively simple overlaps, such as Cap-Sca, Radius-Lu, Ham-MC5, and Tm-Td, obtain high DSC values for the best supervised models. For example, SwinUMamba achieves 89.55\% on Cap-Sca, 89.10\% on Ham-MC5, and 91.29\% on Tm-Td, while SwinUNETR achieves 88.24\% on Radius-Lu. These results indicate that current models can handle some overlap regions when the local anatomical structure remains clear. In contrast, severe overlap pairs show much lower performance. Cap-MC3, Ham-MC4, and Td-MC2 are particularly difficult, with the best DSC values reaching only 57.34\%, 67.21\%, and 46.11\%, respectively. These regions involve weak boundaries and heavy superimposition, making accurate separation of adjacent bones difficult.

Among supervised models, SwinUMamba provides the most balanced performance across overlapping regions. It achieves the best DSC on several representative pairs, including Cap-Sca, Radius-Sca, Ham-MC5, Lu-Sca, Tm-Td, Tm-MC1, Tm-MC2, Td-MC2, and MC2-MC3. UMambaEnc is also competitive, achieving the best DSC on Cap-Td and Ham-MC4, while SwinUNETR performs best on Radius-Lu and Sca-Tm. This suggests that models with stronger contextual modeling and boundary representation are more robust under anatomical ambiguity.

The NSD results in Table~\ref{tab:overlap_nsd} further highlight the importance of boundary accuracy. SwinUMamba achieves the best NSD on many overlap pairs, including Cap-Sca, Radius-Sca, Ham-MC5, Lu-Sca, Tm-Td, Tm-MC1, Tm-MC2, Td-MC2, and MC2-MC3. However, even the best NSD values remain far below the corresponding whole-bone NSD values in Table~\ref{tab:all_bone_nsd}, showing that overlap regions are highly sensitive to boundary shifts. Qualitative results in Fig.~\ref{fig:bone_seg_result1_AP} and Fig.~\ref{fig:bone_seg_result2_AP} show the same pattern: errors are concentrated around the wrist and metacarpal bases, where models may merge neighboring bones, leak across boundaries, or miss thin visible structures.

Foundation models perform poorly in overlap-specific evaluation. In Table~\ref{tab:overlap_dsc} and Table~\ref{tab:overlap_nsd}, SAM(Box), SAM(Point), and MedSAM(Box) obtain near-zero values on many overlap pairs, especially Cap-MC3, Ham-MC4, and MC2-MC3. Although MedSAM(Box) performs slightly better on some larger overlaps, such as Radius-Lu and Tm-Td, it remains far below supervised models. This confirms that prompt-based foundation models do not reliably resolve fine-grained anatomical overlap in hand radiographs.

Overall, the results indicate that supervised models already achieve strong performance for general hand bone structure segmentation, especially on large bones and clearly visible anatomical regions. The remaining challenges are therefore concentrated less on coarse whole-bone localization and more on fine-grained anatomical details, including small structures, subtle boundary variations, and instance separation in densely arranged regions. In particular, overlapping regions remain the main bottleneck: although current models can recover the overall hand skeleton, they still struggle to assign accurate boundaries when adjacent bones are heavily superimposed. Future work should therefore focus on overlap-aware and anatomy-aware segmentation, for example by introducing stronger boundary supervision, anatomical relationship modeling, topology-aware constraints, or prior-guided strategies that improve the separation of neighboring bones under projection ambiguity.

\begin{table*}[!t]
\centering
\caption{Spearman correlation results between bone segmentation-derived overlap size and the ground-truth total SvdH JSN score on the Test set. Significance levels: $*p<0.05$, $**p<0.01$, $***p<0.001$.}
\label{tab:stat_eva_boneseg_AP}
\begin{threeparttable}
\resizebox{0.6\linewidth}{!}{
\setlength{\tabcolsep}{10pt}
\begin{tabular}{lccc}
\toprule
\textbf{Model} 
& \textbf{Spearman $\rho$} 
& \textbf{$p$-value}
& \textbf{Significance} \\
\midrule
\multicolumn{4}{c}{\textbf{Supervised Models}} \\
\midrule
Unet & -0.1234 & $0.0440$ & $*$ \\
Unet++ & -0.1184 & $0.0533$ & -- \\
SegFormer & -0.1251 & $0.0411$ & $*$ \\
TransUnet & -0.1163 & $0.0576$ & -- \\
SwinUNETR & -0.1193 & $0.0515$ & -- \\
UMambaEnc & -0.1198 & $0.0506$ & -- \\
SwinUMamba & -0.1195 & $0.0511$ & -- \\
MambaVision & -0.1063 & $0.0830$ & -- \\
\midrule
\multicolumn{4}{c}{\textbf{Foundation Models}} \\
\midrule
SAM(Box) & -0.1270 & $0.0382$ & $*$ \\
SAM(Point) & 0.0704 & $0.2516$ & -- \\
MedSAM(Box) & 0.0246 & $0.6896$ & -- \\
\bottomrule
\end{tabular}
}
\end{threeparttable}
\end{table*}

\begin{figure}[!t]
    \centering
    \includegraphics[width=\textwidth]{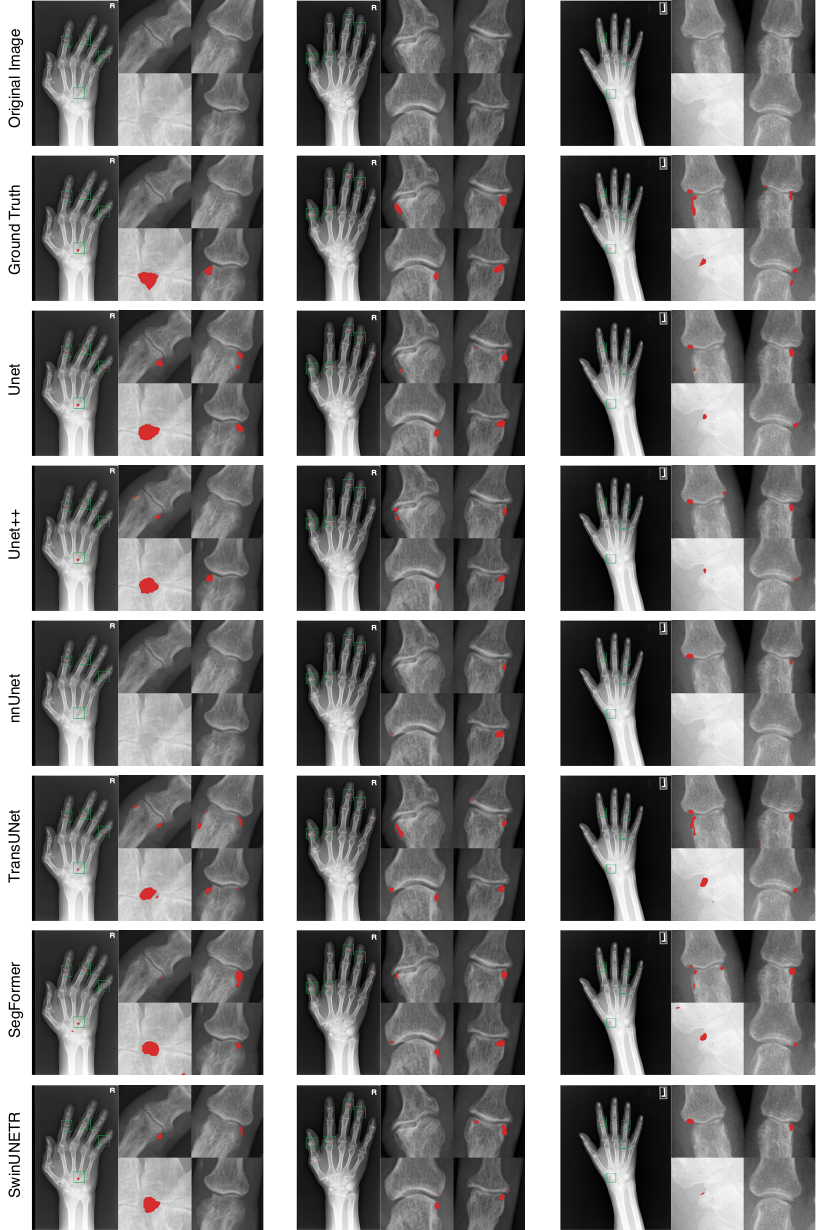}
    \caption{SvdH-BE-90 segmentation results (A).}
    \label{fig:be_seg_90_result1_AP}
\end{figure}

\subsubsection{Correlation Analysis Between Bone Segmentation-Derived Overlap Size and Ground-Truth Total SvdH JSN Score}
The clinical relevance of overlap size was evaluated using Spearman's rank correlation analysis with the total JSN score. As summarized in Table~\ref{tab:stat_eva_boneseg_AP}, the associations between predicted overlap size and total JSN score were generally weak across models. Among the supervised models, Unet ($\rho=-0.1234$, $p=0.0440$) and SegFormer ($\rho=-0.1251$, $p=0.0411$) showed statistically significant correlations, whereas the remaining supervised models did not reach statistical significance. For the foundation models, SAM(Box) also exhibited a weak but significant negative correlation ($\rho=-0.1270$, $p=0.0382$), while SAM(Point) and MedSAM(Box) showed no significant association.

These findings indicate that predicted overlap size alone has limited monotonic association with total JSN severity. Although several models reached nominal statistical significance, the small absolute correlation coefficients suggest that overlap size is not a strong standalone surrogate for JSN score. This may be because the total JSN score reflects localized joint-space narrowing patterns, whereas overlap size represents a global area-based measurement. Therefore, overlap size should be interpreted as a complementary structural descriptor rather than a direct proxy for JSN severity.

\subsection{Hand BE Segmentation}
\label{sec:be_seg_result_AP}
We evaluate BE segmentation using two complementary settings: SvdH-BE-90 and multi-class BE segmentation. The former focuses on high-confidence, clinically defined erosions, while the latter includes multiple BE categories (e.g., SvdH-BE-90, SvdH-BE-50, and Non-SvdH-BE), reflecting different confidence levels and definition criteria. Results from both settings show that BE segmentation remains highly challenging due to small lesion size, low contrast, and ambiguous boundaries. Together, these two protocols provide a more comprehensive evaluation, covering both reliable clinical targets and broader erosion patterns.
\subsubsection{SvdH-BE-90 Segmentation}
\label{sec:benchmark_be90_all_AP}

\begin{figure}[!t]
    \centering
    \includegraphics[width=\textwidth]{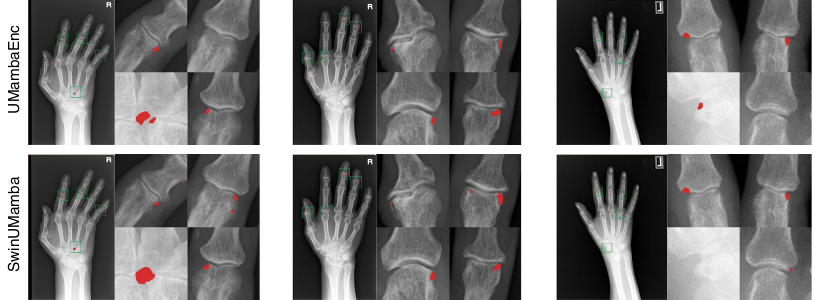}
    \caption{SvdH-BE-90 segmentation results (B).}
    \label{fig:be_seg_90_result2_AP}
\end{figure}

Qualitative results for SvdH-BE-90 segmentation are shown in Fig.~\ref{fig:be_seg_90_result1_AP} and Fig.~\ref{fig:be_seg_90_result2_AP}. The visual comparisons show that SvdH-defined erosion segmentation remains highly challenging even when models are trained specifically for this task. Across different cases, the predicted masks are generally sparse and unstable, and many small erosion regions are either missed or only partially detected. This is particularly evident in low-contrast regions, where the boundary between true cortical erosion and normal anatomical variation is visually ambiguous.

The results also reveal a strong tendency toward either under-detection or over-segmentation. Some models produce conservative predictions and miss subtle erosion regions, while others generate scattered false-positive masks around cortical edges, joint spaces, or overlapping bone structures. This indicates that SvdH-BE-90 segmentation is not only a small-target segmentation problem, but also a fine-grained discrimination problem: the model must distinguish clinically meaningful erosions from normal radiographic irregularities. In Fig.~\ref{fig:be_seg_90_result1_AP} and Fig.~\ref{fig:be_seg_90_result2_AP}, even relatively successful predictions often show imperfect boundary alignment, suggesting that lesion localization and precise contour delineation remain difficult to optimize simultaneously.

Future work should therefore focus on improving the balance between sensitivity and false-positive control for clinically defined BE regions. Promising directions include lesion-aware sampling, hard negative mining, boundary-sensitive supervision, and anatomy-guided segmentation. Incorporating bone structure priors or joint-level SvdH context may also help models distinguish true erosion regions from normal cortical variations and projection artifacts.

\begin{table*}[!t]
\centering
\caption{Independent-reader and model performance for SvdH-BE-90 segmentation on the same 50-image test subset. All systems are evaluated against the same consensus annotations using the benchmark evaluation protocol. The independent reader is reported as a clinical reference and is not included in the model ranking. Best and second-best model results are shown in \textbf{bold} and \underline{underline}, respectively.}
\label{tab:be_human_resegmentation_AP}
\begin{threeparttable}
\resizebox{\textwidth}{!}{
\begin{tabular}{lccccccc}
\toprule
 &
\textbf{DSC (\%)} &
\textbf{NSD (\%)} &
\textbf{Inst.} &
\textbf{Inst./GT} &
\textbf{Det. Rec. (\%)} &
\textbf{Det. Prec. (\%)} &
\textbf{Ctr. DSC (\%)} \\
\midrule
Independent reader & 41.16 & 51.51 & 153 & 1.0625 & 57.64 & 54.25 & 68.20 \\
\midrule
Unet & 15.48 & 18.69 & 533 & 3.7014 & 52.78 & 14.26 & 53.37 \\
Unet++ & 12.41 & 15.23 & 510 & 3.5417 & 39.58 & 11.18 & 52.68 \\
nnUnet & 12.22 & 15.86 & 94 & 0.6528 & 20.14 & \textbf{30.85} & 59.08 \\
TransUnet & \underline{16.82} & 19.32 & 587 & 4.0764 & \textbf{55.56} & 13.63 & 53.28 \\
SegFormer & 9.19 & 10.69 & 812 & 5.6389 & 53.47 & 9.48 & 44.03 \\
SwinUNETR & 14.46 & 15.71 & 736 & 5.1111 & 51.39 & 10.05 & 57.14 \\
UMambaEnc & 16.55 & \underline{20.41} & 456 & 3.1667 & 46.53 & \underline{14.69} & \underline{61.54} \\
SwinUMamba & \textbf{19.99} & \textbf{21.85} & 552 & 3.8333 & \underline{54.86} & 14.31 & \textbf{62.09} \\
\bottomrule
\end{tabular}
}

\begin{tablenotes}[flushleft]
\footnotesize
\item \textbf{Inst.}: Number of connected lesion instances.
\item \textbf{Inst./GT}: Ratio of detected instances to the 144 consensus instances.
\item \textbf{Det. Rec.}: Lesion co-detection recall.
\item \textbf{Det. Prec.}: Lesion co-detection precision.
\item \textbf{Ctr. DSC}: Contour DSC over co-detected lesions.
\item \textbf{Instance matching}: Connected components of at least 5 pixels are greedily matched when mask IoU $>0$.
\end{tablenotes}
\end{threeparttable}
\end{table*}

\subsubsection{Independent-Reader Analysis of SvdH-BE-90 Segmentation}
\label{sec:be_human_resegmentation_AP}

To further characterize reader-to-consensus variability under the SvdH-BE-90 segmentation setting, we conducted an additional re-segmentation analysis on a 50-image subset of the test set. An independent radiologist re-segmented the SvdH-BE-90 lesions without access to the consensus masks. The independent reader and all BE segmentation models were evaluated against the same consensus reference masks on the same subset using the benchmark evaluation protocol. In addition to pixel-level DSC and NSD, we decompose segmentation performance into lesion co-detection and contour delineation. BE instances are defined as connected components of at least 5 pixels and are greedily matched when their mask IoU is greater than zero.

As shown in Table~\ref{tab:be_human_resegmentation_AP}, the independent reader achieves a DSC of 41.16\% and an NSD of 51.51\% against the consensus annotations, whereas the best model, SwinUMamba, reaches 19.99\% DSC and 21.85\% NSD. The reader--consensus DSC indicates substantial variability in the precise delineation of SvdH-BE-90 lesions. At the same time, all evaluated models remain below the independent reader on the same images and reference masks, indicating that the low model-to-consensus segmentation performance cannot be explained solely by reader-to-consensus variability.

The instance-level decomposition further reveals different lesion-detection behaviors across models. The consensus annotations contain 144 connected lesion instances, compared with 153 identified by the independent reader. Most evaluated models produce substantially more instances than the consensus, ranging from 456 to 812, with co-detection precision between 9.48\% and 14.69\%. In contrast, nnUnet produces only 94 instances and exhibits the opposite pattern, with higher co-detection precision (30.85\%) but substantially lower co-detection recall (20.14\%). Thus, the evaluated models exhibit different lesion-detection failure modes rather than a uniform tendency toward over-detection.

For lesions co-detected with the consensus, contour DSC reaches 68.20\% for the independent reader and up to 62.09\% for SwinUMamba. Notably, nnUnet also achieves a relatively high contour DSC of 59.08\% despite its low co-detection recall. These results indicate that whole-mask segmentation performance reflects multiple factors, including missed lesions, false-positive detections, and contour disagreement, rather than a single detection failure mode. Overall, this analysis quantifies reader-to-consensus variability in SvdH-BE-90 delineation and provides a clinical reference for interpreting model segmentation performance on the same evaluation subset.

\begin{table}[!t]
\centering
\caption{Joint- and lesion-level detection performance for SvdH-BE-90 segmentation. The best results in each column are highlighted in \textbf{bold}, and the second-best values are \underline{underlined}.}
\label{tab:be_detection_metrics_AP}
\resizebox{\linewidth}{!}{
\setlength{\tabcolsep}{10pt}
\begin{tabular}{lcccccc}
\toprule
& \multicolumn{4}{c}{\textbf{Joint-level Detection}} & \multicolumn{2}{c}{\textbf{Lesion-level Detection}} \\
\cmidrule(lr){2-5} \cmidrule(lr){6-7}
\textbf{Method} &
\textbf{SEN $\uparrow$ (\%)} &
\textbf{SPE $\uparrow$ (\%)} &
\textbf{F1 $\uparrow$ (\%)} &
\textbf{Cohen's $\kappa$ $\uparrow$} &
\textbf{mAP$_{[0.5:0.95]}$ $\uparrow$} &
\textbf{F1 $\uparrow$ (\%)} \\
\midrule
Unet          & 69.22 & 64.95 & 36.62 & 0.1963 & 0.0119 & 8.34 \\
Unet++        & 69.22 & 63.97 & 36.06 & 0.1880 & \underline{0.0137} & 9.62 \\
nnUnet        & 26.38 & \textbf{93.00} & 31.40 & \textbf{0.2236} & 0.0118 & \textbf{12.77} \\
TransUnet     & 73.78 & 61.97 & \underline{36.86} & 0.1950 & 0.0128 & 8.93 \\
SegFormer     & \textbf{79.15} & 46.47 & 31.79 & 0.1144 & 0.0035 & 4.92 \\
SwinUNETR     & \underline{76.87} & 48.66 & 31.85 & 0.1173 & 0.0135 & 7.49 \\
UMambaEnc     & 62.05 & \underline{71.05} & 37.10 & \underline{0.2122} & 0.0112 & 10.22 \\
SwinUMamba    & 73.94 & 62.36 & \textbf{37.14} & 0.1989 & \textbf{0.0153} & \underline{11.07} \\
\bottomrule
\end{tabular}
}
\end{table}

\subsubsection{Detection-oriented Evaluation of SvdH-BE-90 Segmentation}
\label{sec:be_detection_metrics_AP}

Pixel-level overlap metrics do not distinguish between failure to identify an erosion and inaccurate delineation of a localized lesion. We therefore complement DSC and NSD with detection-oriented evaluation at both the SvdH joint level and the individual lesion level.

For joint-level evaluation, the SvdH-BE-90 segmentation results are converted into binary BE detection outcomes for each SvdH-defined joint region using a spatial tolerance of $\tau=10$ pixels. We report sensitivity, specificity, F1 score, and Cohen's $\kappa$. For lesion-level evaluation, connected components of at least 5 pixels are treated as individual BE instances and greedily matched one-to-one to reference lesions according to mask IoU. We report mAP averaged over IoU thresholds from 0.5 to 0.95 and detection F1 at an IoU threshold of 0.5. Table~\ref{tab:be_detection_metrics_AP} summarizes the detection-oriented results on the test set.

At the joint level, the evaluated models exhibit markedly different sensitivity--specificity trade-offs. Most models achieve sensitivities above 60\%, but their specificities range from 46.47\% to 71.05\%. SegFormer achieves the highest sensitivity (79.15\%) while exhibiting the lowest specificity (46.47\%), whereas nnUnet shows the opposite behavior, with the highest specificity (93.00\%) but substantially lower sensitivity (26.38\%). These results indicate that the detection errors underlying low pixel-level overlap vary across architectures, ranging from frequent false-positive predictions to substantial under-detection.

Lesion-level detection remains challenging under stricter instance matching. The best mAP$_{[0.5:0.95]}$ is 0.0153, achieved by SwinUMamba, while nnUnet achieves the highest lesion detection F1 of 12.77\%. Performance at this level is affected by missed and false-positive lesions, confidence ranking, and the sensitivity of IoU-based matching to small spatial deviations in these tiny erosion instances. Together with the co-detection and contour analysis in Sec.~\ref{sec:be_human_resegmentation_AP}, these results characterize complementary failure modes at the joint, lesion, and contour levels.

\begin{table*}[!t]
\centering
\caption{Multi-class BE segmentation results obtained on the Test set.}
\begin{threeparttable}
\resizebox{\textwidth}{!}{
\setlength{\tabcolsep}{5pt}
\begin{tabular}{lcccccccc}
\toprule
\textbf{Model} 
& \textbf{DSC $\uparrow$ (\%)} 
& \textbf{NSD $\uparrow$ (\%)} 
& \textbf{REC $\uparrow$ (\%)} 
& \textbf{PREC $\uparrow$ (\%)} 
& \textbf{VOE $\downarrow$ (\%)} 
& \textbf{MSD $\downarrow$ (pix)} 
& \textbf{\#P (M)} 
& \textbf{Time (ms)} \\
\midrule

Unet
& 12.78$\pm$11.41 
& 8.22$\pm$7.08 
& 9.58$\pm$10.01 
& \underline{7.27$\pm$7.37}
& 92.45$\pm$7.62 
& \textbf{206.61$\pm$106.07 }
& 7.94 
& 365.42 \\

Unet++
& \underline{14.48$\pm$10.80}
& 9.65$\pm$7.16 
& 11.72$\pm$9.90 
& \textbf{7.96$\pm$7.81 }
& \underline{91.48$\pm$6.96}
& \underline{220.90$\pm$99.59}
& 2.41 
& 772.82 \\

TransUnet
& \textbf{14.53$\pm$11.71 }
& \underline{9.59$\pm$8.06}
& \underline{14.16$\pm$11.64}
& 7.10$\pm$7.72 
& \textbf{91.28$\pm$8.30 }
& 236.23$\pm$109.68 
& 105.32 
& 849.10 \\

SegFormer
& 9.75$\pm$9.41 
& 6.18$\pm$5.70
& 10.37$\pm$10.24 
& 4.60$\pm$5.80 
& 94.35$\pm$6.35
& 231.40$\pm$99.01 
& 21.88 
& 272.00 \\

SwinUNETR
& 13.13$\pm$10.16 
& 7.78$\pm$5.96 
& 13.03$\pm$11.05 
& 6.17$\pm$5.96 
& 92.36$\pm$6.53 
& 230.94$\pm$100.65 
& 25.14 
& 826.32 \\

UMambaEnc
& 14.14$\pm$10.41 
& 8.46$\pm$6.26 
& \textbf{15.38$\pm$11.99 }
& 6.20$\pm$5.82 
& 91.76$\pm$7.06 
& 239.12$\pm$96.71 
& 4.58 
& 780.26 \\

SwinUMamba
& 12.25$\pm$9.68 
& \textbf{10.72$\pm$8.27 }
& 11.11$\pm$10.16 
& 5.76$\pm$5.70 
& 92.95$\pm$6.06 
& 226.17$\pm$103.98 
& 59.89 
& 1361.45 \\

\bottomrule
\end{tabular}
}
\raggedright
\end{threeparttable}
\label{tab:beseg_3ch_AP}
\end{table*}

\begin{table*}[!t]
\centering
\caption{Segmentation results of three categories of BE on the Test set. (a) DSC (\%), (b) NSD (\%).}
\begin{subtable}[!t]{0.49\textwidth}
\centering
\resizebox{\textwidth}{!}{
\setlength{\tabcolsep}{5pt}
\begin{tabular}{lccc}
\toprule
\textbf{Model} & \textbf{SvdH-BE-90} & \textbf{SvdH-BE-50} & \textbf{Non-SvdH-BE} \\
\midrule
Unet & 15.55$\pm$12.83 & 7.70$\pm$11.21 & 22.98$\pm$27.89 \\
Unet++ & \textbf{18.50$\pm$12.17} & 8.06$\pm$13.43 & 21.13$\pm$23.72 \\
TransUnet & 16.81$\pm$11.42 & \textbf{9.74$\pm$14.67} & \textbf{32.15$\pm$32.49} \\
SegFormer & 11.67$\pm$9.47 & 5.58$\pm$9.86 & 24.32$\pm$28.44 \\
SwinUNETR & 16.81$\pm$11.83 & 6.89$\pm$11.11 & \underline{24.84$\pm$22.11} \\
UMambaEnc & 16.61$\pm$11.44 & \underline{9.67$\pm$12.21} & 22.17$\pm$24.87 \\
SwinUMamba & \underline{17.03$\pm$11.79} & 7.23$\pm$12.37 & - \\
\bottomrule
\end{tabular}
}
\end{subtable}
\hfill
\begin{subtable}[!t]{0.49\textwidth}
\centering
\resizebox{\textwidth}{!}{
\setlength{\tabcolsep}{5pt}
\begin{tabular}{lccc}
\toprule
\textbf{Model} & \textbf{SvdH-BE-90} & \textbf{SvdH-BE-50} & \textbf{Non-SvdH-BE} \\
\midrule
Unet & 16.15$\pm$14.46 & 5.43$\pm$9.26 & 2.79$\pm$12.76 \\
Unet++ & \textbf{18.07$\pm$13.95} & 5.65$\pm$11.01 & \underline{3.26$\pm$12.03} \\
TransUnet & 16.62$\pm$12.88 & \underline{5.82$\pm$10.43} & \textbf{4.89$\pm$17.90} \\
SegFormer & 12.04$\pm$10.65 & 3.68$\pm$7.05 & 2.65$\pm$11.31 \\
SwinUNETR & 16.58$\pm$13.39 & 4.30$\pm$7.80 & 2.46$\pm$9.19 \\
UMambaEnc & 16.56$\pm$12.92 & \textbf{6.09$\pm$9.34} & 2.53$\pm$10.93 \\
SwinUMamba & \underline{17.09$\pm$13.35} & 4.98$\pm$8.80 & - \\
\bottomrule
\end{tabular}
}
\end{subtable}

\label{tab:beseg_3ch_results_dsc_nsd_AP}
\end{table*}

\begin{figure}[!t]
    \centering
    \includegraphics[width=\textwidth]{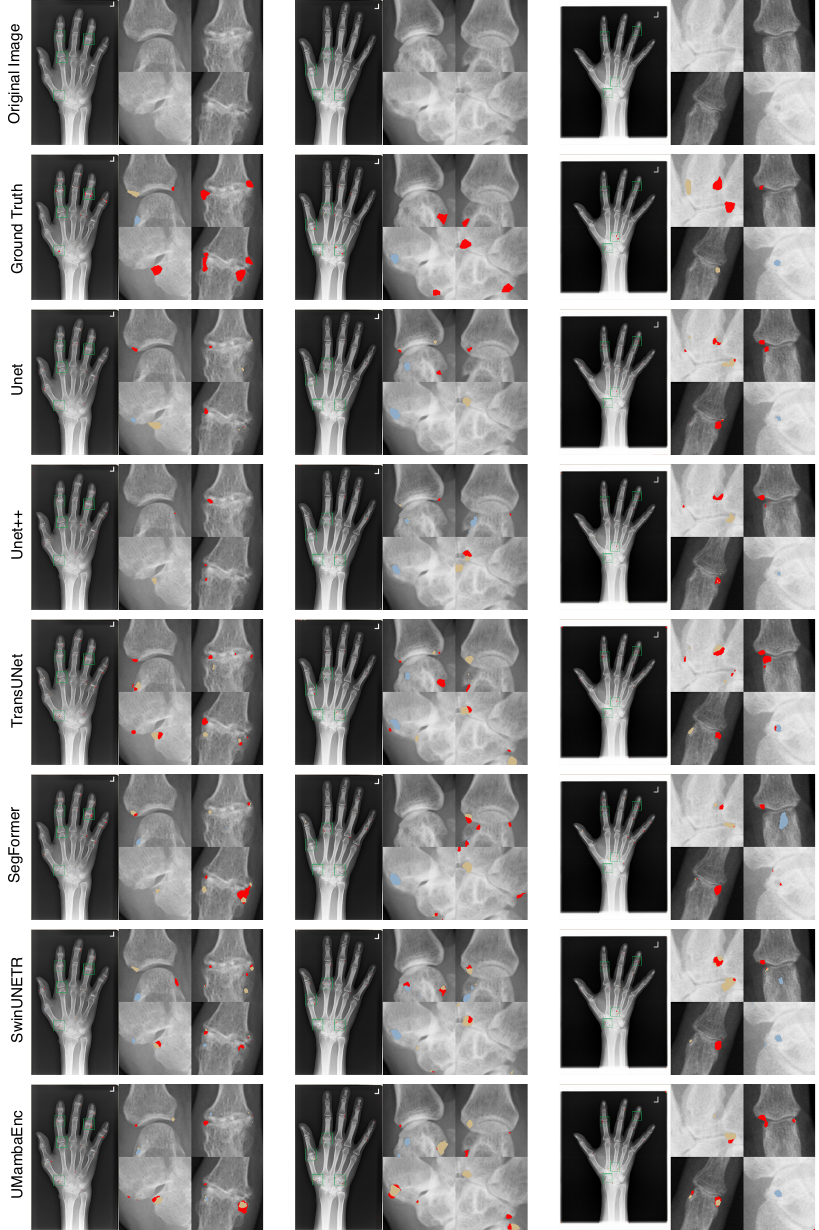}
    \caption{Multi-class BE segmentation results (A).}
    \label{fig:be_seg_3ch_result1_AP}
\end{figure}

\begin{figure}[!t]
    \centering
    \includegraphics[width=\textwidth]{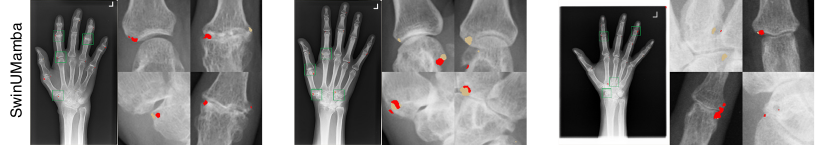}
    \caption{Multi-class BE segmentation results (B).}
    \label{fig:be_seg_3ch_result2_AP}
\end{figure}

\subsubsection{Multi-class BE segmentation}
\label{sec:benchmark_be_all_AP}

Multi-class BE segmentation results are summarized in Table~\ref{tab:beseg_3ch_AP} and Table~\ref{tab:beseg_3ch_results_dsc_nsd_AP}, with qualitative examples shown in Fig.~\ref{fig:be_seg_3ch_result1_AP} and Fig.~\ref{fig:be_seg_3ch_result2_AP}. Compared with SvdH-BE-90 segmentation alone, the multi-class setting is more difficult because the model must simultaneously segment SvdH-BE-90, SvdH-BE-50, and Non-SvdH-BE regions. As shown in Table~\ref{tab:beseg_3ch_AP}, all models obtain low overall performance, with the best DSC reaching only 14.53\% for TransUnet and the best NSD reaching 10.72\% for SwinUMamba. These results indicate that separating different erosion categories remains highly challenging.

Different models show different strengths across metrics. TransUnet achieves the best overall DSC and VOE, suggesting relatively better region overlap, while SwinUMamba obtains the highest NSD, indicating better boundary agreement. UMambaEnc reaches the highest recall of 15.38\%, showing stronger sensitivity to possible erosion regions, whereas Unet++ achieves the highest precision of 7.96\%. However, the absolute precision and recall values remain low for all models, showing that multi-class BE segmentation is still dominated by missed detections and false positives.

The class-wise results in Table~\ref{tab:beseg_3ch_results_dsc_nsd_AP} further show that different BE categories have different levels of difficulty. SvdH-BE-90 is the most stable category, with Unet++ achieving the best DSC of 18.50\% and NSD of 18.07\%. In contrast, SvdH-BE-50 is more difficult, with the best DSC only reaching 9.74\% and the best NSD only 6.09\%. Non-SvdH-BE obtains a higher DSC for some models, especially TransUnet with 32.15\%, but its NSD remains low, indicating that models may capture approximate lesion areas without accurately delineating their boundaries. The qualitative results in Fig.~\ref{fig:be_seg_3ch_result1_AP} and Fig.~\ref{fig:be_seg_3ch_result2_AP} are consistent with this pattern, showing frequent category confusion, missed small lesions, and scattered predictions around anatomically complex regions.

Future work on multi-class BE segmentation should address both lesion detection and category discrimination. Beyond improving binary erosion localization, models need stronger clinical and anatomical priors to distinguish SvdH-defined erosions from non-SvdH erosive changes. Multi-task learning with bone segmentation, joint ROI localization, or SvdH scoring may provide useful contextual constraints. In addition, class-balanced optimization, uncertainty-aware supervision, and category-specific hard example mining may help reduce confusion between visually similar BE categories.

\subsubsection{Correlation Analysis Between Predicted BE Size and Ground-Truth Total SvdH BE Score}
The association between predicted BE size and the total BE score was further examined using Spearman's rank correlation. As shown in Table~\ref{tab:stat_eva_be_seg_AP}, all BE segmentation models exhibited positive correlations with the total BE score, indicating that larger predicted erosion areas were generally associated with higher clinical erosion severity. The strongest correlations were observed for TransUnet ($\rho=0.3980$, $p<0.001$) and UMambaEnc ($\rho=0.3742$, $p<0.001$), followed by SwinUMamba ($\rho=0.3279$, $p<0.001$) and Unet ($\rho=0.3222$, $p<0.001$). In contrast, SegFormer and SwinUNETR showed weaker but still statistically significant associations ($\rho=0.1358$, $p=0.0265$ and $\rho=0.1359$, $p=0.0264$, respectively).

These results suggest that predicted BE size captures clinically relevant information related to bone erosion severity. However, the moderate magnitude of the correlations indicates that BE size should not be regarded as a complete substitute for total BE score. This is expected because the clinical BE score reflects joint-specific erosion patterns and ordinal severity grades, whereas the predicted BE size is a global area-based measurement. Therefore, BE size can serve as a useful quantitative imaging descriptor complementary to established clinical scoring.

\begin{table*}[!t]
\centering
\caption{Spearman correlation results between predicted BE size and the ground-truth total SvdH BE score on the Test set. Significance levels: $*p<0.05$, $**p<0.01$, $***p<0.001$.}
\label{tab:stat_eva_be_seg_AP}
\begin{threeparttable}
\resizebox{0.55\linewidth}{!}{
\setlength{\tabcolsep}{7pt}
\begin{tabular}{lccc}
\toprule
\textbf{Model} 
& \textbf{Spearman $\rho$} 
& \textbf{$p$-value}
& \textbf{Significance} \\
\midrule
Unet & 0.3222 & $<0.001$ & $***$ \\
Unet++ & 0.2533 & $<0.001$ & $***$ \\
SegFormer & 0.1358 & $0.0265$ & $*$ \\
TransUnet & 0.3980 & $<0.001$ & $***$ \\
SwinUNETR & 0.1359 & $0.0264$ & $*$ \\
UMambaEnc & 0.3742 & $<0.001$ & $***$ \\
SwinUMamba & 0.3279 & $<0.001$ & $***$ \\
\bottomrule
\end{tabular}
}
\end{threeparttable}
\end{table*}

\subsection{Scoring of SvdH BE}
\label{sec:be_score_result_AP}
\subsubsection{Joint-level Results and Confusion Matrices}
Tables~\ref{tab:be_joint_qwk_results} and~\ref{tab:be_joint_bacc_results} report the joint-level QWK and BACC results for the SvdH BE scoring task on the test set.
Overall, the results show clear heterogeneity across anatomical locations, indicating that BE scoring difficulty is strongly joint-dependent.

\begin{table*}[!t]
\centering
\caption{SvdH BE score classification QWK results for each joint on the Test set. The best results in each column are highlighted in \textbf{bold}, and the second-best values are \underline{underlined}. Mean denotes the overall QWK computed over all BE scoring samples.}
\label{tab:be_joint_qwk_results}

\begin{subtable}[!t]{\textwidth}
\centering
\resizebox{\textwidth}{!}{
\setlength{\tabcolsep}{12pt}
\begin{tabular}{lcccccccc}
\toprule
\textbf{Model} & \textbf{Radius} & \textbf{Ulna} & \textbf{IP} & \textbf{Lu} & \textbf{MCP-T} & \textbf{MCP-I} & \textbf{MCP-M} & \textbf{MCP-R} \\
\midrule
ResNet & 0.3032 & \underline{0.5499} & 0.0490 & 0.6623 & 0.3796 & \underline{0.6758} & \textbf{0.5981} & 0.3030 \\
DenseNet & \textbf{0.4157} & 0.5603 & 0.0978 & \textbf{0.7208} & 0.3160 & 0.6037 & 0.4368 & 0.3090 \\
EfficientNetV2 & \underline{0.3733} & 0.3754 & 0.0992 & 0.5226 & 0.3539 & 0.4233 & \underline{0.5582} & 0.2219 \\
MobileViT & 0.2028 & \textbf{0.6522} & 0.0888 & 0.6346 & \underline{0.4151} & 0.6387 & 0.4208 & 0.2594 \\
LeViT & 0.1547 & 0.2423 & 0.0330 & 0.4599 & 0.3505 & 0.4502 & 0.2458 & 0.1029 \\
EfficientFormer & 0.3162 & 0.4493 & 0.1205 & 0.5749 & 0.3877 & 0.5759 & 0.4158 & \textbf{0.3358} \\
ConvNeXtV2 & 0.2126 & 0.3191 & \textbf{0.2653} & 0.4435 & 0.3752 & 0.6216 & 0.3226 & 0.1725 \\
MedMamba & 0.3364 & 0.5497 & \underline{0.1908} & \underline{0.6781} & \textbf{0.5138} & \textbf{0.7087} & 0.5275 & \underline{0.3214} \\
MambaVision & 0.3561 & \underline{0.5793} & 0.0740 & 0.5189 & 0.3660 & 0.6396 & 0.5094 & 0.3102 \\
\bottomrule
\end{tabular}}
\end{subtable}

\vspace{0.6cm}

\begin{subtable}[!t]{\textwidth}
\centering
\resizebox{\textwidth}{!}{
\setlength{\tabcolsep}{10pt}
\begin{tabular}{lccccccccc}
\toprule
\textbf{Model} & \textbf{MCP-S} & \textbf{CMC-T} & \textbf{PIP-I} & \textbf{PIP-M} & \textbf{PIP-R} & \textbf{PIP-S} & \textbf{Sca} & \textbf{Tr} & \textbf{Mean} \\
\midrule
ResNet & \textbf{0.5228} & 0.1367 & \underline{0.3301} & \underline{0.5011} & \textbf{0.3573} & \underline{0.2093} & 0.3063 & 0.2069 & \underline{0.4408} \\
DenseNet & 0.4185 & 0.2770 & 0.2665 & 0.4488 & 0.2922 & \textbf{0.2125} & \underline{0.3294} & 0.1039 & 0.3905 \\
EfficientNetV2 & 0.3947 & 0.0967 & 0.2847 & 0.3472 & 0.1318 & 0.1209 & 0.1485 & \textbf{0.3560} & 0.3358 \\
MobileViT & 0.3514 & 0.1190 & 0.2594 & 0.4505 & 0.2184 & 0.1568 & 0.2846 & 0.0898 & 0.3920 \\
LeViT & 0.1354 & -0.0347 & 0.0987 & 0.2821 & 0.0803 & 0.1029 & 0.1861 & \underline{0.2538} & 0.2346 \\
EfficientFormer & 0.3942 & \textbf{0.2956} & 0.2464 & 0.4478 & 0.1914 & 0.1168 & 0.2703 & 0.2080 & 0.3504 \\
ConvNeXtV2 & 0.3549 & -0.0244 & 0.2650 & 0.4387 & 0.0693 & 0.1073 & 0.1316 & 0.2188 & 0.3058 \\
MedMamba & 0.3629 & \underline{0.2801} & \textbf{0.4392} & \textbf{0.5045} & \underline{0.3188} & 0.1936 & \textbf{0.3622} & 0.2331 & \textbf{0.4522} \\
MambaVision & \underline{0.4457} & 0.1472 & 0.2087 & 0.3563 & 0.2618 & 0.1825 & 0.3098 & 0.1759 & 0.3667 \\
\bottomrule
\end{tabular}}
\end{subtable}
\end{table*}

\begin{table*}[!t]
\centering
\caption{SvdH BE score classification BACC results (\%) for each joint on the Test set. The best results in each column are highlighted in \textbf{bold}, and the second-best values are \underline{underlined}. Mean denotes the overall BACC computed over all BE scoring samples.}
\label{tab:be_joint_bacc_results}

\begin{subtable}[!t]{\textwidth}
\centering
\resizebox{\textwidth}{!}{
\setlength{\tabcolsep}{12pt}
\begin{tabular}{lcccccccc}
\toprule
\textbf{Model} & \textbf{Radius} & \textbf{Ulna} & \textbf{IP} & \textbf{Lu} & \textbf{MCP-T} & \textbf{MCP-I} & \textbf{MCP-M} & \textbf{MCP-R} \\
\midrule
ResNet & 25.07 & \textbf{41.46} & 19.55 & \underline{31.51} & 41.54 & \underline{38.20} & \textbf{41.73} & 36.99 \\
DenseNet & \textbf{30.03} & 25.81 & 20.15 & \textbf{33.26} & 40.77 & 34.44 & \underline{35.08} & \underline{39.77} \\
EfficientNetV2 & \underline{29.53} & 24.33 & 19.75 & 26.12 & 42.29 & 33.45 & 35.04 & 25.42 \\
MobileViT & 23.29 & \underline{39.86} & 21.01 & 26.04 & \underline{43.71} & 32.65 & 27.08 & 26.33 \\
LeViT & 23.45 & 26.87 & 20.00 & 22.40 & 42.69 & 26.70 & 25.83 & 25.99 \\
EfficientFormer & 21.49 & 36.13 & 20.11 & 24.30 & 42.71 & 27.21 & 26.17 & 32.71 \\
ConvNeXtV2 & 24.74 & 31.65 & \textbf{21.77} & 19.69 & 42.59 & 33.58 & 26.20 & 26.16 \\
MedMamba & 25.07 & 29.89 & \underline{21.41} & 28.18 & \textbf{47.74} & \textbf{40.82} & 25.41 & 34.12 \\
MambaVision & 23.95 & 30.99 & 20.00 & 30.89 & 41.12 & 33.57 & 31.23 & \textbf{39.89} \\
\bottomrule
\end{tabular}}
\end{subtable}

\vspace{0.6cm}

\begin{subtable}[!t]{\textwidth}
\centering
\resizebox{\textwidth}{!}{
\setlength{\tabcolsep}{10pt}
\begin{tabular}{lccccccccc}
\toprule
\textbf{Model} & \textbf{MCP-S} & \textbf{CMC-T} & \textbf{PIP-I} & \textbf{PIP-M} & \textbf{PIP-R} & \textbf{PIP-S} & \textbf{Sca} & \textbf{Tr} & \textbf{Mean} \\
\midrule
ResNet & \textbf{34.05} & 21.52 & 27.29 & 37.38 & \underline{32.81} & 21.18 & 26.60 & \textbf{25.89} & \textbf{35.87} \\
DenseNet & \underline{30.48} & \textbf{41.31} & 31.25 & \underline{37.93} & 29.48 & \textbf{26.49} & 21.46 & 21.53 & 33.06 \\
EfficientNetV2 & 25.42 & 22.10 & \underline{33.43} & 30.42 & 20.92 & 19.96 & 22.17 & \underline{25.87} & 30.67 \\
MobileViT & 27.51 & 20.21 & 28.28 & 35.70 & \textbf{33.70} & 22.40 & 26.87 & 23.81 & 31.88 \\
LeViT & 25.63 & 19.85 & 22.24 & 22.45 & 26.08 & 20.06 & 22.95 & 24.77 & 25.66 \\
EfficientFormer & 27.03 & \underline{26.42} & 21.55 & 34.46 & 23.65 & 20.81 & 22.13 & 22.27 & 27.32 \\
ConvNeXtV2 & 25.30 & 20.46 & 31.52 & 33.76 & 20.86 & 18.67 & 20.83 & 22.79 & 28.15 \\
MedMamba & 25.54 & 20.29 & \textbf{42.30} & \textbf{40.92} & 31.89 & \underline{25.50} & \underline{29.74} & 23.65 & \underline{34.91} \\
MambaVision & 25.58 & 21.14 & 26.47 & 25.73 & 25.48 & 21.99 & \textbf{33.10} & 24.46 & 30.59 \\
\bottomrule
\end{tabular}}
\end{subtable}
\end{table*}

\begin{figure}[!t]
    \centering
    \includegraphics[width=0.97\linewidth]{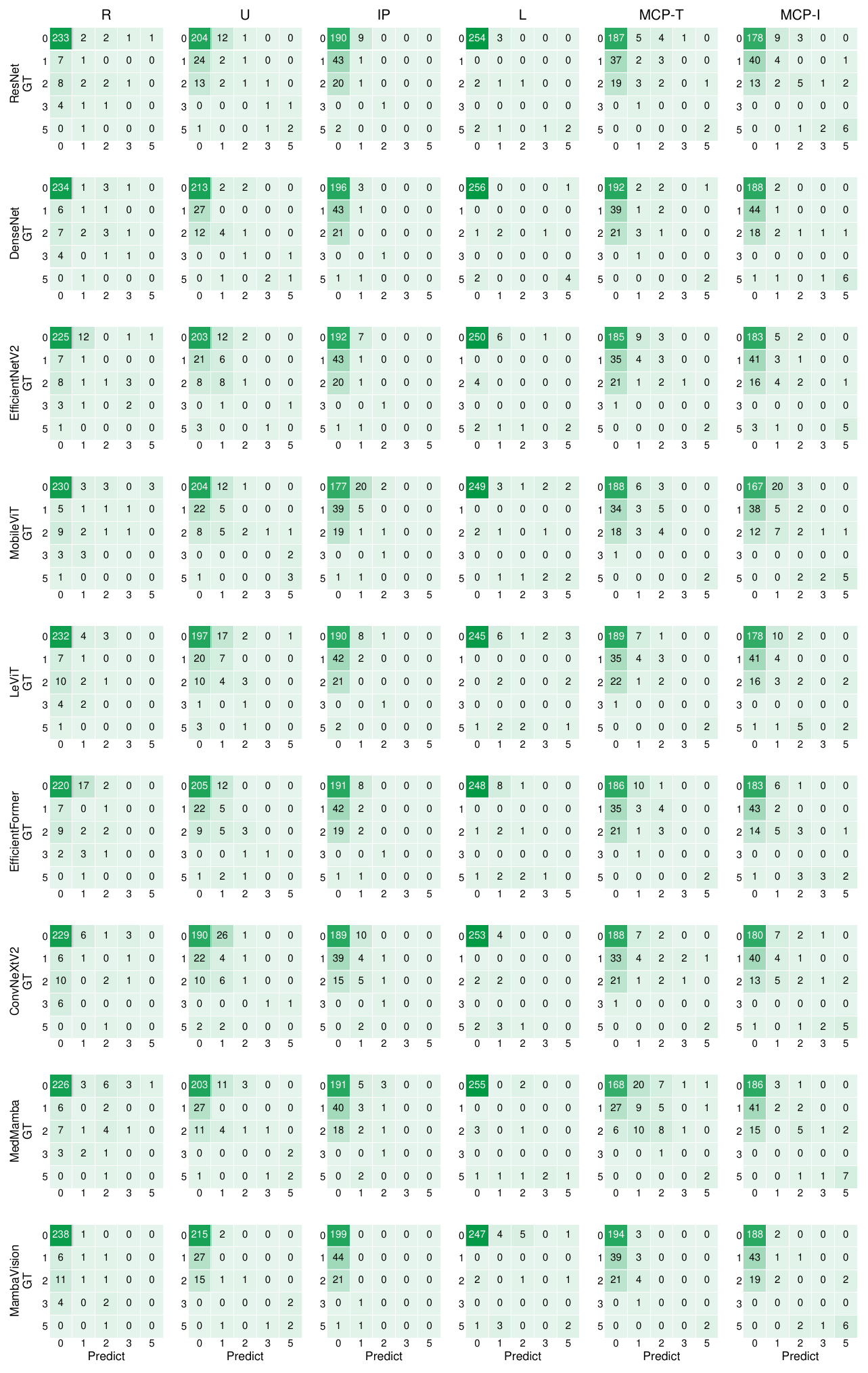}
    \caption{Joint-wise confusion matrices of SvdH BE scoring (A)}
    \label{fig:be_confusion_joint_1_AP}
\end{figure}

\begin{figure}
    \centering
    \includegraphics[width=0.97\linewidth]{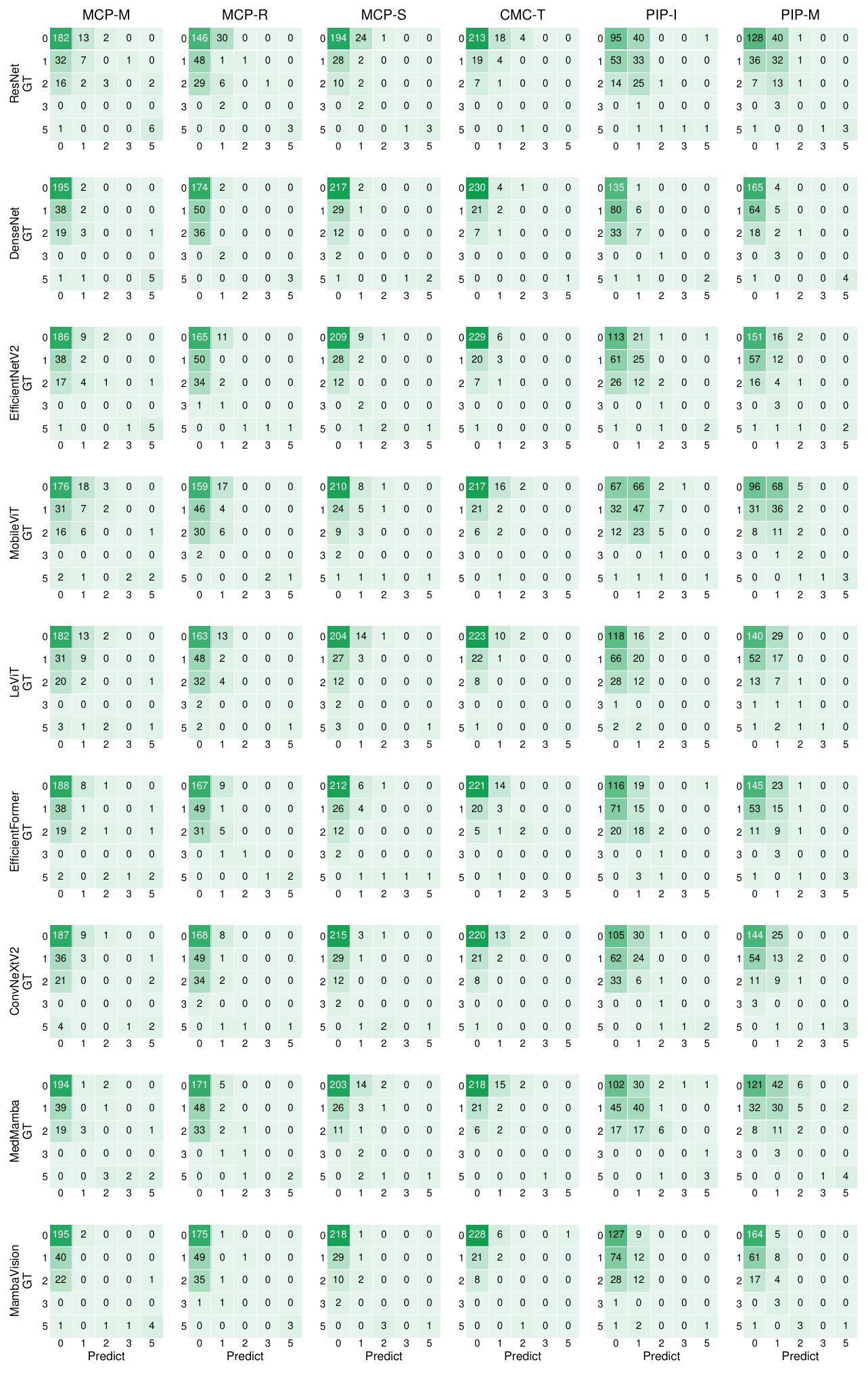}
    \caption{Joint-wise confusion matrices of SvdH BE scoring (B)}
    \label{fig:be_confusion_joint_2_AP}
\end{figure}

\begin{figure}
    \centering
    \includegraphics[width=0.67\linewidth]{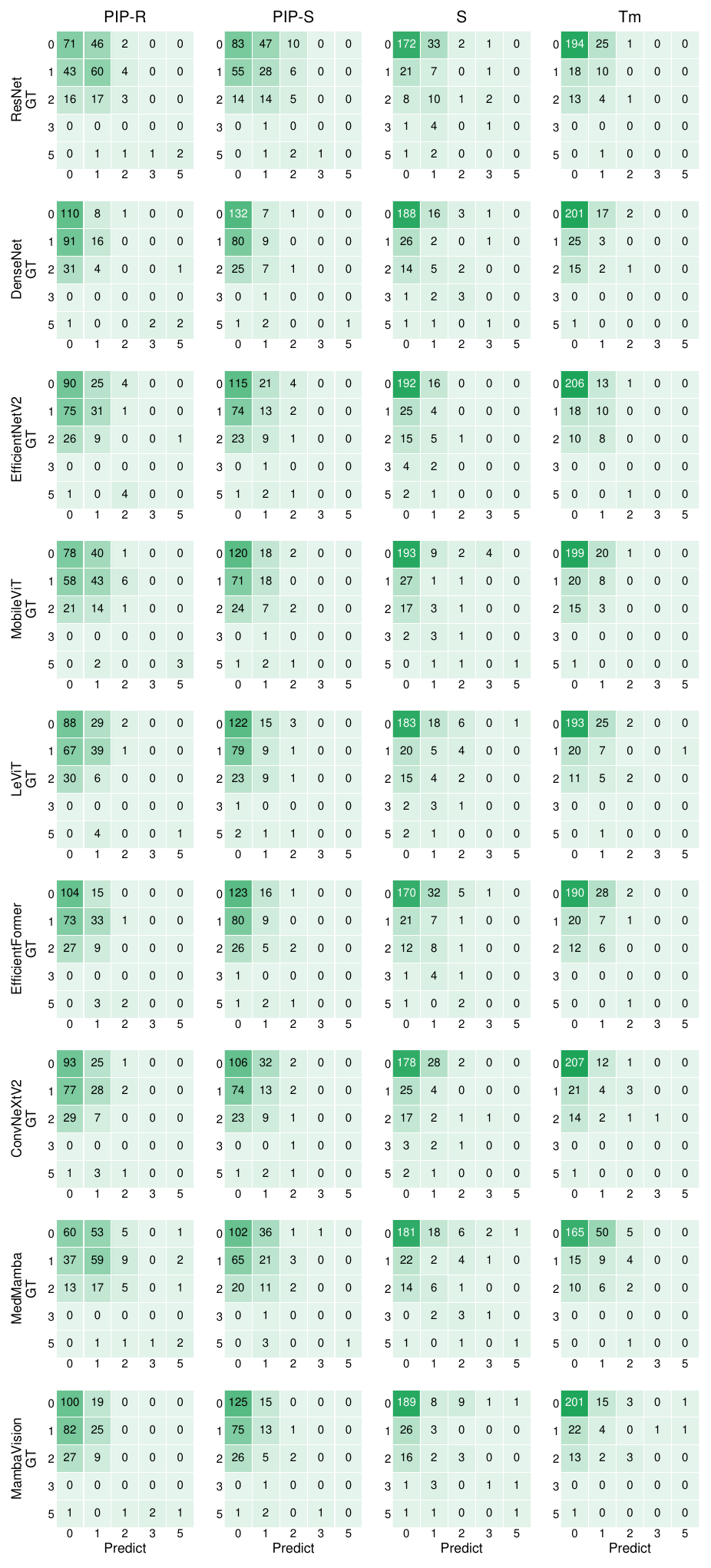}
    \caption{Joint-wise confusion matrices of SvdH BE scoring (C)}
    \label{fig:be_confusion_joint_3_AP}
\end{figure}

For joint-level QWK, the models show noticeable performance variation across anatomical sites. Higher agreement is generally observed for joints such as Lu, MCP-I, MCP-M, and Ulna, where several models achieve relatively strong ordinal consistency. In contrast, IP, CMC-T, PIP-S, and Tr are more challenging, with lower QWK values across many methods. Among the evaluated models, MedMamba achieves the best overall QWK reported in the Mean column and shows strong performance across several key joints, particularly MCP-T, MCP-I, PIP-I, and PIP-M. ResNet also demonstrates competitive joint-level ordinal agreement, especially on Ulna, Lu, MCP-I, MCP-M, and PIP-M, suggesting relatively robust performance across different joint types.

For joint-level BACC, the results show a similar pattern. ResNet obtains the highest overall BACC reported in the Mean column, indicating stronger balanced classification ability across joints, while MedMamba also achieves competitive performance and performs well on several MCP and PIP joints. In general, joints with clearer and more reliable visual patterns tend to achieve better BACC, whereas smaller or anatomically ambiguous joints remain difficult. This suggests that BE scoring performance is affected not only by model architecture, but also by joint-specific anatomical structure and lesion characteristics.

The joint-wise confusion matrices in Figs.~\ref{fig:be_confusion_joint_1_AP}, \ref{fig:be_confusion_joint_2_AP}, and~\ref{fig:be_confusion_joint_3_AP} further illustrate these differences. Joints with better quantitative results show clearer diagonal patterns, while more difficult joints have predictions concentrated in the lower BE scores or scattered across neighboring classes. This indicates that many errors are related to under-recognition of positive or severe erosion grades, especially in joints where lesions are small, subtle, or visually ambiguous.

These joint-level results further confirm that BE score classification remains highly heterogeneous across anatomical locations. The relatively low performance on several small or ambiguous joints suggests that improving joint-specific feature learning and ordinal-aware classification is important for more reliable BE assessment. Future work may benefit from anatomy-aware models that better capture local structural differences across hand and wrist regions. In addition, ordinal-aware loss functions, ranking-based learning strategies, and more effective class-imbalance handling may help reduce severe grading errors and improve recognition of underrepresented positive grades. Incorporating multi-joint contextual information, anatomical priors, and uncertainty estimation could further improve robustness for subtle or ambiguous erosive changes.

\subsubsection{Per-grade Scoring Analysis.}
\label{sec:be_pergrade_AP}
To further examine performance across individual SvdH-BE grades, we report per-grade recall and precision for all scoring models in Table~\ref{tab:be_pergrade_AP}. We additionally report 95\% confidence intervals for the higher grades (grades 2, 3, and 5) in Table~\ref{tab:be_pergrade_ci_AP}, estimated using patient-level cluster bootstrap ($B=2000$) by resampling patients while retaining all associated hands and joints. The corresponding numbers of predictions assigned to each grade are also reported as the denominators for the precision estimates. The results show clear grade-specific differences that are not fully reflected by the aggregate scoring metrics. In particular, grade-2 recall remains low across all nine models, ranging from 3.13\% to 11.11\%, despite having 351 test samples. In comparison, the less frequent grade 5 achieves recall between 14.75\% and 54.10\%, with a higher point estimate than grade 2 for every model. Performance therefore does not decrease monotonically with either grade severity or class frequency, but instead exhibits substantial class-specific confusion. The confidence intervals for the higher grades are generally broad, particularly for grade 3, which contains only 25 test samples, reflecting greater uncertainty in the performance estimates for these less represented grades.

\begin{table*}[!t]
\centering
\caption{
Per-grade recall (Rec.) and precision (Prec.) for SvdH BE scoring on the complete test set. All values are reported in percentage (\%).
}
\label{tab:be_pergrade_AP}
\resizebox{\textwidth}{!}{
\setlength{\tabcolsep}{9pt}
\begin{tabular}{lcccccccccc}
\toprule
& \multicolumn{2}{c}{\textbf{Grade 0}}
& \multicolumn{2}{c}{\textbf{Grade 1}}
& \multicolumn{2}{c}{\textbf{Grade 2}}
& \multicolumn{2}{c}{\textbf{Grade 3}}
& \multicolumn{2}{c}{\textbf{Grade 5}} \\
\cmidrule(lr){2-3}
\cmidrule(lr){4-5}
\cmidrule(lr){6-7}
\cmidrule(lr){8-9}
\cmidrule(lr){10-11}
\textbf{Model}
& \textbf{Rec.} & \textbf{Prec.}
& \textbf{Rec.} & \textbf{Prec.}
& \textbf{Rec.} & \textbf{Prec.}
& \textbf{Rec.} & \textbf{Prec.}
& \textbf{Rec.} & \textbf{Prec.} \\
\midrule
ResNet
& 87.36 & 78.96 & 27.06 & 28.61 & 7.41 & 31.33 & 8.00 & 9.09 & 49.18 & 76.92 \\
DenseNet
& 97.05 & 75.69 & 6.97 & 27.03 & 3.13 & 30.56 & 4.00 & 7.14 & 54.10 & 84.62 \\
EfficientNetV2
& 92.82 & 76.62 & 16.32 & 29.03 & 3.42 & 21.82 & 8.00 & 18.18 & 32.79 & 76.92 \\
MobileViT
& 87.56 & 78.58 & 26.50 & 29.23 & 5.98 & 22.58 & 0.00 & 0.00 & 39.34 & 70.59 \\
LeViT
& 91.31 & 76.25 & 18.13 & 29.15 & 3.99 & 20.29 & 0.00 & 0.00 & 14.75 & 45.00 \\
EfficientFormer
& 92.01 & 76.88 & 14.92 & 23.99 & 5.98 & 30.43 & 4.00 & 11.11 & 19.67 & 75.00 \\
ConvNeXtV2
& 91.47 & 76.15 & 15.34 & 25.64 & 3.70 & 22.81 & 4.00 & 5.56 & 26.23 & 69.57 \\
MedMamba
& 88.07 & 79.50 & 25.38 & 30.28 & 11.11 & 26.17 & 4.00 & 4.35 & 45.90 & 62.22 \\
MambaVision
& 96.31 & 76.03 & 10.18 & 31.60 & 3.13 & 23.91 & 4.00 & 11.11 & 39.34 & 66.67 \\
\bottomrule
\end{tabular}
}
\end{table*}

\begin{table*}[!t]
\centering
\caption{
Patient-level bootstrap 95\% confidence intervals for per-grade SvdH BE scoring ($B=2000$).
Pred.\ $n$ denotes the predicted-class count and therefore the denominator used to compute precision.
Values are reported as percentages (\%).
}
\label{tab:be_pergrade_ci_AP}

\resizebox{\textwidth}{!}{
\begin{tabular}{llccccc}
\toprule
\textbf{Model} & \textbf{Metric} &
\textbf{Grade 0} & \textbf{Grade 1} & \textbf{Grade 2} &
\textbf{Grade 3} & \textbf{Grade 5} \\
\midrule

\multirow{3}{*}{ResNet}
& Rec.\ CI
& [85.61, 89.38]
& [22.41, 29.95]
& [4.30, 11.24]
& [0.00, 25.00]
& [21.43, 71.61] \\
& Prec.\ CI
& [74.49, 83.49]
& [22.13, 33.71]
& [20.27, 39.81]
& [0.00, 22.74]
& [46.67, 88.89] \\
& Pred.\ $n$
& 3450 & 678 & 83 & 22 & 39 \\
\midrule

\multirow{3}{*}{DenseNet}
& Rec.\ CI
& [95.90, 97.98]
& [4.77, 10.65]
& [1.43, 6.18]
& [0.00, 16.67]
& [27.27, 69.51] \\
& Prec.\ CI
& [70.80, 81.11]
& [20.29, 33.98]
& [15.79, 48.39]
& [0.00, 22.22]
& [62.50, 94.29] \\
& Pred.\ $n$
& 3998 & 185 & 36 & 14 & 39 \\
\midrule

\multirow{3}{*}{EfficientNetV2}
& Rec.\ CI
& [91.31, 94.08]
& [12.51, 22.83]
& [1.60, 6.17]
& [0.00, 26.67]
& [13.88, 46.58] \\
& Prec.\ CI
& [71.64, 82.11]
& [23.86, 34.12]
& [11.54, 32.56]
& [0.00, 50.00]
& [43.73, 92.86] \\
& Pred.\ $n$
& 3777 & 403 & 55 & 11 & 26 \\
\midrule

\multirow{3}{*}{MobileViT}
& Rec.\ CI
& [85.75, 89.35]
& [23.35, 30.55]
& [3.66, 9.39]
& [0.00, 0.00]
& [26.19, 54.10] \\
& Prec.\ CI
& [74.14, 83.55]
& [23.11, 34.89]
& [16.25, 30.67]
& [0.00, 0.00]
& [48.14, 90.48] \\
& Pred.\ $n$
& 3474 & 650 & 93 & 21 & 34 \\
\midrule

\multirow{3}{*}{LeViT}
& Rec.\ CI
& [89.78, 92.86]
& [14.87, 23.06]
& [1.73, 7.27]
& [0.00, 0.00]
& [3.03, 31.75] \\
& Prec.\ CI
& [71.41, 81.55]
& [23.03, 34.09]
& [11.36, 29.73]
& [0.00, 0.00]
& [11.11, 70.00] \\
& Pred.\ $n$
& 3734 & 446 & 69 & 3 & 20 \\
\midrule

\multirow{3}{*}{EfficientFormer}
& Rec.\ CI
& [90.27, 93.88]
& [11.52, 19.17]
& [2.49, 11.77]
& [0.00, 18.18]
& [4.44, 37.50] \\
& Prec.\ CI
& [72.00, 82.01]
& [17.72, 28.63]
& [18.91, 39.44]
& [0.00, 83.25]
& [42.86, 100.00] \\
& Pred.\ $n$
& 3732 & 446 & 69 & 9 & 16 \\
\midrule

\multirow{3}{*}{ConvNeXtV2}
& Rec.\ CI
& [90.09, 92.67]
& [12.13, 20.47]
& [1.87, 6.83]
& [0.00, 18.18]
& [8.33, 43.33] \\
& Prec.\ CI
& [71.24, 81.74]
& [19.60, 30.66]
& [13.16, 33.82]
& [0.00, 25.00]
& [41.67, 91.68] \\
& Pred.\ $n$
& 3745 & 429 & 57 & 18 & 23 \\
\midrule

\multirow{3}{*}{MedMamba}
& Rec.\ CI
& [86.00, 90.46]
& [20.73, 29.39]
& [6.77, 16.09]
& [0.00, 20.00]
& [28.20, 60.87] \\
& Prec.\ CI
& [74.67, 84.47]
& [23.23, 35.69]
& [17.39, 34.00]
& [0.00, 15.79]
& [41.46, 78.26] \\
& Pred.\ $n$
& 3454 & 601 & 149 & 23 & 45 \\
\midrule

\multirow{3}{*}{MambaVision}
& Rec.\ CI
& [95.02, 97.80]
& [7.61, 12.80]
& [1.11, 5.06]
& [0.00, 16.67]
& [24.44, 58.46] \\
& Prec.\ CI
& [71.30, 81.10]
& [24.00, 38.29]
& [11.11, 38.46]
& [0.00, 50.00]
& [50.00, 83.35] \\
& Pred.\ $n$
& 3950 & 231 & 46 & 9 & 36 \\

\bottomrule
\end{tabular}
}
\end{table*}

\subsubsection{Correlation Analysis Between Predicted and Ground-Truth SvdH BE Scores}
To assess the consistency between model predictions and clinical evaluation, Spearman's rank correlation analysis was conducted between the predicted BE scores and the reference total BE scores. As shown in Table~\ref{tab:stat_eva_be_AP}, all models exhibited statistically significant positive correlations with the clinical BE scores ($p<0.001$), indicating that the predicted scores were generally aligned with expert annotations. The strongest association was achieved by MambaVision ($\rho=0.5063$), followed by ResNet ($\rho=0.4708$), MedMamba ($\rho=0.4692$), and MobileViT ($\rho=0.4679$). EfficientFormer also showed a moderate correlation ($\rho=0.4120$), whereas the remaining models demonstrated weaker but still significant associations.

These results suggest that the BE score prediction models capture clinically meaningful ranking information related to bone erosion severity. However, the correlations were moderate rather than high, indicating that model predictions are not fully interchangeable with expert-derived BE scores. This may reflect the ordinal and joint-specific nature of clinical BE scoring, as well as the difficulty of aggregating local erosion patterns into a total score. Therefore, predicted BE scores should be interpreted as clinically informative estimates rather than direct replacements for expert assessment.

\begin{table*}[!t]
\centering
\caption{Spearman correlation results between predicted SvdH BE scores and ground-truth SvdH BE scores on the Test set. Significance levels: $*p<0.05$, $**p<0.01$, $***p<0.001$.}
\label{tab:stat_eva_be_AP}
\begin{threeparttable}
\resizebox{0.6\linewidth}{!}{
\setlength{\tabcolsep}{8pt}
\begin{tabular}{lccc}
\toprule
\textbf{Model} 
& \textbf{Spearman $\rho$} 
& \textbf{$p$-value}
& \textbf{Significance} \\
\midrule
ResNet & 0.4708 & $<0.001$ & $***$ \\
DenseNet & 0.3447 & $<0.001$ & $***$ \\
EfficientNetV2 & 0.2923 & $<0.001$ & $***$ \\
MobileViT & 0.4679 & $<0.001$ & $***$ \\
LeViT & 0.3236 & $<0.001$ & $***$ \\
EfficientFormer & 0.4120 & $<0.001$ & $***$ \\
ConvNeXtV2 & 0.3206 & $<0.001$ & $***$ \\
MedMamba & 0.4692 & $<0.001$ & $***$ \\
MambaVision & 0.5063 & $<0.001$ & $***$ \\
\bottomrule
\end{tabular}
}
\end{threeparttable}
\end{table*}

\subsection{Scoring of SvdH JSN}
\label{sec:jsn_score_result_AP}
\subsubsection{Joint-level Results and Confusion Matrices}
\begin{table*}[!t]
\centering
\caption{SvdH JSN score classification QWK results for each joint on the Test set. The best results in each column are highlighted in \textbf{bold}, and the second-best values are \underline{underlined}. Mean denotes the overall QWK computed over all JSN scoring samples.}
\label{tab:jsn_joint_qwk_results}

\begin{subtable}[!t]{\textwidth}
\centering
\resizebox{\textwidth}{!}{
\setlength{\tabcolsep}{12pt}
\begin{tabular}{lcccccccc}
\toprule
\textbf{Model} & \textbf{MCP-T} & \textbf{MCP-I} & \textbf{MCP-M} & \textbf{MCP-R} & \textbf{MCP-S} & \textbf{PIP-I} & \textbf{PIP-M} & \textbf{PIP-R} \\
\midrule
ResNet & \textbf{0.5400} & \underline{0.9124} & 0.7464 & 0.6870 & 0.6739 & 0.3068 & 0.5391 & 0.3344 \\
DenseNet & 0.4686 & 0.8513 & 0.7236 & 0.6918 & 0.6694 & 0.4250 & 0.4572 & 0.4241 \\
EfficientNetV2 & 0.3823 & 0.9042 & 0.7207 & 0.5914 & 0.6612 & 0.5419 & 0.3261 & \underline{0.4720} \\
MobileViT & \underline{0.5188} & \textbf{0.9129} & 0.7759 & \textbf{0.7211} & \textbf{0.7187} & 0.3170 & \textbf{0.5683} & 0.3042 \\
LeViT & 0.4692 & 0.8438 & \textbf{0.8382} & 0.6800 & \underline{0.6742} & 0.4127 & 0.4513 & 0.4056 \\
EfficientFormer & 0.4702 & 0.8763 & 0.7067 & \underline{0.7052} & 0.6587 & 0.5310 & \underline{0.5397} & \textbf{0.5596} \\
ConvNeXtV2 & 0.4619 & 0.8968 & \underline{0.8130} & 0.5962 & 0.4747 & \underline{0.5498} & 0.4453 & 0.4645 \\
MedMamba & 0.4739 & 0.9033 & 0.7181 & 0.7001 & 0.6245 & \textbf{0.5588} & 0.5148 & 0.4405 \\
MambaVision & 0.5058 & 0.7873 & 0.7975 & 0.5752 & 0.5538 & 0.3477 & 0.4300 & 0.3408 \\
\bottomrule
\end{tabular}}
\end{subtable}

\vspace{0.6cm}

\begin{subtable}[!t]{\textwidth}
\centering
\resizebox{\textwidth}{!}{
\setlength{\tabcolsep}{12pt}
\begin{tabular}{lcccccccc}
\toprule
\textbf{Model} & \textbf{PIP-S} & \textbf{STT} & \textbf{SC} & \textbf{SR} & \textbf{CMC-M} & \textbf{CMC-R} & \textbf{CMC-S} & \textbf{Mean} \\
\midrule
ResNet & 0.1058 & \textbf{0.7203} & \textbf{0.5521} & \textbf{0.6464} & \underline{0.5986} & 0.3160 & 0.4843 & 0.5884 \\
DenseNet & \underline{0.3139} & \underline{0.6853} & \underline{0.5138} & 0.6276 & 0.5669 & \textbf{0.3856} & \textbf{0.5042} & 0.5829 \\
EfficientNetV2 & \textbf{0.3676} & 0.6686 & 0.3052 & 0.5380 & 0.4974 & 0.2186 & 0.4048 & 0.5393 \\
MobileViT & 0.1950 & 0.6754 & 0.4546 & 0.6171 & \textbf{0.6316} & 0.3370 & 0.4521 & \textbf{0.5967} \\
LeViT & 0.1214 & 0.5220 & 0.3236 & 0.5668 & 0.4214 & \underline{0.3760} & 0.4691 & 0.5445 \\
EfficientFormer & 0.2928 & 0.6731 & 0.4889 & \underline{0.6283} & 0.5886 & 0.2821 & 0.4644 & \underline{0.5919} \\
ConvNeXtV2 & 0.2189 & 0.4722 & 0.3581 & 0.4618 & 0.3070 & 0.1290 & 0.1737 & 0.5151 \\
MedMamba & 0.2960 & 0.6409 & 0.4303 & 0.6257 & 0.4956 & 0.2622 & \underline{0.4859} & 0.5738 \\
MambaVision & 0.0988 & 0.6745 & 0.4678 & 0.6080 & 0.4954 & 0.3543 & 0.4741 & 0.5457 \\
\bottomrule
\end{tabular}}
\end{subtable}
\end{table*}

\begin{table*}[!t]
\centering
\caption{SvdH JSN score classification BACC results (\%) for each joint on the Test set. The best results in each column are highlighted in \textbf{bold}, and the second-best values are \underline{underlined}. Mean denotes the overall BACC computed over all JSN scoring samples.}
\label{tab:jsn_joint_bacc_results}

\begin{subtable}[!t]{\textwidth}
\centering
\resizebox{\textwidth}{!}{
\setlength{\tabcolsep}{12pt}
\begin{tabular}{lcccccccc}
\toprule
\textbf{Model} & \textbf{MCP-T} & \textbf{MCP-I} & \textbf{MCP-M} & \textbf{MCP-R} & \textbf{MCP-S} & \textbf{PIP-I} & \textbf{PIP-M} & \textbf{PIP-R} \\
\midrule
ResNet & \underline{50.75} & \underline{55.49} & 37.57 & 39.65 & 43.49 & 28.80 & 36.95 & 28.27 \\
DenseNet & 46.13 & 40.82 & 24.85 & 39.83 & 46.92 & \textbf{29.54} & 25.04 & \underline{36.02} \\
EfficientNetV2 & 44.97 & \textbf{56.83} & 27.69 & 23.83 & 39.15 & 27.16 & 19.80 & 27.82 \\
MobileViT & \textbf{53.16} & 54.08 & 35.43 & \textbf{55.13} & \underline{49.30} & 25.74 & 32.70 & 21.68 \\
LeViT & 46.89 & 52.82 & \underline{41.57} & \underline{49.83} & \textbf{53.58} & 26.00 & 31.03 & 28.85 \\
EfficientFormer & 50.04 & 49.08 & \textbf{42.19} & 41.39 & 45.89 & 21.76 & \textbf{41.37} & 31.65 \\
ConvNeXtV2 & 37.99 & 50.91 & 33.91 & 22.61 & 29.35 & 25.32 & 35.03 & \textbf{39.21} \\
MedMamba & 50.48 & 49.41 & 33.74 & 35.25 & 40.45 & 28.82 & \underline{39.21} & 32.07 \\
MambaVision & 45.02 & 43.08 & 35.35 & 21.48 & 35.90 & \underline{29.49} & 30.32 & 23.85 \\
\bottomrule
\end{tabular}}
\end{subtable}

\vspace{0.6cm}

\begin{subtable}[!t]{\textwidth}
\centering
\resizebox{\textwidth}{!}{
\setlength{\tabcolsep}{12pt}
\begin{tabular}{lcccccccc}
\toprule
\textbf{Model} & \textbf{PIP-S} & \textbf{STT} & \textbf{SC} & \textbf{SR} & \textbf{CMC-M} & \textbf{CMC-R} & \textbf{CMC-S} & \textbf{Mean} \\
\midrule
ResNet & 20.53 & 41.09 & 31.56 & 32.32 & 32.52 & 26.56 & \textbf{35.56} & 39.51 \\
DenseNet & 22.01 & 32.53 & 31.56 & 35.52 & 36.41 & \textbf{27.99} & \underline{35.19} & 36.44 \\
EfficientNetV2 & 21.50 & 29.32 & 30.26 & 31.17 & 30.54 & 22.93 & 29.61 & 32.83 \\
MobileViT & \underline{22.51} & \underline{43.62} & 30.82 & \textbf{57.83} & \textbf{56.84} & \underline{27.24} & 27.87 & \textbf{42.80} \\
LeViT & 21.07 & 34.17 & \underline{38.12} & \underline{37.24} & \underline{49.21} & 25.76 & 31.52 & \underline{40.62} \\
EfficientFormer & 21.42 & 33.31 & 33.69 & 28.92 & 35.61 & 25.12 & 33.01 & 39.25 \\
ConvNeXtV2 & 21.78 & 29.48 & 25.55 & 25.04 & 22.42 & 21.39 & 23.56 & 32.51 \\
MedMamba & \textbf{27.88} & 38.10 & 32.86 & 30.43 & 25.57 & 25.11 & 26.16 & 38.66 \\
MambaVision & 19.97 & \textbf{57.73} & \textbf{45.73} & 32.06 & 40.37 & 27.14 & 32.40 & 34.62 \\
\bottomrule
\end{tabular}}
\end{subtable}
\end{table*}

\begin{figure}[!t]
    \centering
    \includegraphics[width=0.97\linewidth]{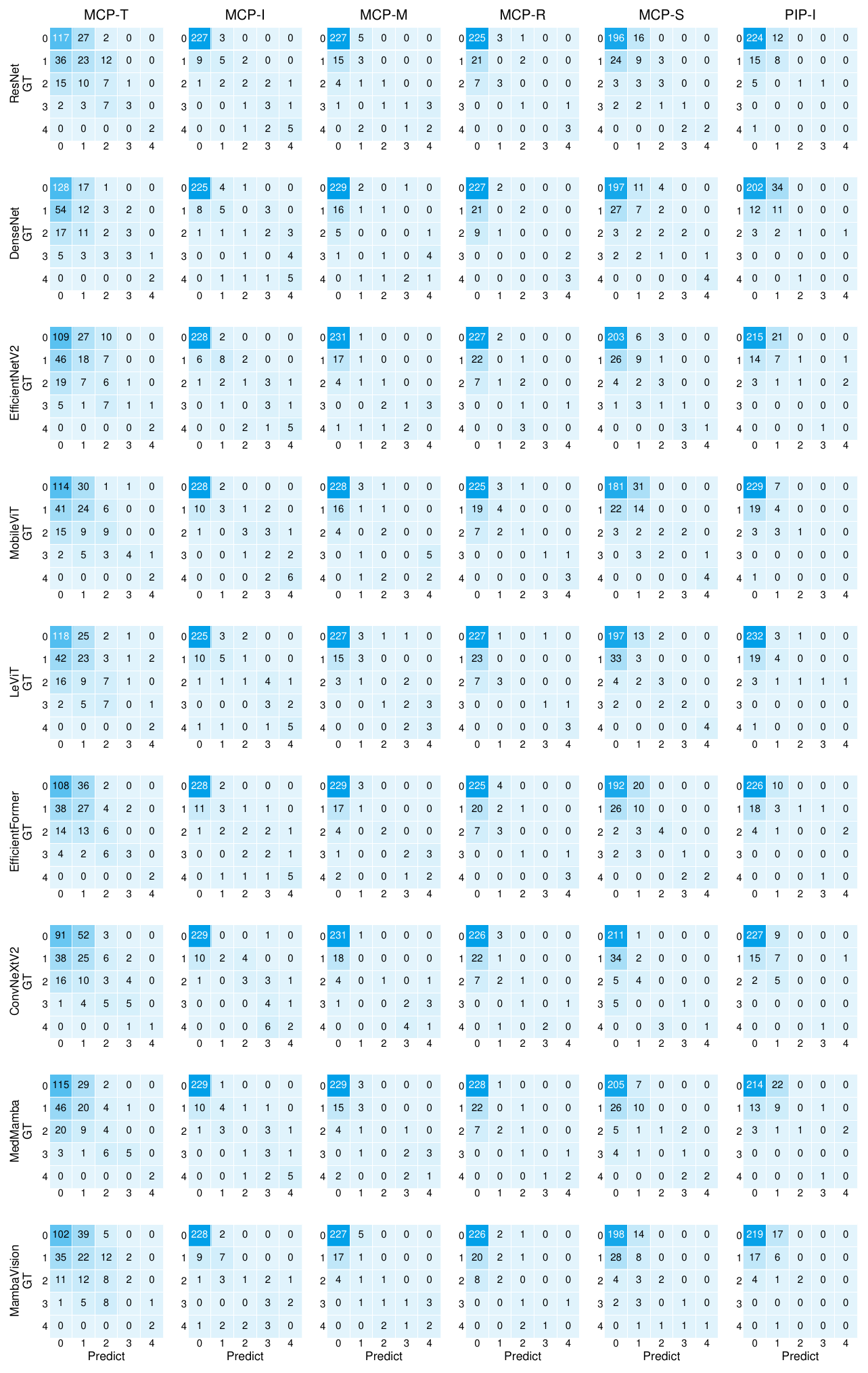}
    \caption{Joint-wise confusion matrices of SvdH JSN scoring (A).}
    \label{fig:jsn_confusion_joint_1_AP}
\end{figure}

\begin{figure}[!t]
    \centering
    \includegraphics[width=0.97\linewidth]{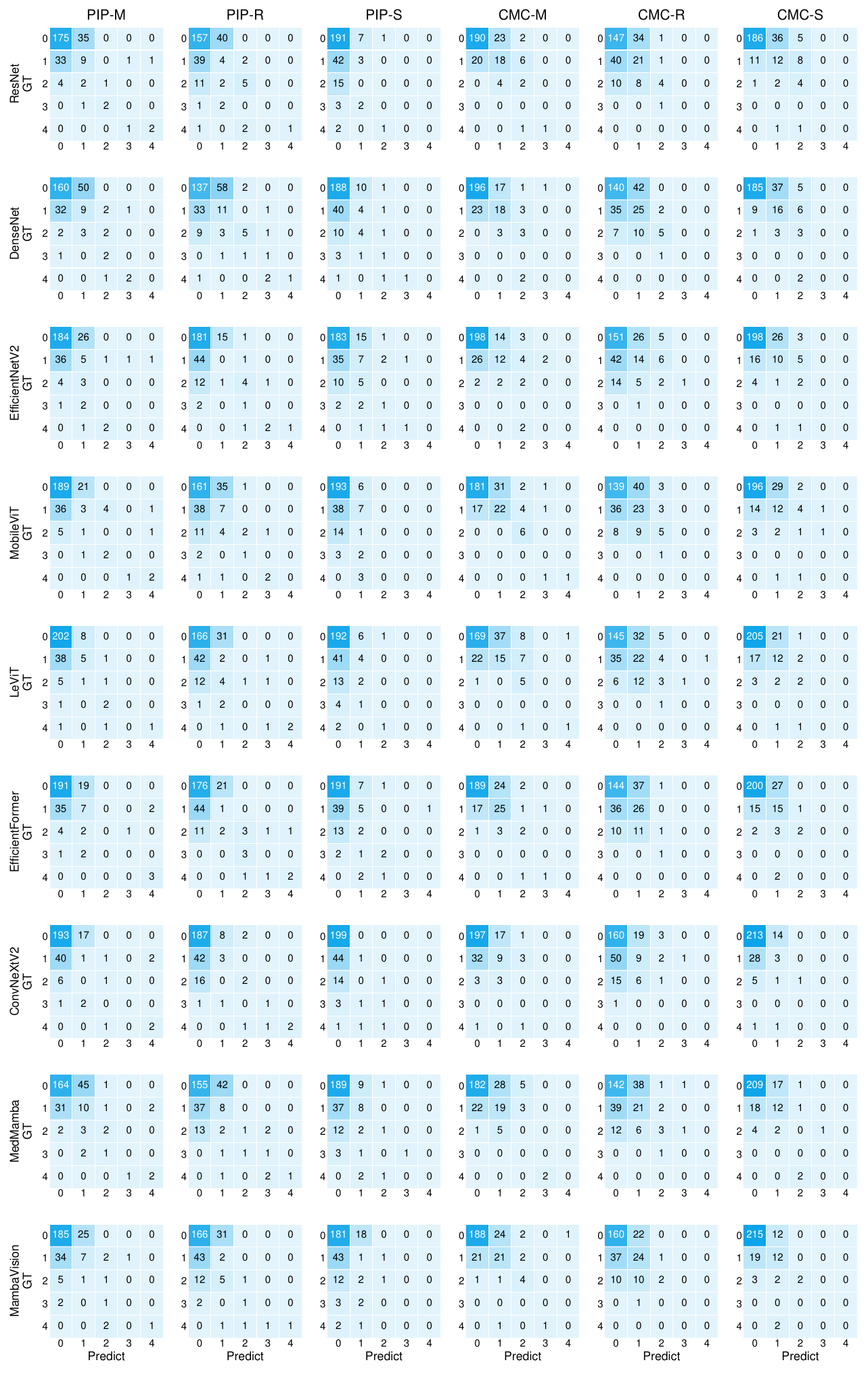}
    \caption{Joint-wise confusion matrices of SvdH JSN scoring (B).}
    \label{fig:jsn_confusion_joint_2_AP}
\end{figure}

\begin{figure}[!t]
    \centering
    \includegraphics[width=0.55\linewidth]{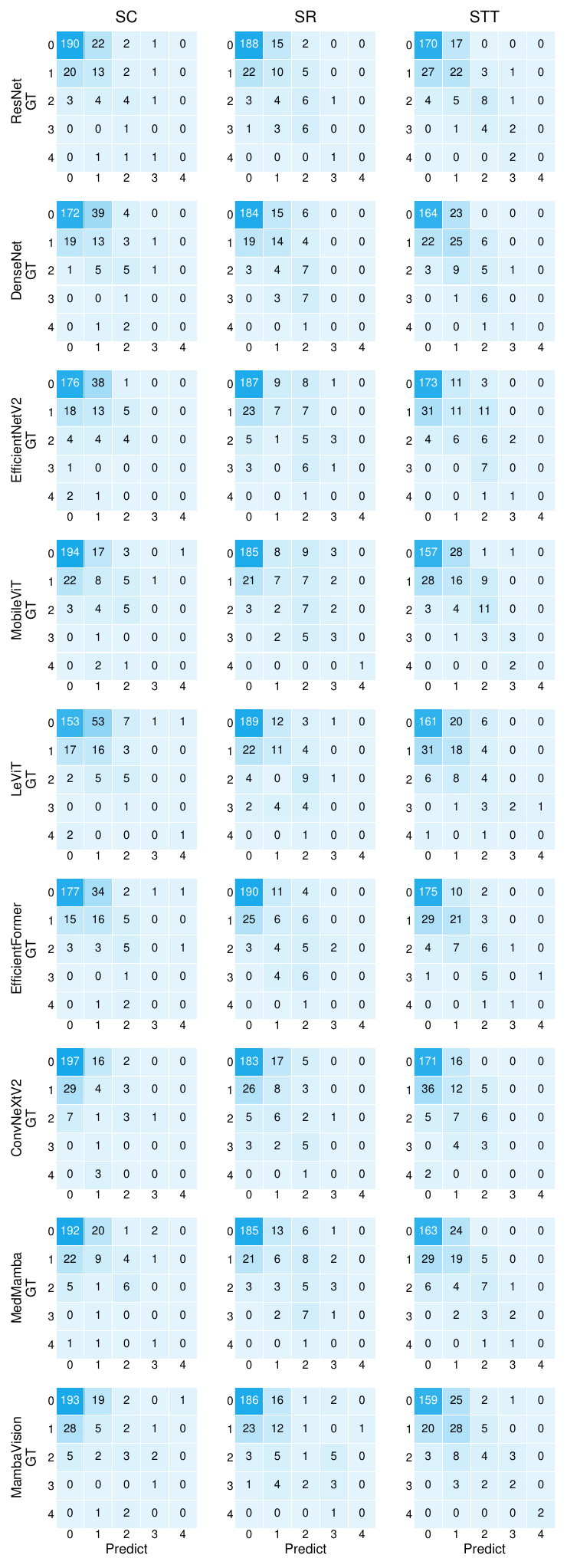}
    \caption{Joint-wise confusion matrices of SvdH JSN scoring (C).}
    \label{fig:jsn_confusion_joint_3_AP}
\end{figure}

Tables~\ref{tab:jsn_joint_qwk_results} and~\ref{tab:jsn_joint_bacc_results} report the joint-level QWK and BACC results for the SvdH JSN scoring task on the test set.
Overall, the results show clear joint-dependent performance differences, suggesting that JSN scoring difficulty varies substantially across anatomical locations.

For joint-level QWK, the models show relatively high agreement on MCP joints and several wrist-related joints, indicating that JSN patterns in these regions can be captured more reliably. In particular, MCP-I, MCP-M, MCP-R, MCP-S, STT, SR, and CMC-M generally achieve stronger ordinal consistency across different architectures. In contrast, PIP joints, especially PIP-S, as well as CMC-R, are more challenging, with noticeably lower QWK values across many models. Among the evaluated methods, MobileViT achieves the best overall QWK reported in the Mean column, while ResNet, DenseNet, EfficientFormer, and MedMamba also show competitive joint-level ordinal agreement.

For joint-level BACC, the results follow a similar trend. MobileViT obtains the strongest overall BACC reported in the Mean column and performs well across both MCP and wrist-related joints, suggesting relatively robust classification ability under class imbalance. ResNet, LeViT, EfficientFormer, and MedMamba also achieve competitive BACC, showing that several architectures can provide reasonably balanced predictions across anatomical sites. However, performance remains uneven across joints, with some PIP and CMC joints showing lower balanced accuracy, indicating that class imbalance and subtle visual differences still affect JSN recognition.

The joint-wise confusion matrices in Figs.~\ref{fig:jsn_confusion_joint_1_AP}, \ref{fig:jsn_confusion_joint_2_AP}, and~\ref{fig:jsn_confusion_joint_3_AP} further illustrate these joint-specific patterns. Joints with stronger quantitative performance tend to show clearer diagonal alignment, while more difficult joints have predictions concentrated in lower JSN scores or confused between neighboring grades. This suggests that many errors come from mild or moderate narrowing cases, where the visual difference between adjacent JSN grades is subtle.

These joint-level results further show that JSN classification is highly joint-dependent. While several models achieve strong performance on MCP and wrist-related joints, smaller or more ambiguous joints remain difficult. Future work should therefore focus on ordinal-aware learning strategies that better preserve the ordered nature of JSN grades, together with joint-specific modeling approaches that account for anatomical differences across regions. More effective solutions for class imbalance, such as cost-sensitive learning, adaptive re-weighting, or balanced sampling, may also help improve sensitivity to underrepresented positive grades. Incorporating anatomical priors, multi-joint contextual information, and uncertainty-aware prediction could further improve robustness, especially for subtle or ambiguous JSN changes.

\begin{table*}[!t]
\centering
\caption{
Per-grade recall (Rec.) and precision (Prec.) for SvdH JSN scoring on the complete test set.
Values are reported as percentages (\%).
}
\label{tab:jsn_pergrade_AP}
\resizebox{\textwidth}{!}{
\setlength{\tabcolsep}{9pt}
\begin{tabular}{lcccccccccc}
\toprule
& \multicolumn{2}{c}{\textbf{Grade 0}}
& \multicolumn{2}{c}{\textbf{Grade 1}}
& \multicolumn{2}{c}{\textbf{Grade 2}}
& \multicolumn{2}{c}{\textbf{Grade 3}}
& \multicolumn{2}{c}{\textbf{Grade 4}} \\
\cmidrule(lr){2-3}
\cmidrule(lr){4-5}
\cmidrule(lr){6-7}
\cmidrule(lr){8-9}
\cmidrule(lr){10-11}
\textbf{Model}
& \textbf{Rec.} & \textbf{Prec.}
& \textbf{Rec.} & \textbf{Prec.}
& \textbf{Rec.} & \textbf{Prec.}
& \textbf{Rec.} & \textbf{Prec.}
& \textbf{Rec.} & \textbf{Prec.} \\
\midrule
ResNet
& 90.01 & 85.57
& 27.40 & 30.59
& 25.00 & 33.80
& 15.62 & 31.25
& 39.53 & 70.83 \\
DenseNet
& 87.57 & 85.65
& 29.28 & 28.17
& 21.88 & 30.00
& 6.25 & 12.12
& 37.21 & 48.48 \\
EfficientNetV2
& 91.10 & 84.62
& 20.89 & 29.19
& 20.31 & 22.67
& 10.94 & 20.59
& 20.93 & 45.00 \\
MobileViT
& 89.69 & 85.65
& 26.54 & 30.21
& 28.65 & 37.93
& 20.31 & 30.23
& 48.84 & 60.00 \\
LeViT
& 89.94 & 84.55
& 24.49 & 29.92
& 21.88 & 30.66
& 15.62 & 31.25
& 51.16 & 59.46 \\
EfficientFormer
& 91.00 & 85.52
& 28.77 & 32.94
& 19.79 & 34.55
& 12.50 & 27.59
& 44.19 & 55.88 \\
ConvNeXtV2
& 93.37 & 83.02
& 14.90 & 25.36
& 15.10 & 30.21
& 9.38 & 27.27
& 27.91 & 54.55 \\
MedMamba
& 89.72 & 85.60
& 26.03 & 29.92
& 23.96 & 32.86
& 25.00 & 37.21
& 41.86 & 60.00 \\
MambaVision
& 91.35 & 84.51
& 20.55 & 30.17
& 22.40 & 31.16
& 12.50 & 25.81
& 39.53 & 56.67 \\
\bottomrule
\end{tabular}
}
\end{table*}

\begin{table*}[!t]
\centering
\caption{
Patient-level bootstrap 95\% confidence intervals for per-grade SvdH JSN scoring ($B=2000$).
Pred.\ $n$ denotes the predicted-class count and therefore the denominator used to compute precision.
Values are reported as percentages (\%).
}
\label{tab:jsn_pergrade_ci_AP}

\resizebox{\textwidth}{!}{
\begin{tabular}{llccccc}
\toprule
\textbf{Model} & \textbf{Metric} &
\textbf{Grade 0} & \textbf{Grade 1} & \textbf{Grade 2} &
\textbf{Grade 3} & \textbf{Grade 4} \\
\midrule

\multirow{3}{*}{ResNet}
& Rec.\ CI
& [85.92, 94.31]
& [20.63, 32.89]
& [16.13, 32.00]
& [3.80, 27.78]
& [8.57, 66.67] \\
& Prec.\ CI
& [82.64, 88.24]
& [22.50, 42.00]
& [24.84, 43.53]
& [7.69, 55.00]
& [20.00, 94.56] \\
& Pred.\ $n$
& 3284 & 523 & 142 & 32 & 24 \\
\midrule

\multirow{3}{*}{DenseNet}
& Rec.\ CI
& [82.17, 92.73]
& [21.72, 35.70]
& [14.86, 28.64]
& [0.00, 10.45]
& [10.00, 58.82] \\
& Prec.\ CI
& [82.69, 88.45]
& [21.53, 38.36]
& [22.85, 38.54]
& [0.00, 20.51]
& [10.00, 81.25] \\
& Pred.\ $n$
& 3192 & 607 & 140 & 33 & 33 \\
\midrule

\multirow{3}{*}{EfficientNetV2}
& Rec.\ CI
& [87.15, 94.89]
& [15.85, 25.34]
& [12.42, 26.64]
& [0.00, 22.58]
& [9.08, 46.17] \\
& Prec.\ CI
& [81.56, 87.49]
& [19.73, 42.18]
& [13.66, 33.98]
& [0.00, 45.45]
& [14.29, 70.00] \\
& Pred.\ $n$
& 3361 & 418 & 172 & 34 & 20 \\
\midrule

\multirow{3}{*}{MobileViT}
& Rec.\ CI
& [86.11, 93.55]
& [20.42, 32.02]
& [19.53, 36.89]
& [6.49, 31.82]
& [28.57, 69.23] \\
& Prec.\ CI
& [82.30, 88.69]
& [23.63, 38.79]
& [28.46, 47.25]
& [11.76, 46.15]
& [25.00, 82.05] \\
& Pred.\ $n$
& 3269 & 513 & 145 & 43 & 35 \\
\midrule

\multirow{3}{*}{LeViT}
& Rec.\ CI
& [87.37, 92.99]
& [18.63, 29.33]
& [13.97, 28.23]
& [2.94, 29.55]
& [8.33, 68.18] \\
& Prec.\ CI
& [81.16, 87.40]
& [23.01, 38.17]
& [22.88, 38.68]
& [7.13, 55.00]
& [7.13, 79.69] \\
& Pred.\ $n$
& 3321 & 478 & 137 & 32 & 37 \\
\midrule

\multirow{3}{*}{EfficientFormer}
& Rec.\ CI
& [87.42, 94.48]
& [21.69, 34.50]
& [11.24, 27.54]
& [1.64, 25.00]
& [13.79, 66.67] \\
& Prec.\ CI
& [82.24, 88.42]
& [25.65, 43.16]
& [23.08, 45.19]
& [5.26, 48.15]
& [14.29, 85.71] \\
& Pred.\ $n$
& 3322 & 510 & 110 & 29 & 34 \\
\midrule

\multirow{3}{*}{ConvNeXtV2}
& Rec.\ CI
& [91.37, 95.63]
& [10.96, 18.14]
& [7.69, 17.76]
& [8.33, 33.82]
& [5.00, 46.15] \\
& Prec.\ CI
& [79.39, 86.09]
& [19.91, 32.57]
& [19.30, 37.68]
& [15.00, 58.33]
& [10.53, 80.00] \\
& Pred.\ $n$
& 3511 & 343 & 91 & 41 & 19 \\
\midrule

\multirow{3}{*}{MedMamba}
& Rec.\ CI
& [85.81, 93.63]
& [22.01, 30.84]
& [9.88, 23.96]
& [10.52, 40.91]
& [4.88, 58.82] \\
& Prec.\ CI
& [81.77, 87.79]
& [22.66, 40.79]
& [18.52, 42.98]
& [12.49, 48.15]
& [10.00, 85.71] \\
& Pred.\ $n$
& 3300 & 518 & 107 & 55 & 25 \\
\midrule

\multirow{3}{*}{MambaVision}
& Rec.\ CI
& [86.35, 94.99]
& [18.04, 34.84]
& [10.46, 22.58]
& [5.66, 26.79]
& [11.32, 53.33] \\
& Prec.\ CI
& [81.76, 88.25]
& [23.95, 39.85]
& [23.30, 42.19]
& [7.81, 52.95]
& [17.64, 72.73] \\
& Pred.\ $n$
& 3327 & 516 & 101 & 41 & 20 \\

\bottomrule
\end{tabular}
}
\end{table*}

\subsubsection{Per-grade Scoring Analysis}
\label{sec:jsn_pergrade_AP}
To further examine performance across individual SvdH-JSN grades, we report per-grade recall and precision for all scoring models in Table~\ref{tab:jsn_pergrade_AP}. We additionally report 95\% confidence intervals for the higher grades (grades 2, 3, and 4) in Table~\ref{tab:jsn_pergrade_ci_AP}, estimated using patient-level cluster bootstrap ($B=2000$) by resampling patients while retaining all associated hands and joints. The corresponding numbers of predictions assigned to each grade are also reported as the denominators for the precision estimates. The results show clear grade-specific differences that are not fully reflected by the aggregate scoring metrics. In particular, grade-3 recall remains low across all nine models, ranging from 6.25\% to 25.00\%, despite having 64 test samples. In comparison, the less frequent grade 4 achieves recall between 20.93\% and 51.16\%, with a higher point estimate than grade 3 for every model. Performance therefore does not decrease monotonically with either grade severity or class frequency, but instead exhibits substantial class-specific confusion. The confidence intervals for the higher grades are generally broad, particularly for grade 4, which contains only 43 test samples, reflecting greater uncertainty in the performance estimates for these less represented grades.

\begin{table*}[!t]
\centering
\caption{Spearman correlation results between predicted SvdH JSN scores and ground-truth SvdH JSN scores on the Test set. Significance levels: $*p<0.05$, $**p<0.01$, $***p<0.001$.}
\label{tab:stat_eva_jsn_AP}
\begin{threeparttable}
\resizebox{0.6\linewidth}{!}{
\setlength{\tabcolsep}{8pt}
\begin{tabular}{lccc}
\toprule
\textbf{Model} 
& \textbf{Spearman $\rho$} 
& \textbf{$p$-value}
& \textbf{Significance} \\
\midrule
ResNet & 0.5610 & $<0.001$ & $***$ \\
DenseNet & 0.5395 & $<0.001$ & $***$ \\
EfficientNetV2 & 0.5580 & $<0.001$ & $***$ \\
MobileViT & 0.5765 & $<0.001$ & $***$ \\
LeViT & 0.5658 & $<0.001$ & $***$ \\
EfficientFormer & 0.5986 & $<0.001$ & $***$ \\
ConvNeXtV2 & 0.4792 & $<0.001$ & $***$ \\
MedMamba & 0.5308 & $<0.001$ & $***$ \\
MambaVision & 0.5184 & $<0.001$ & $***$ \\
\bottomrule
\end{tabular}
}
\end{threeparttable}
\end{table*}

\subsubsection{Correlation Analysis Between Predicted and Ground-Truth SvdH JSN Scores}
The relationship between predicted JSN scores and reference total JSN scores was analyzed using Spearman's rank correlation. As shown in Table~\ref{tab:stat_eva_jsn_AP}, all models demonstrated statistically significant positive correlations with the clinical JSN scores ($p<0.001$), indicating that the predicted scores preserved the relative ordering of joint space narrowing severity. The strongest correlation was achieved by EfficientFormer ($\rho=0.5986$), followed by MobileViT ($\rho=0.5765$), LeViT ($\rho=0.5658$), ResNet ($\rho=0.5610$), and EfficientNetV2 ($\rho=0.5580$). ConvNeXtV2 showed the lowest correlation among the evaluated models, but still maintained a significant positive association ($\rho=0.4792$, $p<0.001$).

These results suggest that the JSN score prediction models capture clinically meaningful ranking information related to joint space narrowing severity. Compared with BE score prediction, the JSN correlations were generally higher, indicating better consistency between model predictions and expert-derived JSN scores. Nevertheless, the correlations remain moderate rather than near-perfect, suggesting that predicted JSN scores should be interpreted as supportive quantitative estimates rather than direct substitutes for expert clinical assessment.

\subsection{Hand-Level SvdH Score Aggregation}
\label{app:hand_level_aggregation}

To evaluate whether joint-level SvdH predictions can be summarized into clinically interpretable hand-level severity measures, we further aggregate the joint-level predictions on the test set. For each of the 267 test hands, Total BE is computed by summing the predicted grades over the 16 BE ROIs, while Total JSN is computed over the 15 JSN ROIs. Total SvdH is then obtained as the sum of Total BE and Total JSN. The same aggregation is applied to the reference joint-level scores. For each target, we report the MAE and Pearson correlation coefficient ($r$) between the predicted and reference hand-level totals. This analysis uses ground-truth ROI crops and therefore evaluates the scoring-to-aggregation stage independently of ROI localization error.

\begin{table*}[!t]
\centering
\caption{
Hand-level aggregation of joint-level SvdH predictions on the test set
($N=267$ hands). Total BE and Total JSN are obtained by summing the
predicted grades across the corresponding ROIs, and Total SvdH is
their sum. MAE and Pearson correlation ($r$) are computed against the
corresponding reference hand-level totals. Ground-truth ROI crops are
used in this analysis.
}
\label{tab:hand_level_aggregation}
\resizebox{\textwidth}{!}{
\setlength{\tabcolsep}{12pt}
\begin{tabular}{lcccccc}
\toprule
& \multicolumn{2}{c}{Total SvdH}
& \multicolumn{2}{c}{Total BE}
& \multicolumn{2}{c}{Total JSN} \\
\cmidrule(lr){2-3}
\cmidrule(lr){4-5}
\cmidrule(lr){6-7}
Model
& MAE $\downarrow$ & Pearson $r$ $\uparrow$
& MAE $\downarrow$ & Pearson $r$ $\uparrow$
& MAE $\downarrow$ & Pearson $r$ $\uparrow$ \\
\midrule
ResNet         & 5.1648 & 0.8403 & 3.6929 & 0.7616 & 2.7453 & 0.7627 \\
DenseNet         & 6.3258 & 0.8416 & 5.1498 & 0.7381 & 2.9213 & 0.7352 \\
EfficientNetV2   & 6.5356 & 0.7909 & 4.7903 & 0.6424 & 3.0337 & 0.7334 \\
MobileViT        & 5.1461 & 0.8158 & 3.8127 & 0.6889 & 2.7416 & 0.7725 \\
LeViT            & 6.4195 & 0.7577 & 4.8764 & 0.5314 & 2.8015 & 0.7603 \\
EfficientFormer  & 6.4232 & 0.8102 & 4.8015 & 0.6470 & 2.8652 & 0.7700 \\
ConvNeXtV2       & 7.0150 & 0.7561 & 4.6929 & 0.5587 & 3.1236 & 0.7571 \\
MedMamba         & 5.3596 & 0.8258 & 3.6854 & 0.7554 & 2.9101 & 0.7549 \\
MambaVision    & 6.8614 & 0.7976 & 5.1049 & 0.6754 & 2.9850 & 0.7311 \\
\bottomrule
\end{tabular}}
\end{table*}

As shown in Table~\ref{tab:hand_level_aggregation}, the aggregated Total SvdH predictions show Pearson correlations of $r=0.7561$--$0.8416$ with the reference hand-level totals across the nine models. Total JSN shows a comparatively narrow range ($r=0.7311$--$0.7725$), whereas Total BE is more variable across
backbones ($r=0.5314$--$0.7616$), consistent with the greater difficulty of BE scoring observed in the joint-level benchmark.

Importantly, this experiment evaluates within-task aggregation from joint-level SvdH predictions to hand-level scores. It does not constitute evidence of information transfer or performance improvement across the separate bone segmentation, BE segmentation, and SvdH scoring tasks.

\subsection{Center-Held-Out Evaluation on SARC}
\label{sec:sarc_heldout_AP}
To further assess robustness to institutional variation, we conducted a center-held-out evaluation using SARC as the unseen test center. SARC was selected because it is the second-largest contributing institution in RAM-H1200, accounting for 368 of the 1,200 radiographs (30.7\%), corresponding to 5,888 BE joints and 5,520 JSN joints. In addition, SARC is the only contributing center using a KONICA MINOLTA imaging system, with an acquisition configuration distinct from that of HMCRD, the largest contributing center. These characteristics provide a sufficiently large held-out cohort while introducing a meaningful acquisition-domain shift. All 368 SARC images were excluded from model development and used exclusively for testing.
The remaining data were divided at the patient level into 684 training and 148 validation images, corresponding to 208 and 38 patients, respectively, while the SARC test set contained 45 patients. No patient appeared in more than one split.

We evaluated the models from each of the four benchmark tasks under this center-held-out setting, including hand bone structure segmentation, hand BE segmentation, SvdH BE scoring, and SvdH JSN scoring. The results are reported separately for each task below.

\subsubsection{Hand Bone Structure Segmentation}
\label{sec:sarc_bone_seg_AP}

Table~\ref{tab:sarc_bone_seg_AP} reports hand bone structure segmentation performance when SARC is held out as the unseen test center. Compared with the main benchmark (Table~\ref{tab:seg_results}), cross-center performance exhibits clear architecture-dependent variation. The DSC range of the supervised models broadens from 96.37--97.32\% in the main benchmark to 89.65--96.04\% on SARC. Nevertheless, the strongest model remains relatively robust: SwinUMamba achieves the best DSC (96.04\%) and NSD (90.11\%), with its DSC decreasing by only 1.27 percentage points from 97.31\% in the main benchmark. In contrast, larger degradation is observed for several other architectures, indicating that robustness to the held-out center is not uniform across models. Interestingly, overlap-region performance does not show the same degradation. SwinUMamba achieves DSC$_O$/NSD$_O$ of 78.59\%/81.55\% on SARC, compared with 76.32\%/77.68\% in the main benchmark. This suggests that the cross-center shift primarily affects overall structural delineation rather than uniformly increasing the difficulty of anatomically overlapping regions. Overall, the results indicate that high in-distribution segmentation accuracy does not necessarily translate into equivalent cross-center robustness, while the best-performing architecture retains relatively strong performance on the unseen SARC center.

\begin{table*}[!t]
\centering
\caption{Hand bone structure segmentation results when SARC is held out as the unseen test center. The best results in each column are highlighted in \textbf{bold}, and the second-best values are \underline{underlined}.}
\begin{threeparttable}
\resizebox{\textwidth}{!}{
\setlength{\tabcolsep}{7pt}
\begin{tabular}{lcccccc}
\toprule
\textbf{Model} 
& \textbf{DSC (\%)} 
& \textbf{NSD (\%)} 
& \textbf{DSC$_O$ (\%)} 
& \textbf{NSD$_O$ (\%)} 
& \textbf{VOE (\%)} 
& \textbf{MSD (pix)} \\
\midrule
\multicolumn{7}{c}{\textbf{Supervised Models}} \\
\midrule

Unet
& 92.27$\pm$5.42
& 83.15$\pm$8.36
& 76.12$\pm$7.12
& 77.99$\pm$12.48
& 11.95$\pm$6.87
& 20.82$\pm$25.70 \\

Unet++
& 89.65$\pm$7.51
& 75.19$\pm$12.45
& \underline{77.73$\pm$6.36}
& 78.68$\pm$10.78
& 15.59$\pm$9.53
& 82.09$\pm$66.76 \\

SegFormer
& \underline{95.19$\pm$3.59}
& 87.71$\pm$6.14
& 73.99$\pm$5.54
& 73.49$\pm$10.35
& \underline{8.12$\pm$4.68}
& \textbf{11.13$\pm$21.12} \\

TransUnet
& 94.11$\pm$4.86
& \underline{88.43$\pm$7.23}
& 75.86$\pm$5.48
& 79.36$\pm$10.17
& 9.01$\pm$6.03
& \underline{15.36$\pm$20.35} \\

SwinUNETR
& 94.71$\pm$3.89
& 85.30$\pm$7.45
& 78.16$\pm$5.72
& \underline{80.13$\pm$9.56}
& 8.70$\pm$5.35
& 37.21$\pm$32.71 \\

UMambaEnc
& 91.10$\pm$7.80
& 78.86$\pm$12.51
& 74.52$\pm$10.48
& 75.38$\pm$14.97
& 13.40$\pm$9.91
& 89.18$\pm$85.18 \\

SwinUMamba
& \textbf{96.04$\pm$4.14}
& \textbf{90.11$\pm$7.58}
& \textbf{78.59$\pm$7.22}
& \textbf{81.55$\pm$11.37}
& \textbf{6.58$\pm$5.76}
& 29.58$\pm$47.97 \\

MambaVision
& 94.71$\pm$3.22
& 87.09$\pm$5.75
& 73.36$\pm$6.52
& 73.25$\pm$11.17
& 8.82$\pm$4.21
& 15.82$\pm$19.73 \\

\midrule
\multicolumn{7}{c}{\textbf{Foundation Models}} \\
\midrule

SAM(Box)
& 92.95$\pm$2.20
& 79.40$\pm$4.41
& 7.02$\pm$3.64
& 4.74$\pm$3.65
& 11.50$\pm$2.57
& 3.30$\pm$2.17 \\

SAM(Point)
& 75.52$\pm$8.79
& 59.14$\pm$8.67
& 4.02$\pm$1.97
& 3.34$\pm$2.13
& 33.49$\pm$9.73
& 42.35$\pm$29.35 \\

MedSAM(Box)
& 75.45$\pm$3.14
& 26.03$\pm$4.74
& 8.58$\pm$4.27
& 6.20$\pm$3.03
& 37.74$\pm$3.94
& 12.06$\pm$2.38 \\

\bottomrule
\end{tabular}
}
\end{threeparttable}
\label{tab:sarc_bone_seg_AP}
\end{table*}

\begin{table*}[!t]
\centering
\caption{SvdH-BE-90 segmentation results with SARC held out as the unseen test center. The best results in each column are highlighted in \textbf{bold}, and the second-best values are \underline{underlined}.}
\label{tab:sarc_be_seg_AP}
\resizebox{\textwidth}{!}{
\setlength{\tabcolsep}{6pt}
\begin{tabular}{lcccccc}
\toprule
\textbf{Model} &
\textbf{DSC (\%)} &
\textbf{NSD (\%)} &
\textbf{REC (\%)} &
\textbf{PREC (\%)} &
\textbf{VOE (\%)} &
\textbf{MSD (pix)} \\
\midrule
Unet
& 13.18 $\pm$ 13.61
& 9.31 $\pm$ 13.97
& 12.37 $\pm$ 21.37
& 7.53 $\pm$ 12.55
& 92.32 $\pm$ 8.64
& 244.66 $\pm$ 149.55 \\

Unet++
& 12.92 $\pm$ 11.60
& 8.74 $\pm$ 11.94
& \textbf{15.70 $\pm$ 24.71}
& 6.24 $\pm$ 9.69
& 92.67 $\pm$ 6.91
& 297.39 $\pm$ 144.64 \\

nnUnet
& 8.45 $\pm$ 15.93
& 8.98 $\pm$ 18.35
& 3.69 $\pm$ 10.39
& \textbf{12.73 $\pm$ 28.91}
& 94.71 $\pm$ 10.70
& 285.03 $\pm$ 186.45 \\

TransUnet
& \textbf{15.23 $\pm$ 14.76}
& \textbf{10.83 $\pm$ 15.35}
& 13.23 $\pm$ 22.38
& \underline{9.30 $\pm$ 15.04}
& \textbf{91.01 $\pm$ 9.43}
& 270.20 $\pm$ 167.82 \\

SegFormer
& 10.27 $\pm$ 11.21
& 7.20 $\pm$ 11.02
& 11.65 $\pm$ 20.92
& 5.05 $\pm$ 8.78
& 94.19 $\pm$ 6.73
& \underline{242.67 $\pm$ 123.14} \\

SwinUNETR
& 12.77 $\pm$ 11.78
& 8.71 $\pm$ 12.01
& \underline{15.05 $\pm$ 23.84}
& 6.29 $\pm$ 9.96
& 92.74 $\pm$ 7.13
& 303.84 $\pm$ 151.84 \\

UMambaEnc
& 13.96 $\pm$ 13.66
& 9.35 $\pm$ 13.61
& 14.65 $\pm$ 24.73
& 7.23 $\pm$ 11.64
& 91.88 $\pm$ 8.48
& 286.12 $\pm$ 160.96 \\

SwinUMamba
& \underline{14.09 $\pm$ 14.78}
& \underline{10.39 $\pm$ 15.70}
& 12.39 $\pm$ 22.38
& 8.35 $\pm$ 13.27
& \underline{91.65 $\pm$ 9.79}
& \textbf{238.27 $\pm$ 147.42} \\
\bottomrule
\end{tabular}
}
\end{table*}

\subsubsection{Hand BE Segmentation}
\label{sec:sarc_be_seg_AP}

Table~\ref{tab:sarc_be_seg_AP} reports SvdH-BE-90 segmentation performance when SARC is held out as the unseen test center. Compared with the main benchmark in Table~\ref{tab:beseg_results}, BE segmentation performance generally decreases under the SARC-held-out setting. In the main benchmark, the best DSC and NSD are 19.59\% and 19.93\%, respectively, both achieved by SwinUMamba; under the held-out-center setting, the corresponding best values decrease to 15.23\% and 10.83\%, achieved by TransUnet. SwinUMamba itself decreases from 19.59\% to 14.09\% in DSC and from 19.93\% to 10.39\% in NSD. Similar reductions are observed across most architectures, particularly for NSD, indicating reduced overlap and boundary agreement on the unseen SARC center. The model ranking also changes between the two evaluations, with TransUnet outperforming SwinUMamba on both DSC and NSD under the SARC-held-out setting. Overall, these results indicate that the already challenging BE segmentation task is further affected by the evaluated center shift, although the magnitude of degradation remains architecture dependent.

\subsubsection{SvdH BE Scoring}
\label{sec:sarc_be_scoring_AP}

Table~\ref{tab:sarc_be_scoring_AP} reports SvdH BE scoring performance when SARC is held out as the unseen test center. Compared with the main benchmark in Table~\ref{tab:be_results}, the most evident degradation occurs in QWK. The best QWK decreases from 0.4522 in the main benchmark to 0.3472 under the SARC-held-out setting, and the model ranking also changes, with DenseNet achieving the highest held-out QWK instead of MedMamba. In contrast, exact and tolerance-based classification metrics remain broadly comparable to those of the main benchmark: ACC remains around 70\% across most models and W1-ACC remains above 90\%. The P/N ACC metrics likewise show model-dependent changes rather than a uniform deterioration. These results suggest that the center shift has a stronger effect on fine-grained ordinal consistency than on coarse exact/tolerance-based or positive/negative classification performance. Nevertheless, the substantial variation across architectures indicates that cross-center robustness for BE scoring remains model dependent.

\begin{table}[!t]
\centering
\caption{SvdH BE scoring performance on the center-held-out SARC test set. The best result in each column is highlighted in \textbf{bold}, and the second-best value is \underline{underlined}.}
\label{tab:sarc_be_scoring_AP}
\resizebox{\linewidth}{!}{
\setlength{\tabcolsep}{7pt}
\begin{tabular}{lccccccc}
\toprule
\textbf{Model} & \textbf{QWK} & \textbf{MAE} & \textbf{BACC (\%)} & \textbf{ACC (\%)} & \textbf{W1-ACC (\%)} & \textbf{P/N-SEN (\%)} & \textbf{P/N-ACC (\%)} \\
\midrule
ResNet        & \underline{0.3455} & 0.3874 & 34.39 & 70.04 & 94.19 & \underline{35.93} & 73.47 \\
DenseNet        & \textbf{0.3472} & 0.4049 & \textbf{35.22} & 67.99 & 94.29 & \textbf{39.51} & 71.60 \\
EfficientNetV2  & 0.2447 & \underline{0.3762} & 30.84 & 71.55 & 93.50 & 21.51 & 73.52 \\
MobileViT       & 0.2993 & 0.4066 & \underline{34.72} & 68.75 & 93.50 & 34.96 & 71.65 \\
LeViT           & 0.2264 & 0.3939 & 31.13 & \underline{71.64} & 92.39 & 18.58 & 73.51 \\
EfficientFormer & 0.3185 & 0.3808 & 32.34 & 69.79 & \textbf{94.50} & 35.22 & 73.28 \\
ConvNeXtV2      & 0.2023 & 0.3911 & 26.97 & 70.47 & 92.85 & 20.21 & 72.35 \\
MedMamba        & 0.2992 & 0.3954 & 30.78 & 70.67 & 93.19 & 28.78 & \underline{73.61} \\
MambaVision   & 0.2546 & \textbf{0.3684} & 29.11 & \textbf{72.96} & 92.90 & 16.57 & \textbf{74.83} \\
\bottomrule
\end{tabular}
}
\end{table}

\subsubsection{SvdH JSN Scoring}
\label{sec:sarc_jsn_scoring_AP}

Table~\ref{tab:sarc_jsn_scoring_AP} reports SvdH JSN scoring performance when SARC is held out as the unseen test center. Compared with the main benchmark in Table~\ref{tab:jsn_results}, JSN scoring shows relatively limited degradation in QWK under the held-out-center setting. The best QWK decreases from 0.5967 in the main benchmark to 0.5571 on SARC, while most models remain within a relatively narrow QWK range. In contrast to BE scoring, where the reduction in QWK is more pronounced, these results suggest that ordinal JSN scoring is comparatively less sensitive to the evaluated center shift. Exact and tolerance-based classification performance also remains broadly comparable: ACC remains around 70--75\% across models and W1-ACC remains above 95\%. Nevertheless, individual metrics and model rankings vary between the main and SARC-held-out evaluations, indicating that cross-center effects are architecture- and metric-dependent rather than a uniform deterioration across JSN scoring performance.

\begin{table}[!t]
\centering
\caption{SvdH JSN scoring performance on the center-held-out SARC test set. The best result in each column is highlighted in \textbf{bold}, and the second-best value is \underline{underlined}.}
\label{tab:sarc_jsn_scoring_AP}
\resizebox{\linewidth}{!}{
\setlength{\tabcolsep}{7pt}
\begin{tabular}{lccccccc}
\toprule
\textbf{Model} & \textbf{QWK} & \textbf{MAE} & \textbf{BACC (\%)} & \textbf{ACC (\%)} & \textbf{W1-ACC (\%)} & \textbf{P/N-SEN (\%)} & \textbf{P/N-ACC (\%)} \\
\midrule
ResNet
& 0.5399
& 0.3962
& \underline{39.54}
& 64.22
& \underline{96.70}
& \textbf{65.83}
& 71.54 \\

DenseNet
& \underline{0.5433}
& 0.3766
& \textbf{42.88}
& 67.23
& 96.11
& 61.23
& 74.22 \\

EfficientNetV2
& 0.4666
& 0.3692
& 35.77
& \underline{69.82}
& 94.40
& 46.86
& 75.29 \\

MobileViT
& 0.5411
& \underline{0.3612}
& 37.17
& 68.75
& 95.72
& 56.42
& \underline{75.67} \\

LeViT
& 0.4034
& 0.3973
& 30.70
& 66.87
& 94.51
& 49.44
& 72.77 \\

EfficientFormer
& 0.5358
& 0.3665
& 36.02
& 67.12
& \textbf{96.74}
& \underline{64.09}
& 74.31 \\

ConvNeXtV2
& 0.4071
& 0.4034
& 30.31
& 66.63
& 94.20
& 44.42
& 72.12 \\

MedMamba
& \textbf{0.5554}
& \textbf{0.3377}
& 36.71
& \textbf{70.38}
& 96.30
& 59.14
& \textbf{77.03} \\

MambaVision
& 0.4798
& 0.3737
& 33.97
& 68.46
& 95.13
& 51.81
& 75.09 \\
\bottomrule
\end{tabular}
}
\end{table}

\subsection{End-to-End Automatic ROI Localization and SvdH Scoring}
To evaluate the practical applicability of the benchmark under automatic localization, all scoring models were evaluated using ROIs detected by the YOLOv12-m joint detector instead of Ground-Truth joint annotations.

\subsubsection{Automatic Joint ROI Localization}

The YOLOv12-m detector was initialized from COCO-pretrained weights and trained at an input resolution of $1024\times1024$ using the same patient-level split as the scoring experiments, with mosaic, mild affine, and HSV-V augmentations and flip augmentation disabled.
Table~\ref{tab:joint_detection_AP} summarizes the performance of the YOLOv12-m joint detector on the test set. The detector achieved mAP@50 values of 0.9914 and 0.9945 for BE and JSN joint localization, respectively, with both precision and recall exceeding 0.98. These results indicate that the proposed detector accurately localizes the clinically defined joint ROIs required for downstream SvdH scoring, providing a reliable basis for evaluating the automatic ROI localization-to-scoring pipeline.

\begin{table}[!t]
\centering
\caption{Performance of the automatic joint ROI detector on the test set.}
\label{tab:joint_detection_AP}
\resizebox{0.8\linewidth}{!}{
\setlength{\tabcolsep}{10pt}
\begin{tabular}{lccccc}
\toprule
\textbf{Target} & \textbf{Classes} & \textbf{mAP@50} & \textbf{mAP@50--95} & \textbf{Precision} & \textbf{Recall} \\
\midrule
BE  & 16 & 0.9914 & 0.8202 & 0.9931 & 0.9828 \\
JSN & 15 & 0.9945 & 0.7961 & 0.9956 & 0.9954 \\
\bottomrule
\end{tabular}}
\end{table}

\subsubsection{SvdH BE Scoring using Automatically Detected ROIs}
\label{sec:auto_roi_be_AP}
To assess the influence of automatic ROI localization on downstream BE scoring, all SvdH BE scoring models were re-evaluated using automatically detected joint ROIs. Table~\ref{tab:auto_roi_be_AP} compares the scoring performance obtained using ground-truth and automatically detected ROIs, together with the paired patient-level bootstrap confidence intervals of $\Delta$QWK.

Overall, replacing manually annotated ROIs with automatically detected ones resulted in only minor changes in BE scoring performance. Across the nine models, the average $\Delta$QWK was +0.0003 (95\% CI: $[-0.0190,+0.0212]$), indicating highly consistent overall performance with that obtained using ground-truth ROIs. Although the overall difference remained close to zero, MobileViT and MambaVision-T exhibited confidence intervals that excluded zero, indicating statistically significant performance degradation for these two models. The detected-ROI MAE remained comparable across all models, suggesting that automatic joint ROI localization provides sufficiently accurate localization for reliable downstream BE scoring.

\begin{table}[!t]
\centering
\caption{Comparison of SvdH BE scoring performance using ground-truth (GT) and automatically detected (Det) joint ROIs. $\Delta$QWK is estimated using paired patient-level bootstrap ($B=2000$), with 95\% confidence intervals reported in parentheses.}
\label{tab:auto_roi_be_AP}

\begin{threeparttable}
\resizebox{0.85\columnwidth}{!}{
\setlength{\tabcolsep}{12pt}
\begin{tabular}{lcccc}
\toprule
\textbf{Model} &
\textbf{GT ROI} &
\textbf{Det ROI} &
\textbf{$\Delta$QWK (95\% CI)} &
\textbf{MAE} \\
\midrule

ResNet
&0.4408&0.4686&+0.0266 $[-0.0084,+0.0659]$&0.3763\\

DenseNet
&0.3905&0.4076&+0.0177 $[-0.0104,+0.0569]$&0.3617\\

EfficientNetV2
&0.3358&0.3437&+0.0082 $[-0.0304,+0.0508]$&0.3917\\

MobileViT
&0.3920&0.3666&-0.0252 $[-0.0504,-0.0010]$&0.4107\\

LeViT
&0.2346&0.2367&+0.0021 $[-0.0553,+0.0683]$&0.4254\\

EfficientFormer
&0.3504&0.3492&-0.0012 $[-0.0346,+0.0246]$&0.3919\\

ConvNeXtV2
&0.3058&0.3119&+0.0063 $[-0.0243,+0.0314]$&0.4118\\

MedMamba
&0.4522&0.4504&-0.0018 $[-0.0282,+0.0213]$&0.3919\\

MambaVision
&0.3667&0.3364&-0.0300 $[-0.0524,-0.0054]$&0.3845\\

\midrule

Average
&0.3632&0.3635&
\textbf{+0.0003 $[-0.0190,+0.0212]$}
&0.3940\\

\bottomrule
\end{tabular}
}
\end{threeparttable}
\end{table}

\subsubsection{SvdH JSN Scoring using Automatically Detected ROIs}
\label{sec:auto_roi_jsn_AP}
The same evaluation was performed for SvdH JSN scoring models using automatically detected joint ROIs. Table~\ref{tab:auto_roi_jsn_AP} summarizes the corresponding scoring performance and paired patient-level bootstrap confidence intervals.

Across all evaluated models, the average $\Delta$QWK was $-0.0057$ (95\% CI: $[-0.0228,+0.0107]$), indicating only a minor overall difference compared with GT ROIs. Among all evaluated models, only MedMamba exhibited a confidence interval that excluded zero, while the remaining models showed no statistically significant performance change. The detected-ROI MAE also remained stable across models, demonstrating that automatic ROI localization has only a limited influence on downstream JSN scoring.

\begin{table}[!t]
\centering
\caption{Comparison of SvdH JSN scoring performance using ground-truth (GT) and automatically detected (Det) joint ROIs. $\Delta$QWK is estimated using paired patient-level bootstrap ($B=2000$), with 95\% confidence intervals reported in parentheses.}
\label{tab:auto_roi_jsn_AP}

\begin{threeparttable}
\resizebox{0.85\columnwidth}{!}{
\setlength{\tabcolsep}{12pt}
\begin{tabular}{lcccc}
\toprule
\textbf{Model} &
\textbf{GT ROI} &
\textbf{Det ROI} &
\textbf{$\Delta$QWK (95\% CI)} &
\textbf{MAE} \\
\midrule

ResNet
&0.5884&0.5885&+0.0001 $[-0.0281,+0.0308]$&0.2789\\

DenseNet
&0.5829&0.5756&-0.0074 $[-0.0277,+0.0088]$&0.2996\\

EfficientNetV2
&0.5393&0.5354&-0.0038 $[-0.0377,+0.0332]$&0.2996\\

MobileViT
&0.5967&0.6025&+0.0056 $[-0.0171,+0.0304]$&0.2844\\

LeViT
&0.5445&0.5516&+0.0073 $[-0.0282,+0.0458]$&0.2999\\

EfficientFormer
&0.5919&0.5813&-0.0108 $[-0.0343,+0.0136]$&0.2759\\

ConvNeXtV2
&0.5151&0.5108&-0.0044 $[-0.0210,+0.0157]$&0.2894\\

MedMamba
&0.5738&0.5356&-0.0378 $[-0.0732,-0.0051]$&0.3051\\

MambaVision
&0.5457&0.5462&+0.0004 $[-0.0174,+0.0228]$&0.2909\\

\midrule

Average
&0.5643&0.5588&
\textbf{-0.0057 $[-0.0228,+0.0107]$}
&0.2915\\

\bottomrule
\end{tabular}
}
\end{threeparttable}
\end{table}

\subsection{Cross-Task Evaluation with Intermediate Segmentation Masks}
\label{sec:cross_task_mask_AP}

To evaluate whether intermediate segmentation information can support downstream SvdH scoring, we concatenate the corresponding segmentation mask with each radiographic ROI as an additional input channel. We evaluate SvdH-BE-90 segmentation masks for BE scoring and joint-specific bone segmentation masks for JSN scoring using the same models as in the corresponding scoring benchmarks.

\subsubsection{SvdH BE Scoring with BE Segmentation Masks}
\label{sec:be_mask_scoring_AP}

For BE scoring, we evaluate three mask conditions: GT/GT, GT/Pred, and Pred/Pred, denoting the mask sources used for training and inference, respectively. Predicted masks are generated using SwinUMamba, the best-performing SvdH-BE-90 segmentation model in our benchmark. Table~\ref{tab:be_mask_ablation_AP} reports changes relative to the corresponding no-mask baseline for each model.

\begin{table}[!t]
\centering
\caption{Effect of BE segmentation masks on SvdH BE scoring. All values denote changes relative to the corresponding no-mask baseline for the same model. GT and Pred denote ground-truth and predicted masks, respectively.}
\label{tab:be_mask_ablation_AP}
\resizebox{0.9\linewidth}{!}{
\setlength{\tabcolsep}{8pt}
\begin{tabular}{lccccc}
\toprule
\textbf{Model} &
\textbf{$\Delta$QWK} &
\textbf{$\Delta$MAE} &
\textbf{$\Delta$BACC (\%)} &
\textbf{$\Delta$P/N-SEN (\%)} &
\textbf{$\Delta$P/N-ACC (\%)} \\
\midrule
\multicolumn{6}{c}{\textit{Train: GT \quad / \quad Infer: GT}} \\
\midrule
ResNet           & +0.0581 & -0.0368 & -1.76 & +4.16  & +2.83 \\
DenseNet         & +0.0905 & -0.0405 & +3.15 & +4.16  & +3.89 \\
MedMamba         & +0.0890 & -0.0325 & +3.92 & +11.09 & +3.02 \\
EfficientFormer  & +0.0261 & -0.0197 & -0.86 & +2.69  & +2.41 \\
LeViT            & +0.0418 & -0.0393 & -0.33 & +10.05 & +3.84 \\
MobileViT        & +0.0982 & -0.0143 & +4.43 & +14.73 & +1.24 \\
ConvNeXtV2       & +0.1284 & -0.0375 & +3.08 & +15.68 & +3.21 \\
EfficientNetV2   & +0.0305 & -0.0213 & +0.14 & +9.01  & +2.20 \\
MambaVision      & +0.0409 & -0.0410 & -0.13 & +6.85  & +3.82 \\
\midrule
\textbf{Mean} &
\textbf{+0.0670} &
\textbf{-0.0314} &
\textbf{+1.29} &
\textbf{+8.71} &
\textbf{+2.94} \\
\midrule
\multicolumn{6}{c}{\textit{Train: GT \quad / \quad Infer: Pred}} \\
\midrule
ResNet           & +0.0288 & +0.0302 & -1.09 & +22.10 & -2.20 \\
DenseNet         & +0.0722 & +0.0147 & +3.04 & +20.71 & -0.63 \\
MedMamba         & +0.0352 & +0.0522 & +2.76 & +28.34 & -2.46 \\
EfficientFormer  & +0.0106 & +0.0330 & +0.93 & +19.67 & -2.25 \\
LeViT            & +0.0163 & +0.0274 & +0.24 & +29.90 & -1.69 \\
MobileViT        & +0.0781 & +0.0581 & +4.07 & +32.67 & -4.05 \\
ConvNeXtV2       & +0.0451 & +0.0475 & +2.35 & +22.53 & -1.45 \\
EfficientNetV2   & -0.0037 & +0.0424 & -0.21 & +20.10 & -2.74 \\
MambaVision      & -0.0223 & +0.0684 & -0.44 & +27.30 & -4.54 \\
\midrule
\textbf{Mean} &
\textbf{+0.0289} &
\textbf{+0.0415} &
\textbf{+1.29} &
\textbf{+24.81} &
\textbf{-2.44} \\
\midrule
\multicolumn{6}{c}{\textit{Train: Pred \quad / \quad Infer: Pred}} \\
\midrule
ResNet           & +0.0151 & -0.0225 & +1.42 & -7.02 & +2.41 \\
DenseNet         & +0.0271 & -0.0075 & +0.39 & +6.07 & +0.33 \\
MedMamba         & +0.0190 & -0.0286 & -1.97 & +13.69 & +3.00 \\
EfficientFormer  & -0.0636 & +0.0080 & -2.30 & -3.03 & -0.23 \\
LeViT            & -0.0304 & +0.0162 & -0.88 & +5.72 & -1.17 \\
MobileViT        & -0.0043 & -0.0035 & +1.07 & -2.25 & +0.44 \\
ConvNeXtV2       & +0.0092 & -0.0101 & -1.45 & +12.13 & +0.80 \\
EfficientNetV2   & +0.0190 & -0.0047 & +1.64 & +5.46 & +0.19 \\
MambaVision      & -0.1212 & -0.0080 & -6.01 & -4.68 & +0.96 \\
\midrule
\textbf{Mean} &
\textbf{-0.0145} &
\textbf{-0.0067} &
\textbf{-0.90} &
\textbf{+2.90} &
\textbf{+0.75} \\
\bottomrule
\end{tabular}
}
\end{table}

As shown in Table~\ref{tab:be_mask_ablation_AP}, GT BE masks improve downstream scoring across all models in terms of QWK, yielding a mean $\Delta$QWK of +0.067. The average BACC also increases by 1.29 \%, although this improvement is not consistent across individual models (5/9 models improve). Mean MAE, P/N-SEN, and P/N-ACC also improve relative to the no-mask baselines. These results indicate across the full set of scoring models that lesion-level BE annotations provide information that can benefit downstream SvdH BE scoring, while the magnitude of the benefit depends on the evaluation metric.

The benefit becomes less consistent when predicted masks are introduced. With GT masks for training and predicted masks at inference, seven of nine models retain a positive QWK change (mean $\Delta$QWK = +0.0289), and six of nine models improve in BACC, with a mean increase of 1.29 \%. However, this setting exhibits a clear trade-off across metrics: P/N-SEN increases substantially, whereas MAE increases and P/N-ACC decreases, suggesting that the predicted masks alter the balance between positive-case sensitivity and overall scoring errors. When predicted masks are used for both training and inference, the mean QWK change becomes slightly negative ($-0.0145$), mean BACC decreases by 0.9 \%, and the effects vary substantially across models. Overall, accurate BE masks provide measurable information for downstream BE scoring, whereas the benefit is only partially retained when replacing GT masks with the current predicted masks.

\subsubsection{SvdH JSN Scoring with Bone Segmentation Masks}
\label{sec:jsn_mask_scoring_AP}

For JSN scoring, the masks of the two bones adjacent to each evaluated joint are provided as additional input channels. We evaluate the same three Train/Infer conditions as for BE scoring: GT/GT, GT/Pred, and Pred/Pred. The validation mask source follows the training mask source. All changes in Table~\ref{tab:jsn_mask_ablation_AP} are calculated relative to the corresponding no-mask scoring baseline for each model.

\begin{table}[!t]
\centering
\caption{Effect of joint-specific bone segmentation masks on SvdH JSN scoring. All values denote changes relative to the corresponding no-mask baseline for the same model. GT and Pred denote ground-truth and predicted masks, respectively.}
\label{tab:jsn_mask_ablation_AP}
\resizebox{0.9\linewidth}{!}{
\setlength{\tabcolsep}{10pt}
\begin{tabular}{lccccc}
\toprule
\textbf{Model} &
\textbf{$\Delta$QWK} &
\textbf{$\Delta$MAE} &
\textbf{$\Delta$BACC (\%)} &
\textbf{$\Delta$P/N-SEN (\%)} &
\textbf{$\Delta$P/N-ACC (\%)} \\
\midrule
\multicolumn{6}{c}{\textit{Train: GT \quad / \quad Infer: GT}} \\
\midrule
ResNet           & -0.0345 & +0.0397 & -1.83 & -1.02 & -2.27 \\
DenseNet         & -0.0172 & -0.0087 & -1.39 & -5.09 & +0.67 \\
MedMamba         & -0.0021 & -0.0043 & +0.32 & +0.22 & +0.37 \\
EfficientFormer  & -0.0379 & +0.0337 & -2.97 & -2.38 & -3.29 \\
LeViT            & -0.0071 & -0.0022 & -2.12 & -6.23 & +0.18 \\
MobileViT        & -0.0358 & +0.0000 & -2.19 & -5.21 & -0.50 \\
ConvNeXtV2       & +0.0029 & -0.0020 & +0.36 & +0.46 & +0.30 \\
EfficientNetV2   & -0.0063 & +0.0040 & -1.33 & -1.47 & -1.25 \\
MambaVision      & +0.0152 & -0.0045 & -2.61 & -7.36 & +0.05 \\
\midrule
\textbf{Mean}    & \textbf{-0.0137} & \textbf{+0.0062} & \textbf{-1.53} & \textbf{-3.12} & \textbf{-0.64} \\
\midrule
\multicolumn{6}{c}{\textit{Train: GT \quad / \quad Infer: Pred}} \\
\midrule
ResNet           & -0.0454 & +0.0365 & -3.39 & -4.08 & -2.99 \\
DenseNet         & -0.0106 & -0.0100 & -1.21 & -4.98 & +0.90 \\
MedMamba         & -0.0313 & -0.0010 & -1.13 & -3.51 & +0.20 \\
EfficientFormer  & -0.0527 & +0.0295 & -3.19 & -3.74 & -2.87 \\
LeViT            & -0.0281 & +0.0040 & -2.58 & -6.79 & -0.22 \\
MobileViT        & -0.0885 & +0.0138 & -4.37 & -9.40 & -1.55 \\
ConvNeXtV2       & -0.0385 & +0.0040 & -0.39 & -1.70 & +0.35 \\
EfficientNetV2   & -0.0483 & +0.0090 & -2.52 & -4.30 & -1.52 \\
MambaVision      & -0.0054 & -0.0055 & -3.23 & -9.06 & +0.02 \\
\midrule
\textbf{Mean}    & \textbf{-0.0388} & \textbf{+0.0089} & \textbf{-2.44} & \textbf{-5.28} & \textbf{-0.85} \\
\midrule
\multicolumn{6}{c}{\textit{Train: Pred \quad / \quad Infer: Pred}} \\
\midrule
ResNet           & -0.0026 & +0.0167 & -0.54 & +1.02 & -1.39 \\
DenseNet         & +0.0255 & -0.0097 & -1.82 & -7.25 & +1.22 \\
MedMamba         & -0.0114 & +0.0274 & +1.50 & +7.70 & -1.98 \\
EfficientFormer  & -0.0788 & +0.0240 & -4.63 & -9.86 & -1.69 \\
LeViT            & +0.0147 & -0.0142 & -1.20 & -5.77 & +1.35 \\
MobileViT        & -0.0144 & +0.0028 & -3.84 & -10.65 & -0.03 \\
ConvNeXtV2       & -0.0232 & +0.0103 & -0.21 & +0.80 & -0.77 \\
EfficientNetV2   & +0.0035 & -0.0092 & -2.39 & -6.34 & -0.17 \\
MambaVision      & +0.0513 & -0.0202 & +0.17 & -1.58 & +1.15 \\
\midrule
\textbf{Mean}    & \textbf{-0.0040} & \textbf{+0.0031} & \textbf{-1.44} & \textbf{-3.55} & \textbf{-0.26} \\
\bottomrule
\end{tabular}
}
\end{table}

As shown in Table~\ref{tab:jsn_mask_ablation_AP}, joint-specific bone masks do not provide a consistent improvement in JSN scoring. Even under the GT/GT condition, mean QWK decreases by 0.0137, with only two of the nine models improving over their corresponding no-mask baselines. This contrasts with the BE scoring experiment, where GT lesion masks consistently improve QWK across all evaluated models. Thus, access to accurate bone boundaries alone does not appear to provide the same complementary scoring information for JSN as lesion-level BE masks do for BE scoring.

Replacing GT masks with predicted masks at inference further reduces performance: under GT/Pred, QWK decreases for all nine models, with a mean change of $-0.0388$. When predicted masks are used during both model development and inference, the effect is smaller but remains inconsistent, with a mean $\Delta$QWK of $-0.0040$ and improvements in four of nine models. The absence of a consistent gain even under the GT/GT condition indicates that the limited benefit cannot be attributed solely to segmentation errors in the predicted masks. Bone-mask prompting therefore does not consistently improve joint-level JSN scoring under the evaluated conditions. This finding should be considered separately from the hand-level correlation analysis in Fig.~\ref{fig:statistics_overview}(F). The effectiveness of bone-mask prompting may depend on multiple factors, including the mask representation, model architecture, and training procedure. The granularity of the incorporated anatomical information may also affect its utility for joint-level scoring.

Together with the BE results, these experiments show that the utility of intermediate segmentation information for downstream scoring is task dependent. Lesion-level BE masks provide complementary information for BE severity scoring when accurate masks are available, whereas joint-specific bone masks yield limited and model-dependent changes in JSN scoring. Future work may explore more robust integration of predicted lesion information for BE scoring and alternative ways of representing and incorporating anatomical information for JSN scoring.

\section{Discussion}
The benchmark results reveal a consistent performance gap across tasks, indicating that fine-grained pathological analysis remains substantially more challenging than global anatomical modeling. These challenges are fundamentally rooted in both the characteristics of radiographic imaging and the pathological nature of RA.

From an imaging perspective, 2D radiographs inherently suffer from projection-induced ambiguity, limited contrast, and the absence of depth information. From a disease perspective, RA manifests through subtle, heterogeneous, and progressively evolving structural changes, which are often difficult to localize and quantify. These factors jointly impose intrinsic limitations on all downstream tasks.

At the task level, distinct challenges can be observed. For hand bone structure segmentation, although overall performance is high, projection-induced overlap, particularly in anatomically dense regions such as the wrist, significantly obscures boundaries and degrades accuracy in overlap-sensitive regions. In addition, severe structural damage in advanced RA, including bone deformation and collapse, further complicates reliable delineation.

For BE analysis, the primary difficulty lies in the ambiguous and subjective nature of lesion annotation. Erosive lesions are typically small, irregular, and locally indistinct, especially in early-stage RA where radiographic signals are weak. As a result, annotation often depends on expert interpretation, introducing inter-observer variability and uncertainty in the ground truth. This ambiguity limits both the reliability of supervision and the upper bound of achievable performance.

For SvdH-based scoring, additional challenges arise from its semi-quantitative and ordinal nature. The scoring criteria are inherently coarse and subject to interpretation, particularly at boundary levels between adjacent grades. Moreover, the distribution of scores is highly imbalanced, with a long-tail effect where severe cases are relatively rare in modern clinical cohorts. These factors introduce both label ambiguity and data imbalance, making accurate prediction difficult and further weakening the assumption that ground truth labels are fully reliable.

These observations also reflect several limitations of the current evaluation and dataset. First, RAM-H1200 follows a depth-first design: its scale is limited to 1,200 radiographs from 291 patients, while providing detailed annotations spanning whole-hand anatomy, pixel-level BE lesions, and joint-level BE and JSN scores within the same cohort. This trade-off prioritizes annotation depth over cohort size and may limit the statistical diversity available for model development and evaluation. Second, although the dataset is collected from six institutions, all are located within the Sapporo region of Japan, making the cohort multi-institutional but geographically concentrated and demographically limited. The current center-held-out evaluation is limited to a single institution and therefore provides only an initial assessment of institutional distribution shift within this regional cohort, rather than external validation across geographically and demographically distinct populations. Third, high-severity cases are relatively scarce, and 176 end-stage RA radiographs exhibiting bony ankylosis were excluded during cohort construction. These radiographically conspicuous cases constitute a readily scored high-severity tail, and their exclusion further reduces the representation of end-stage disease. RAM-H1200 is therefore primarily intended for fine-grained assessment of early-to-moderate RA structural changes rather than for evaluation of end-stage ankylosis.

Furthermore, the inherent ambiguity in BE and SvdH annotations introduces uncertainty into both training and evaluation, and the provided labels should therefore be interpreted as expert consensus references rather than error-free absolute ground truth. Our independent-reader analyses provide additional evidence on annotation variability by comparing an independent radiologist with the consensus reference for both SvdH scoring and BE segmentation. However, these analyses do not capture intra-observer variability, as repeated assessments by the same reader were not performed.

The additional experiments also help identify where limitations arise within the current task pipeline. Automatic joint localization produces only minor changes in downstream BE and JSN scoring, suggesting that ROI localization is not the dominant bottleneck under the current setting. In contrast, the utility of intermediate segmentation information is task dependent. Accurate lesion-level BE masks provide complementary information for BE severity scoring, whereas the benefit decreases when predicted masks are used, highlighting the importance of robust lesion representation. For JSN, directly incorporating joint-specific bone masks yields limited and model-dependent improvements even when ground-truth masks are available, suggesting that simple mask concatenation may not adequately capture the anatomical relationships relevant to joint-space narrowing. Thus, although RAM-H1200 provides anatomically, pathologically, and clinically related annotations within the same cohort, the current benchmark primarily evaluates these tasks independently, and the optimal way to exploit their relationships remains an open problem.

These findings suggest that, despite strong progress in anatomical segmentation, fine-grained pathological analysis remains a key bottleneck for automated RA assessment. Future work may therefore extend RAM-H1200 toward larger and geographically broader cohorts, incorporate repeated and multi-reader annotations for more explicit modeling of label uncertainty, and develop approaches that more effectively exploit relationships across anatomical structure, localized pathology, and clinical scoring. In particular, robust integration of predicted lesion information may improve BE scoring, while JSN assessment may benefit from representations that more explicitly encode joint-space geometry and spatial relationships between adjacent bones rather than directly concatenating segmentation masks.

\section{Broader Impact}
\label{sec:broader_impact}

This work introduces a comprehensively annotated multi-task dataset and evaluation benchmark for RA analysis based on hand radiographs. By providing anatomically detailed annotations and clinically grounded evaluation protocols, the dataset has the potential to facilitate the development of computer-aided diagnosis systems that improve the efficiency and consistency of RA assessment. In particular, automated analysis of BE and JSN may assist clinicians in early detection and longitudinal monitoring, which are critical for timely intervention and improved patient outcomes~\cite{gunkl2026bridging}. The public release of such a dataset may also promote reproducible research and lower the barrier for developing structure-aware and clinically interpretable models in medical imaging.

RAM-H1200 is intended for evaluating structure-aware segmentation, lesion-level BE quantification, and joint-level SvdH scoring, but not for evaluating standalone RA diagnosis, treatment recommendation, or deployment readiness across unseen populations. However, several potential risks and limitations should be considered. First, although the dataset includes six institutions, all are located within the Sapporo region of Japan, making the cohort geographically concentrated. It may therefore not fully represent the diversity of imaging protocols, populations, and disease presentations encountered in broader clinical practice. Models trained on this dataset may exhibit reduced generalization performance in other geographic regions and clinical environments. Second, annotation of BE and JSN involves inherent subjectivity, particularly in early-stage cases, which may introduce bias into both model training and evaluation. Third, automated systems developed using this dataset should not be used as standalone diagnostic tools, as incorrect predictions may lead to misinterpretation of disease severity and potentially impact clinical decision-making.

To mitigate these risks, we emphasize that this dataset is intended for research purposes and should be used to develop assistive tools rather than replace expert judgment. Future work may incorporate multi-institutional data from broader geographic regions and more diverse patient populations, uncertainty modeling, and human-in-the-loop validation to improve robustness and reliability. Careful evaluation under different clinical settings is necessary before any real-world deployment.



\end{document}